# AdaptAD: Foundation Model Adaptation with Missing Modality Robustness for Alzheimer's Disease Staging Across the Disease Continuum

Dawa Chyophel Lepcha [a, b, c], Sophie A. Martin [d, e], Pierrick Coupé [f], Shabbir Syed Abdul [a, b, g, *]

[a] Graduate Institute of Biomedical Informatics, College of Medical Science and Technology, Taipei Medical University, Taipei, 235, Taiwan.
[b] International Center for Health Information Technology, College of Medical Science and Technology, Taipei Medical University, Taipei, 235, Taiwan.
[c] Biomedical Sensors & Systems Lab, University of Memphis, Memphis, TN 38152, USA
[d] UCL Hawkes Institute, University College London, London, UK
[e] Queen Square Institute of Neurology, University College London, London,UK
[f] CNRS, University Bordeaux, Bordeaux INP, LABRI, UMR5800, Talence F-33405, France
[g] School of Gerontology and Long-term Care, College of Nursing, Taipei Medical University, Taipei, Taiwan, 110.

**Abstract**

Staging Alzheimer's disease (AD) across its continuum, from cognitively normal (CN) through mild cognitive impairment (MCI) to dementia from a single structural MRI remains difficult because 3-D networks overfit small cohorts, and FreeSurfer morphometry is often unavailable. We present AdaptAD, which reads a registered T1-weighted scan as 16 anatomically placed 2.5-D slices through the medial temporal lobe and adapts a frozen ImageNet-pretrained ViT-B/16 with low-rank adaptation (LoRA), training 3.5 M of 89 M parameters. Morphometry enters as tokens when present, or as a learned [MISSING] token, and a missing-aware stacker fuses both experts. Under a pre-registered protocol, ten methods were compared for three-class CN/MCI/AD staging on identical 5-fold splits of 597 ADNI participants, 29% without FreeSurfer, with Holm-corrected paired bootstrap tests. The locked model was tested once on 993 OASIS-3 participants (988 scored by every method). On ADNI, AdaptAD achieved a three-class macro-AUC of 0.745 (95% CI 0.709 to 0.781), ranking first and significantly exceeding full fine-tuning, ShaSpec and a 3-D CNN. On OASIS-3 it reached 0.780 (0.748 to 0.814), and on the CDR-based label (0.804) it significantly outperformed every 3-D and image-tabular method. LoRA matched full fine-tuning with 25 times fewer trainable parameters, and removing morphometry changed external macro-AUC by less than 0.003. The disease score tracked MMSE and CDR-SB within diagnostic groups and stratified MCI-to-dementia progression (log-rank $p < 0.001$), while saliency concentrated in medial temporal and periventricular regions. A parameter-efficient 2-D foundation model can therefore stage the full AD continuum from a routine T1 scan without morphometry, offering a scalable route to memory-clinic deployment.

## 1. Introduction

Alzheimer's disease (AD) develops along a continuum. Cognitively normal (CN) individuals accumulate pathology silently, many pass through a stage of mild cognitive impairment (MCI), and a proportion of them progress to dementia. Where a patient sits on this continuum decides counselling, eligibility for disease-modifying trials and therapies, and how closely they are followed. Structural MRI is acquired in almost every memory-clinic work-up, and atrophy of the medial temporal lobe with enlargement of the ventricles is among the best-established imaging markers of AD [1]. The NIA-AA research framework places neurodegeneration on this same biological continuum [2], and neuropathological staging shows that neurofibrillary change begins in the transentorhinal and entorhinal cortex and the hippocampus before it spreads to the neocortex [3]. In principle, then, a single T1-weighted scan carries information about the stage of disease. In practice, turning it into a reliable three-class stage of CN, MCI or AD is hard. Age-related atrophy overlaps with disease-related atrophy, and MCI is clinically and biologically mixed: some patients are in the prodromal phase of AD, others will remain stable or revert. A useful model must therefore do more than separate the extremes. It must place patients in the right order along the continuum, including the difficult middle stage. Deep learning has been applied to this problem for more than a decade, mostly with 3-D convolutional neural networks (CNNs) trained on whole registered volumes, patches or regions of interest. A systematic re-evaluation of this literature found that many high reported accuracies were inflated by data leakage between training and test subjects, and that carefully controlled 3-D CNNs were not clearly better than simpler approaches [4]. A structural reason lies behind this. 3-D networks have many parameters relative to the few hundred to few thousand labelled scans available in neuroimaging cohorts, and 3-D medical pretraining [5] draws on orders of magnitude fewer images than natural-image pretraining. Recent multimodal systems have reached strong results by combining imaging with demographics, cognitive tests, laboratory values and other clinical data [6], [7]. These systems are valuable, but they depend on a broad set of inputs that is rarely complete at the point of care, and they leave open how much can be extracted from the one examination nearly every patient receives: a routine T1-weighted scan.

Foundation models offer a different starting point. Vision transformers (ViTs) [8] pretrained on large natural-image collections with strong augmentation and regularisation [9] or by self-supervision [10] have become general-purpose feature extractors, and foundation models are increasingly seen as a route to generalist medical AI [11]. Their use in medical imaging faces specific obstacles, among them the gap between natural and medical images, the three-dimensional nature of many modalities, and the small size of labelled clinical datasets [12]. Parameter-efficient fine-tuning addresses the last of these. Low-rank adaptation (LoRA) [13] keeps the pretrained weights frozen and learns a small low-rank update to them, so that a model with tens of millions of parameters can be adapted by training about one percent of them. A recent benchmark argued that parameter-efficient fine-tuning is a missed opportunity in medical image analysis, with the largest benefits expected precisely in the low-data regime [14]. Whether a 2-D foundation model can read 3-D brain MRI well enough to stage the AD continuum, how the volume should be presented to it, and whether a small low-rank update suffices are open questions. Answering them convincingly requires a fair comparison against 3-D and multimodal alternatives on identical data, which has not been reported. A second, practical problem is that morphometry is often missing. FreeSurfer [15] provides regional volumes and cortical thickness that are informative for AD, but the pipeline takes hours per scan, needs visual quality control, fails on some images and is often run long before or after the scan of interest. In the ADNI cohort studied here, 176 of 597 participants (29%) had no quality-controlled FreeSurfer result within 180 days of their scan. Image and tabular fusion methods, such as feature-wise linear modulation (FiLM) [16], the dynamic affine feature-map transform (DAFT) [17] and shared-specific feature modelling (ShaSpec) [18], typically assume that the tabular input is present, fill absent values with zeros, or are evaluated under missingness simulated at random. A model intended for routine use should accept morphometry when it exists, never require it, and be evaluated under the missingness that actually occurs. Finally, the evidence base itself needs strengthening. Many reported AD classifiers are tuned and tested on the same data, give no confidence intervals or tests between methods, and are never evaluated on an independent cohort. Without these safeguards, small differences between methods cannot be distinguished from noise or selection, and reported performance rarely survives a change of site, scanner or diagnostic convention.

We present AdaptAD, a foundation-model approach to three-class staging of the AD continuum, designed around these three problems. Rather than training a 3-D network from limited data, AdaptAD brings the volume to a pretrained 2-D model. A registered T1-weighted scan is read as 16 slices at fixed positions in MNI space, eight coronal and eight axial, placed to cross the medial temporal lobe and the structures around it, where neurodegeneration in AD begins [3]. Each slice is combined with its two neighbours into a three-channel image, so that every image carries through-plane context and every participant contributes the same anatomy. A frozen ImageNet-pretrained ViT-B/16 encodes these images, and LoRA [13] adapts it by learning low-rank updates to its attention and feed-forward projections, leaving the pretrained weights intact and keeping the number of trained parameters small enough for cohorts of a few hundred participants. FreeSurfer morphometry [15] enters as a short sequence of tokens alongside the MRI tokens. When it is absent, these tokens are replaced by a learned [MISSING] token, so that the network always sees the same input structure and learns what absent morphometry looks like instead of reading zeros as measurements. Training proceeds in two stages. An MRI-only warm-up first establishes a strong image pathway; joint training then exposes the network to morphometry while hiding it at random and distilling the prediction made with morphometry into the prediction made without it. At inference, a missing-aware late-fusion stacker, fitted on validation data only, combines the MRI expert with a morphometry expert when morphometry is available and reduces to a calibrated MRI-only model when it is not, so that no prediction ever depends on FreeSurfer being present. We evaluate AdaptAD against nine alternatives under a pre-registered protocol, examine how it behaves as morphometry is withheld, relate its output to clinical severity and progression, and test the locked model once on an independent external cohort. Our contributions are as follows. Our contributions are as follows.

1. We cast three-class staging of CN, MCI and AD as reading 16 anatomically fixed 2.5-D slice tokens, each with through-plane context, so that every participant contributes the same anatomy to a frozen ImageNet-pretrained ViT-B/16. LoRA adapts the backbone with 1.18 M parameters (3.53 M trainable in total, 3.9% of the model). Staging is evaluated by three-class macro-AUC and by quadratic weighted kappa, which rewards correct ordering of the stages. A controlled comparison within one backbone and seed shows that adaptation is necessary, since the frozen linear probe was the weakest variant, and that LoRA matches full fine-tuning with 25 times fewer trainable parameters.
2. We design AdaptAD to be robust to missing morphometry by combining a learned [MISSING] token, FreeSurfer dropout, distillation from the prediction with morphometry to the prediction without it, and a missing-aware late-fusion stacker whose missing branch reduces to a calibrated MRI-only model. Unlike prior fusion methods, it is evaluated under the natural missingness of ADNI (29% of participants) and under controlled removal of morphometry from 0% to 100% of an external cohort. Removing morphometry changed external macro-AUC by less than 0.003. Ablations attribute this robustness mainly to the strength of the MRI-only path rather than to any single missing-modality component, a finding we report as it stands.
3. We compare ten methods under a pre-registered protocol: AdaptAD, 3-D CNNs trained from scratch and with medical pretraining, three image and tabular fusion methods (FiLM, DAFT and ShaSpec), full ViT fine-tuning and a morphometry-only model, all trained on identical subject-level folds. All design decisions were taken at three pre-registered gates on validation data only, differences were assessed with paired bootstrap tests and Holm correction, and the final method was locked by cryptographic hashes before it was tested once on 993 OASIS-3 participants with endpoints, including a CDR-based secondary label, fixed in advance.
4. We show that AdaptAD is clinically and anatomically meaningful along the continuum. It yields a continuous disease score that correlates with MMSE and CDR-SB in both cohorts and within each diagnostic group, separates MCI participants who progress to dementia from those who remain stable, and stratifies time to dementia. Saliency analysis, checked with a model-randomisation test, shows that the model attends to periventricular and medial temporal regions affected early in AD and largely ignores regions spared by the disease.

The remainder of the paper is organised as follows. Section 2 reviews related work on deep learning for AD from structural MRI, foundation models and parameter-efficient adaptation, image and tabular fusion with missing modalities, and evaluation practice. Section 3 describes AdaptAD. Section 4 presents the cohorts, comparison methods and pre-registered protocol. Section 5 reports the results, Section 6 discusses the implications and limitations, and Section 7 concludes the study.

## 2. Related work

AdaptAD draws on three lines of research: deep learning for AD staging from structural MRI, foundation models and their parameter-efficient adaptation, and the fusion of images with tabular data when one modality may be missing. A fourth line, on evaluation practice in medical image analysis, shaped the protocol by which we compare methods. We review each in turn and state how AdaptAD relates to it.

### 2.1 Deep learning for AD staging from structural MRI

Most image-based AD classifiers are 3-D convolutional networks that operate on whole registered volumes, on patches sampled from them, or on predefined regions of interest such as the hippocampus. Wen et al. [4] reviewed this literature systematically and showed that data leakage, for example placing scans of the same participant in both training and test sets or selecting models on test data, had inflated many reported accuracies. They released a reproducible framework in which a compact 3-D CNN trained from scratch served as a strong and honest baseline, and we adopt that network as one of our comparison methods. Because 3-D networks have many parameters relative to the size of neuroimaging cohorts, transfer learning has been widely explored. MedicalNet [5] pretrained 3-D ResNets on segmentation tasks pooled from 23 medical datasets and improved convergence and accuracy on downstream tasks, but such medical 3-D pretraining remains small compared with natural-image pretraining. A parallel line of 2-D and 2.5-D methods feeds individual slices, or stacks of adjacent slices, to 2-D networks pretrained on ImageNet. These methods inherit powerful 2-D features, but they typically select slices by hand or by an image statistic such as entropy, discard most 3-D context, and can vary with the slices chosen for each participant.

Transformers have entered this field quickly. Shamshad et al. [19] surveyed transformers across medical imaging tasks and identified their need for large training sets or strong pretraining as a central obstacle in clinical domains. Mubonanyikuzo et al. [20] systematically reviewed and meta-analysed ViT-based detection of AD from neuroimaging, documenting how rapidly such models have been adopted for this task. Dhinagar et al. [21] examined how to train ViTs efficiently on structural MRI for AD detection, and found that the choice of pretraining is decisive when labelled scans are scarce. Most recently, foundation models have been built for brain MRI itself. Kaczmarek et al. [22] trained a self-supervised SimCLR model on 3-D brain MRI from several neurological diseases to serve as a general starting point for diagnosis, and BrainIAC [23] pretrained a 3-D model on a large collection of unlabelled brain MRI and adapted it to tasks ranging from brain age to disease classification and survival prediction. These models are promising, but they require large-scale 3-D pretraining that few groups can reproduce. AdaptAD takes a complementary route. It fixes the slice positions in MNI space, so that every participant contributes the same anatomy at the same location, adds through-plane context by encoding three neighbouring slices as the channels of each image, and reuses a 2-D foundation model that is already available, adapting it with a small number of parameters.

### 2.2 Foundation models and parameter-efficient adaptation

Vision transformers [8] split an image into patches and process them with self-attention, and they reach their full strength when pretrained on very large datasets. Steiner et al. [9] showed that with appropriate augmentation and regularisation, ViTs pretrained on ImageNet-21k transfer well to a wide range of downstream tasks, and the backbone used in AdaptAD comes from this line of work. Self-supervised pretraining offers an alternative: DINOv2 [10] learns general-purpose visual features without labels from a large curated image collection. In medicine, foundation models pretrained on large unlabelled collections have improved disease detection from retinal images [24] and the discovery of imaging biomarkers from computed tomography in cancer [25], with the largest gains when labelled data are limited. Adapting such models to a new task by fully fine-tuning every weight is expensive, requires a separate copy of the model for each task and, with a few hundred labelled participants, risks overfitting. Parameter-efficient fine-tuning instead trains a small set of added or selected weights. Adapters [26] insert small bottleneck layers into each transformer block, and LoRA [13] adds a trainable low-rank update to frozen weight matrices, which can be merged into the original weights after training so that inference costs nothing extra. In medical image diagnosis, MeLo [27] applied LoRA to ViTs and reported performance comparable to full fine-tuning across several diagnostic tasks while storing only a small fraction of the parameters per task. Dutt et al. [14] benchmarked a broad range of parameter-efficient methods on medical imaging tasks and concluded that they are underused, with their advantage most evident in the low-data regime typical of clinical research. Despite this evidence, parameter-efficient adaptation of 2-D foundation models to 3-D brain MRI for staging the AD continuum has received little attention, and it has rarely been compared, within the same backbone and training seed, against a frozen linear probe, different adaptation ranks and full fine-tuning. AdaptAD applies LoRA to every attention and feed-forward projection of a ViT-B/16, performs exactly this controlled comparison, and compares the ViT backbone against DINOv2 at a pre-registered decision gate.

### 2.3 Image and tabular fusion with missing modalities

Morphometry derived with FreeSurfer [15], such as regional volumes and cortical thickness, is the most common tabular companion to structural MRI in AD research. Feature-wise linear modulation (FiLM) [16] conditions convolutional feature maps on side information through a learned scale and shift, and DAFT [17] extends this idea to a dynamic affine transform of a 3-D ResNet block driven by tabular data, designed specifically for AD diagnosis and prognosis. Several strategies exist for the case in which a modality is missing. The simplest fills absent values with zeros and adds an indicator, letting the network learn to ignore them. Modality dropout hides an available modality at random during training, so that the model does not come to rely on it. Knowledge distillation [28] can transfer the behaviour of a model that sees all modalities to one that sees fewer. ShaSpec [18] learns shared and modality-specific representations, so that a missing modality can be compensated from the shared space. For transformers, an absent modality can be represented by a learned token, and Ma et al. [29] showed that the robustness of multimodal transformers to missing inputs depends strongly on how and where the modalities are fused. Large multimodal systems for dementia have combined MRI with demographics, medical history and neuropsychological testing [6], and transformer-based models have extended this to the differential diagnosis of dementia aetiologies from heterogeneous clinical records designed to tolerate incomplete inputs [7]. Feng et al. [30] proposed a unified framework for AD diagnosis that accepts any combination of available modalities. Late fusion offers a further option: stacked generalisation [31] trains a second-level model on the outputs of modality-specific experts, so that each expert can be used or dropped independently. A common limitation across this literature is that robustness is usually assessed by removing modalities at random from complete datasets, which does not reproduce the structured missingness of clinical data, where morphometry is absent because a pipeline failed, did not pass quality control or was run at another time. AdaptAD combines a learned [MISSING] token, morphometry dropout and distillation inside the network with a missing-aware stacker outside it, and we evaluate it, DAFT, FiLM and ShaSpec under the missingness that occurs naturally in ADNI, complemented by controlled removal of morphometry in an external cohort.

### 2.4 Evaluation rigour and deployment

Reviews of machine learning for medical imaging have identified small and leaky test sets, model selection on test data, missing confidence intervals and the absence of external validation as recurring methodological failures [32], and reporting guidelines for AI in medical imaging and clinical prediction now ask explicitly for these elements [33], [34]. Discrimination is not the only concern. Modern deep networks are often poorly calibrated [35], and their calibration degrades further under dataset shift [36]. When a model is deployed in a population whose class prevalence differs from that of the development data, both its calibration and several common metrics change, which calls for explicit prevalence-aware evaluation and recalibration [37]. Finally, when several methods are compared, multiplicity must be controlled, for example by Holm's sequentially rejective procedure [38]. Our protocol responds to each point. Every method uses identical subject-level folds with no participant in more than one split. All design choices are taken at pre-registered gates on validation data. Calibration is fitted on validation data for every method in the same way. Differences are reported with paired bootstrap confidence intervals and Holm correction. The locked model is tested once on an external cohort whose prevalence differs from that of the development cohort, and calibration under this shift is reported alongside discrimination

## 3. Methods

AdaptAD stages a participant on the CN, MCI and AD continuum from a single T1-weighted MRI. The scan is read as 16 anatomically placed 2.5-D images and encoded by a frozen ImageNet vision transformer adapted with low-rank updates. The image representation is fused with FreeSurfer morphometry when it exists and with a learned placeholder when it does not, and a missing-aware stacker produces the final calibrated prediction. Fig. 1 summarises the approach, and the following subsections describe each component in the order in which data flow through the model. Section 3.1 formulates the staging problem and defines the continuous disease score. Section 3.2 describes how each registered scan is sampled into 16 anatomically placed 2.5-D slice tokens through the medial temporal lobe, and Section 3.3 presents the ViT-B/16 encoder and its low-rank adaptation. Section 3.4 introduces the FreeSurfer tokens and the learned [MISSING] token that replaces them when morphometry is unavailable, and Section 3.5 describes the fusion transformer and its prediction heads. Section 3.6 sets out the two-stage training with FreeSurfer dropout and distillation, and Section 3.7 the missing-aware late fusion and calibration that produce the final prediction.

### 3.1 Problem formulation

Let $\mathcal{D} = \{(V_i, \mathbf{f}_i, m_i, y_i)\}$, $i = 1, \ldots, N$, denote a cohort of N participants. $V_i$ is a T1-weighted volume registered to the MNI152 template. $\mathbf{f}_i \in \mathbb{R}^{132}$ is a vector of FreeSurfer features, $m_i \in \{0, 1\}$ indicates whether that vector is available, and $y_i \in \{0, 1, 2\}$ encodes the diagnosis CN, MCI or AD recorded within 180 days of the scan. The aim is a function that returns a probability vector over the three stages for every participant, whether or not morphometry is available:

$$\mathbf{p}_i = F(V_i,\ m_i\mathbf{f}_i,\ m_i) \in \Delta^2, \qquad \Delta^2 = \left\{\mathbf{p} \in [0,1]^3 : \sum_c p_c = 1\right\} \tag{1}$$

Because the three classes are ordered stages of one disease rather than unrelated categories, the predicted distribution also defines a continuous disease score,

$$s_i = p_{i,\mathrm{MCI}} + 2\,p_{i,\mathrm{AD}} \in [0,2] \tag{2}$$

which is the expected stage under the model. A confident CN prediction gives $s_i = 0$ and a confident AD prediction gives $s_i = 2$, while uncertain or intermediate cases fall in between. The score plays no part in training. It is used in Section 4.6 to examine whether the model's output tracks clinical severity and progression across the continuum.

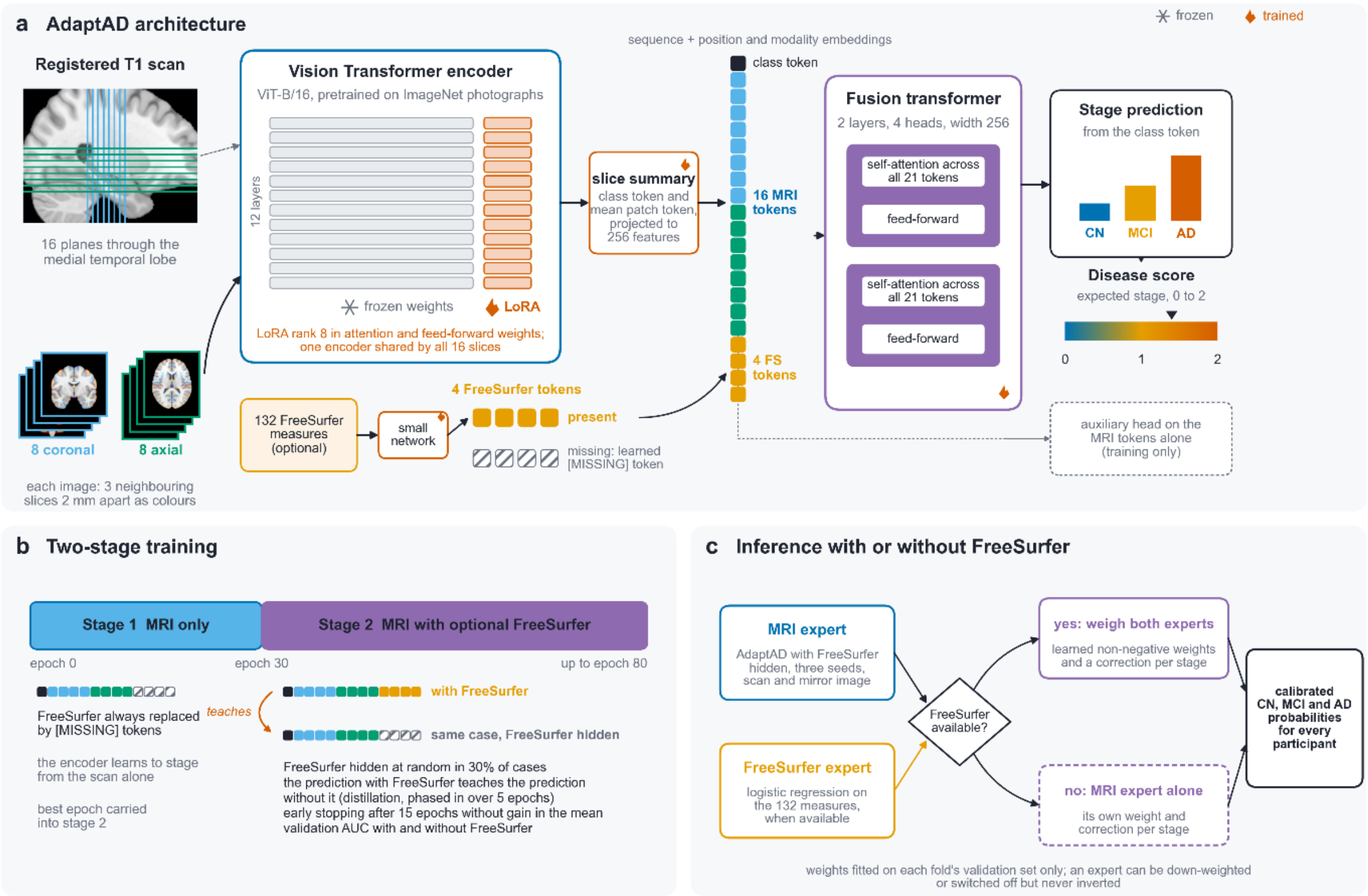

**Fig. 1.** Overview of AdaptAD. (a) Architecture. Sixteen planes through the medial temporal lobe (eight coronal, eight axial) are read by a frozen ViT-B/16 with rank-8 LoRA adapters, giving 16 MRI tokens; 132 FreeSurfer measures give four tokens, or a learned [MISSING] token when unavailable. A two-layer fusion transformer predicts CN, MCI and AD and the disease score; the auxiliary head is used in training only. (b) Two-stage training, with an MRI-only first stage and FreeSurfer dropout and distillation in the second. (c) Missing-aware late fusion of the MRI and FreeSurfer experts at inference. Images show the MNI152 template.

### 3.2 Anatomically placed 2.5-D slice tokens

Each registered volume is cropped to 153 × 189 × 159 voxels at 1 mm isotropic resolution, and its intensities are standardised inside the brain mask $\mathscr{B}$ so that differences in global intensity scale between scanners and participants are removed while contrast between tissues is preserved:

$$\tilde{V}(\mathbf{x}) = \frac{V(\mathbf{x}) - \mu_{\mathcal{B}}}{\sigma_{\mathcal{B}}}, \qquad \mu_{\mathcal{B}} = \frac{1}{|\mathcal{B}|}\sum_{\mathbf{x}\in\mathcal{B}} V(\mathbf{x}), \quad \sigma_{\mathcal{B}}^2 = \frac{1}{|\mathcal{B}|}\sum_{\mathbf{x}\in\mathcal{B}} (V(\mathbf{x}) - \mu_{\mathcal{B}})^2 \tag{3}$$

From the normalised volume, sixteen planes are extracted at fixed MNI coordinates chosen to cross the medial temporal lobe and its surroundings, the region in which neurodegeneration in AD begins [3]. Eight coronal and eight axial planes are placed at

$$y_k = -40 + 5\,(k-1)\ \text{mm}, \qquad z_k = -26 + \frac{40}{7}\,(k-1)\ \text{mm}, \qquad k = 1, \dots, 8 \tag{4}$$

The coronal planes, from y = −40 to −5 mm, cut through the tail, body and head of the hippocampus and through the amygdala. The axial planes, from z = −26 to 14 mm, cut through the hippocampus, the temporal horns and the bodies of the lateral ventricles (Fig. 1a). Because the planes are fixed in template space, token k shows the same anatomy in every participant, and its position in the input sequence carries anatomical meaning that the fusion transformer can learn to exploit. This contrasts with slice-based methods that select slices per participant by hand or by an image statistic, where the same token position may show different structures in different people. A 2-D network pretrained on natural images expects three input channels. Instead of replicating a single slice, AdaptAD fills the channels with the slice at the plane and its neighbours 2 mm on either side, so that every image carries local through-plane context. Writing $S_\pi(u)$ for the slice of $\tilde{V}$ perpendicular to axis π at coordinate u, the k-th image is

$$\mathbf{I}_k = \left[\, S_{\pi_k}(u_k - 2),\ S_{\pi_k}(u_k),\ S_{\pi_k}(u_k + 2) \,\right], \qquad k = 1, \dots, 16 \tag{5}$$

where $\pi_k$ is the coronal axis for k ≤ 8 and the axial axis otherwise, and $u_k$ is the corresponding coordinate from Eq. (4). In this way the colour channels of the pretrained network encode a small 3-D neighbourhood rather than redundant copies. Each image is oriented with superior or anterior up, padded to a square and resized to 224 × 224 pixels. Its intensities are then mapped from the brain z-score window [−2.5, 2.0] to [0, 1] and normalised per channel with the ImageNet statistics (μ_c, σ_c), which places the input in the range the backbone saw during pretraining:

$$\hat{\mathbf{I}}_k^{(c)} = \frac{\operatorname{clip}\left(\frac{\mathbf{I}_k^{(c)} + 2.5}{4.5},\ 0,\ 1\right) - \mu_c}{\sigma_c}, \qquad c \in \{1,2,3\} \tag{6}$$

### 3.3 ViT-B/16 encoder with low-rank adaptation

The encoder is a ViT-B/16 [8] with 224 × 224 input, pretrained on ImageNet-21k [39] with the augmentation and regularisation recipe of [9] and then fine-tuned on ImageNet-1k, as released in the PyTorch Image Models library. The backbone splits each 224 × 224 image into 196 patches of 16 × 16 pixels, embeds them in d = 768 dimensions, prepends a [CLS] token and processes the sequence with 12 transformer blocks. All 86 M pretrained weights remain frozen throughout training, which preserves the general visual features learned from natural images and avoids the overfitting that full fine-tuning risks on a cohort of a few hundred participants.

To adapt the frozen backbone to brain MRI, we use low-rank adaptation [13]. LoRA rests on the observation that the change a pretrained weight matrix needs for a new task has low intrinsic rank. For a frozen matrix $W_0 \in \mathbb{R}^{\{d_{in} \times d_{out}\}}$, it learns two small matrices and modifies the layer's output to

$$\mathbf{h} = W_0\,\mathbf{x} + \frac{\alpha}{r}\,BA\,\mathbf{x}, \qquad A \in \mathbb{R}^{r \times d_{in}},\ B \in \mathbb{R}^{d_{out} \times r},\ r \ll \min(d_{in}, d_{out}) \tag{7}$$

with rank r = 8 and scaling α = 16. A is initialised randomly and B at zero, so the product BA vanishes at the start of training, and the network begins exactly at its pretrained function. The adaptation then grows only as far as the data require, and the factor α/r keeps the magnitude of the update approximately independent of the rank. We apply Eq. (7) to four projections in each of the 12 blocks: the joint query-key-value projection (768 ⟶ 2304), the attention output projection (768 ⟶ 768) and both feed-forward projections (768 ⟶ 3072 and 3072 ⟶ 768). Adapting the feed-forward layers in addition to attention lets the update change both where the model attends and how it transforms features. The number of LoRA parameters is

$$|\theta_{\mathrm{LoRA}}| = L\,r \sum_{\ell \in \mathcal{P}} \left(d_{in}^{(\ell)} + d_{out}^{(\ell)}\right) = 12 \times 8 \times (3072 + 1536 + 3840 + 3840) = 1{,}179{,}648 \tag{8}$$

where L = 12 is the number of blocks and $\mathcal{P}$ the set of adapted projections. This is about 1.3% of the backbone. After training, the product BA can be merged into $W_0$, so the adapted model has the same size and inference cost as the original.

The 16 images of a participant pass through the adapted backbone independently. For image k, let $\mathbf{c}_k \in \mathbb{R}^{768}$ *be the output embedding of the [CLS] token and* $\mathbf{p}_{k,j}$ the output embedding of patch j. The [CLS] embedding summarises the image globally, whereas the mean of the patch embeddings retains spatially distributed evidence, such as diffuse widening of sulci, that the [CLS] token may underweight. Their concatenation is projected to an MRI token of width D = 256,

$$\mathbf{m}_k = g_{\text{MRI}}\left(\text{LN}\left[\mathbf{c}_k\,;\,\frac{1}{196}\sum_{j=1}^{196}\mathbf{p}_{k,j}\right]\right) \in \mathbb{R}^D, \qquad k = 1,\ldots,16 \tag{9}$$

where [·; · denotes concatenation (1,536 values), LN is layer normalisation and g_MRI is a two-layer perceptron (1,536 → 256 → 256, GELU). One projection is shared by all 16 images, and the anatomical identity of each token is supplied later by its position embedding. Together with the FreeSurfer encoder, the fusion transformer and the prediction heads described below, AdaptAD trains 3.53 M of its 89.3 M parameters (3.9%), of which LoRA contributes 1.18 M; a complete breakdown for AdaptAD and every comparison method is given in Table S7.

### 3.4 FreeSurfer tokens and the [MISSING] token

The 132 FreeSurfer features comprise cortical and subcortical volumes, divided by intracranial volume and multiplied by 1,000 to remove the effect of head size, and regional averages of cortical thickness. Individual missing values within an available vector are imputed with the median, and each feature is standardised with statistics computed on the training participants of the current fold only, so that no information from validation or test participants enters the model:

$$\tilde{\mathrm{f}}_i = \frac{\text{impute}(\mathrm{f}_i) - \mu_{\text{tr}}}{\sigma_{\text{tr}}} \tag{10}$$

A two-layer perceptron $g_{\text{FS}}$ (132 → 256 → 4 × 256, GELU, dropout 0.2) then maps the standardised vector to four tokens of the same width as the MRI tokens. Representing morphometry by several tokens rather than one lets the fusion transformer attend to different aspects of it, for example medial temporal volumes and cortical thickness, through different attention patterns:

$$[\mathrm{t}_{i,1},\ldots,\mathrm{t}_{i,4}] = \text{reshape}\left(g_{\text{FS}}(\tilde{\mathrm{f}}_i)\right), \qquad \mathrm{t}_{i,j} \in \mathbb{R}^D \tag{11}$$

When morphometry is absent, or deliberately hidden during training, the four tokens are replaced by four copies of a single learned vector $\mathrm{e} \in \mathbb{R}^D$, the [MISSING] token. With an effective availability flag $\tilde{m}_i$, the tokens presented to the network are

$$\mathrm{t}'_{i,j} = \tilde{m}_i\,\mathrm{t}_{i,j} + (1 - \tilde{m}_i)\,\mathrm{e}, \qquad j = 1,\ldots,4 \tag{12}$$

where the effective flag depends on the training stage and on the prediction being computed:

$$\tilde{m}_i = \begin{cases} 0 & \text{in stage 1, and for the MRI-only prediction in every stage,} \\ m_i\,(1-\delta_i),\;\; \delta_i \sim \text{Bernoulli}(0.3) & \text{for the main prediction in stage 2,} \\ m_i & \text{for the distillation teacher and at inference.} \end{cases} \tag{13}$$

This design keeps the length and structure of the input sequence identical whether or not morphometry exists, and it lets the network learn an explicit representation of absent morphometry. Zero-filling, the most common alternative, cannot distinguish a missing value from a measurement that happens to equal the training mean, and so forces the network to read the absence of data as data.

### 3.5 Fusion transformer and prediction heads

The fusion transformer receives a learned [CLS] token $\mathrm{x}_{\text{cls}}$, the 16 MRI tokens and the 4 FreeSurfer or [MISSING] tokens, 21 tokens in total. Each token is given a learned position embedding $\mathrm{P}_n$, which identifies the anatomical plane or morphometry slot it comes from, and a learned modality embedding E, which identifies it as imaging or morphometry:

$$\mathrm{X}^{(0)} = [\,\mathrm{x}_{\text{cls}};\; \mathrm{m}_1 + \mathrm{P}_1 + \mathrm{E}_{\text{MRI}};\; \ldots;\; \mathrm{m}_{16} + \mathrm{P}_{16} + \mathrm{E}_{\text{MRI}};\; \mathrm{t}'_1 + \mathrm{P}_{17} + \mathrm{E}_{\text{FS}};\; \ldots;\; \mathrm{t}'_4 + \mathrm{P}_{20} + \mathrm{E}_{\text{FS}}\,] \in \mathbb{R}^{21\times D} \tag{14}$$

Two pre-norm transformer layers with width D = 256, 4 attention heads, feed-forward width 1,024 and dropout 0.1 then mix information across anatomical planes and modalities:

$$\mathrm{X}' = \mathrm{X}^{(l-1)} + \text{MHSA}\left(\text{LN}\left(\mathrm{X}^{(l-1)}\right)\right), \qquad \mathrm{X}^{(l)} = \mathrm{X}' + \text{FFN}(\text{LN}(\mathrm{X}')), \qquad l = 1{,}2 \tag{15}$$

where each of the four heads of the multi-head self-attention computes

$$\mathrm{Attn}_h(\mathrm{X}) = \mathrm{softmax}\left(\frac{(\mathrm{X}W_h^Q)(\mathrm{X}W_h^K)^\top}{\sqrt{D/4}}\right)\mathrm{X}W_h^V \tag{16}$$

Through attention, each MRI token can draw on evidence from the other planes, for instance relating hippocampal atrophy in a coronal plane to ventricular enlargement in an axial one, and, when it is available, on morphometry. The [CLS] token gathers the result. A linear head on its final embedding gives the class logits, and the same network yields three predictions that differ only in what it is shown:

$$\mathbf{z}^{\mathrm{full}} = W_o\,\mathbf{x}_{\mathrm{cls}}^{(2)}\big|\tilde{m}, \qquad \mathbf{z}^{\mathrm{MRI}} = W_o\,\mathbf{x}_{\mathrm{cls}}^{(2)}\big|_{\tilde{m}=0}, \qquad \mathbf{z}^{\mathrm{aux}} = W_a\,\frac{1}{16}\sum_{k=1}^{16}\mathbf{m}_k \tag{17}$$

with biases omitted for brevity. The full prediction uses morphometry where it is available, and the MRI-only prediction replaces it with the [MISSING] token for every participant. The auxiliary prediction is a linear classifier on the mean MRI token that bypasses the fusion transformer. It supervises the image pathway directly, in the manner of deep supervision, and serves as a candidate MRI expert at inference.

### 3.6 Two-stage training

All classification terms use a class-balanced cross-entropy with label smoothing. For a participant with logits z and label y, with C = 3 classes and smoothing ε = 0.1, the per-participant loss is

$$\ell(\mathbf{z}, y) = -\sum_{c=1}^{C} w_c\; q_c \log\sigma\,(\mathbf{z})_c, \qquad q_c = (1-\varepsilon)\,\mathbb{1}[c = y] + \frac{\varepsilon}{C}, \qquad w_c = \frac{N_{\mathrm{tr}}}{C\,N_c} \tag{18}$$

where σ is the softmax function, $N_{\mathrm{tr}}$ is the number of training participants in the fold and $N_c$ the number of them in class c. Over a mini-batch $\mathscr{B}$ the loss is normalised by the total weight of the true classes, $CE = \frac{\sum_{i\in\mathcal{B}}\ell(\mathbf{z}_i,y_i)}{\sum_{i\in\mathcal{B}} w_{y_i}}$. The class weights are computed once per fold from its training participants, and they counter the imbalance of the cohort, in which CN participants outnumber AD participants by about 2.6 to 1. Label smoothing discourages over-confident predictions, which suits a task in which the boundaries between neighbouring stages are themselves clinically uncertain.

Training proceeds in two stages. In the first stage, which covers epochs 0 to 29, morphometry is hidden for every participant ($\tilde{m}$ = 0 in Eq. 13). The main prediction therefore coincides with the MRI-only prediction, and the loss combines it with the auxiliary head:

$$\mathcal{L}_1 = \mathrm{CE}(\mathbf{z}^{\mathrm{MRI}}, y) + \mathrm{CE}(\mathbf{z}^{\mathrm{aux}}, y) \tag{19}$$

This warm-up forces the LoRA updates, the token projection and the fusion transformer to extract as much as possible from the image before any morphometry is shown. Without it, the network could come to depend on the tabular features, which are lower-dimensional and easier to exploit, and neglect the image pathway that every participant has. During this stage the validation MRI-only macro-AUC is recorded after every epoch. The checkpoint with the highest trailing three-epoch mean of that AUC is kept, and its weights initialise the second stage. In the second stage, from epoch 30 up to epoch 80, morphometry is shown wherever it exists but hidden at random for 30% of the participants in each batch, following Eq. (13). A participant without morphometry is never shown it, whatever the random draw. Two further terms keep the MRI-only pathway strong while the network learns to use morphometry. The first is a direct loss on the MRI-only prediction. The second is knowledge distillation [28] from a teacher, which is the same network evaluated on the same encoded tokens with all available morphometry and without gradients. The teacher is applied only to the participants of the batch who have morphometry, $\mathcal{M}_{\mathcal{B}} = \{i \in \mathscr{B} : m_i = 1\}$:

$$\mathcal{L}_{\mathrm{KD}} = \frac{T^2}{|\mathcal{M}_{\mathcal{B}}|}\sum_{i\in\mathcal{M}_{\mathcal{B}}} \mathrm{KL}\left(\sigma\left(\mathrm{sg}\left[\mathbf{z}_i^{\mathrm{teach}}\right]/T\right) \,\|\, \sigma\left(\mathbf{z}_i^{\mathrm{MRI}}/T\right)\right) \tag{20}$$

Here sg [·] stops the gradient, T = 2 is the distillation temperature, and the factor $T^2$ keeps gradient magnitudes comparable across temperatures [28]. The teacher's softened distribution encodes how morphometry shifts the relative probabilities of neighbouring stages, and distillation transfers that information to the prediction made without morphometry. The complete stage-2 objective is

$$\mathcal{L}_2 = \mathrm{CE}(\mathbf{z}^{\mathrm{full}}, y) + 0.5\ \mathrm{CE}(\mathbf{z}^{\mathrm{MRI}}, y) + \mathrm{CE}(\mathbf{z}^{\mathrm{aux}}, y) + \lambda_t \mathcal{L}_{\mathrm{KD}}, \qquad \lambda_t = \min\left(1, \frac{t-29}{5}\right), \quad t \geq 30 \tag{21}$$

where t is the epoch. The distillation weight $\lambda_t$ rises linearly from 0.2 in the first epoch of the stage to 1 in the fifth and stays there, so that the student is not pulled strongly towards a teacher that has only just begun to see morphometry. Parameters are optimised with AdamW in separate groups. The LoRA matrices use a peak learning rate of $2 \times 10^{-4}$ and weight decay $10^{-4}$. The remaining trainable weights, namely the token projection, FreeSurfer encoder, fusion transformer and heads, use a peak learning rate of $3 \times 10^{-4}$ and weight decay 0.05. Biases, normalisation parameters and the learned embeddings (the [CLS], position, modality and [MISSING] tokens) are excluded from weight decay. The smaller rate and weaker decay for LoRA reflect that its updates act directly on pretrained features. The learning rate of every group is updated at each optimiser step $\tau$. It rises linearly during the steps of the first five epochs, $\tau < \tau_w$, and then decays along a cosine to 1% of its peak $\eta_{\max}$ by the last step of epoch 80, $\tau_{\max}$:

$$\eta_\tau = \begin{cases} \eta_{\max} \dfrac{\tau+1}{\tau_w}, & \tau < \tau_w, \\ \eta_{\min} + \dfrac{1}{2}(\eta_{\max} - \eta_{\min})\left(1 + \cos\dfrac{\pi(\tau - \tau_w)}{\tau_{\max} - \tau_w}\right), & \tau \geq \tau_w, \end{cases} \qquad \eta_{\min} = 0.01\,\eta_{\max} \tag{22}$$

The schedule runs continuously across both stages, so stage 2 begins at a learning rate already past its peak, which suits a phase that refines rather than relearns the image pathway. Each optimiser step accumulates gradients over two mini-batches of four participants, an effective batch of 8 participants or 128 slice images. Gradients are clipped at a global norm of 1.0, training uses mixed precision, and stochastic depth with rate 0.1 regularises the backbone. Each training participant is augmented by a random left-right flip of the volume with probability 0.5, a random jitter of ±2 voxels of every plane position in Eq. (4), an independent mild affine transform of each image (rotation ±8°, scale 0.92 to 1.08, translation ±4%) and one random change of brightness and contrast of ±10%. The plane jitter and per-image affine transforms expose the model to the small residual misregistrations that occur in practice, so that it does not rely on exact voxel positions.

After every stage-2 epoch e, the network is evaluated on the fold's validation set. The selection score is the mean of the three-class macro-AUCs of the full and MRI-only predictions, with a very small penalty on their mean validation cross-entropy $\ell$ that breaks ties between epochs of equal AUC. The score is then smoothed by a trailing moving average:

$$J_e = \frac{1}{2}\left[\mathrm{AUC}(\mathbf{z}^{\mathrm{full}}) + \mathrm{AUC}(\mathbf{z}^{\mathrm{MRI}})\right]_e - 10^{-3} \cdot \frac{1}{2}\left[\ell^{\mathrm{full}} + \ell^{\mathrm{MRI}}\right]_e, \qquad \bar{J}_e = \frac{1}{|\mathcal{W}_e|} \sum_{e' \in \mathcal{W}_e} J_{e'} \tag{23}$$

where $\mathcal{W}_e$ contains epoch e and up to two preceding stage-2 epochs. Only stage-2 checkpoints are eligible. The checkpoint that maximises $\bar{J}_e$ is kept, and training stops once 15 consecutive stage-2 epochs pass without improvement. Averaging the two AUCs selects a model that performs well both with and without morphometry, and smoothing prevents the selection of a single fortunate epoch on a validation set of about 72 participants.

### 3.7 Missing-aware late fusion and calibration

At inference, every prediction is averaged over the original and the left-right flipped volume and over three models trained with seeds $\mathcal{S} = \{42, 142, 242\}$. Flip averaging exploits the approximate bilateral symmetry of AD-related atrophy, and seed averaging reduces the variance due to random initialisation and data order. The MRI expert is

$$\bar{\mathbf{p}}_i^{\mathrm{MRI}} = \frac{1}{2|\mathcal{S}|} \sum_{s \in \mathcal{S}} \sum_{\phi \in \{\mathrm{id}, \mathrm{flip}\}} \sigma\left(\mathbf{z}_s^{\mathrm{MRI}}(\phi(V_i))\right) \tag{24}$$

The MRI-only and auxiliary predictions of Eq. (17) were the two candidates for this expert, and the choice between them was made automatically on pooled validation AUC, never on test data. The second expert uses morphometry alone. It is a class-balanced multinomial logistic regression on the standardised features of Eq. (10),

$$\mathbf{p}_i^{\mathrm{FS}} = \sigma\left(W_{\mathrm{FS}}\, \tilde{\mathbf{f}}_i + \mathbf{b}_{\mathrm{FS}}\right) \tag{25}$$

fitted on the fold's training participants who have morphometry, with the L2 regularisation strength chosen by inner three-fold cross-validation. Following the principle of stacked generalisation [31], the two experts are combined by a second-level model fitted on each fold's validation participants only. The stacker has one branch for each pattern of availability:

$$\mathbf{z}_i^{\text{stack}} = \begin{cases} a \log \overline{\mathbf{p}}_i^{\text{MRI}} + b \log \mathbf{p}_i^{\text{FS}} + \mathbf{c}, & m_i = 1, \\ a' \log \overline{\mathbf{p}}_i^{\text{MRI}} + \mathbf{c}', & m_i = 0, \end{cases} \qquad \hat{\mathbf{p}}_i = \sigma\left(\mathbf{z}_i^{\text{stack}}\right) \tag{26}$$

The two branches are fitted separately. The first uses the validation participants who have morphometry, $\mathcal{V}_1$, and the second uses all validation participants, $\mathcal{V}$. In each case the class-balanced negative log-likelihood is minimised under box constraints, with a small penalty on the class biases:

$$\min_{a,b,\mathbf{c}} \; -\frac{\sum_{i\in\mathcal{V}_1} w_{y_i} \log \hat{p}_{i,y_i}}{\sum_{i\in\mathcal{V}_1} w_{y_i}} + 0.01\,\|\mathbf{c}\|_2^2, \qquad \min_{a',\mathbf{c}'} \; -\frac{\sum_{i\in\mathcal{V}} w_{y_i} \log \hat{p}_{i,y_i}}{\sum_{i\in\mathcal{V}} w_{y_i}} + 0.01\,\|\mathbf{c}'\|_2^2 \tag{27}$$

subject to a, b, a′ ∈ [0, 3] and **c**, **c**′ ∈ [−5, 5] [3], where $w_c$ is the weight of class c from Eq. (18), computed on the set being fitted, so that $w_{y_i}$ is the weight of the true class of participant i. The constraints allow an expert to be down-weighted, up-weighted or switched off with a weight of zero, but never inverted, and the penalty keeps the class biases moderate when the validation set is small. Both problems are solved with L-BFGS-B. Two properties of this construction matter for deployment. First, no prediction ever depends on morphometry being present, because a participant without it is scored by the second branch of Eq. (26) alone. Second, that branch is exactly a temperature (1/a′) and class-bias calibration of the MRI expert in the sense of [35]. The same per-fold temperature and bias calibration is applied to every deep comparison method (Section 4.6), so that all methods are compared on an equal footing. We report two versions of the model. AdaptAD, the headline model, applies the second branch to every participant and therefore needs nothing but a T1 scan. AdaptAD late fusion applies the full stacker and adds morphometry when it is available. Reporting both separates what the image achieves on its own from what morphometry adds.

## 4. Experimental setup

The experiments test whether AdaptAD stages the AD continuum more accurately than established alternatives, whether it does so without morphometry, and whether its performance carries over to an independent cohort. All models were developed by five-fold cross-validation on the ADNI cohort, locked, and then tested once on the OASIS-3 cohort with endpoints fixed beforehand. Fig. 2 shows how participants were selected in both cohorts. Table 1 summarises their demographic and clinical characteristics (panel A) and the partition of ADNI into training, validation and test sets (panel B). Section 4.1 describes the cohorts, Section 4.2 the preprocessing, Section 4.3 the partitioning, Section 4.4 the pre-registered decision gates, Section 4.5 the comparison methods, Section 4.6 the evaluation metrics and statistical analysis, and Section 4.7 the implementation.

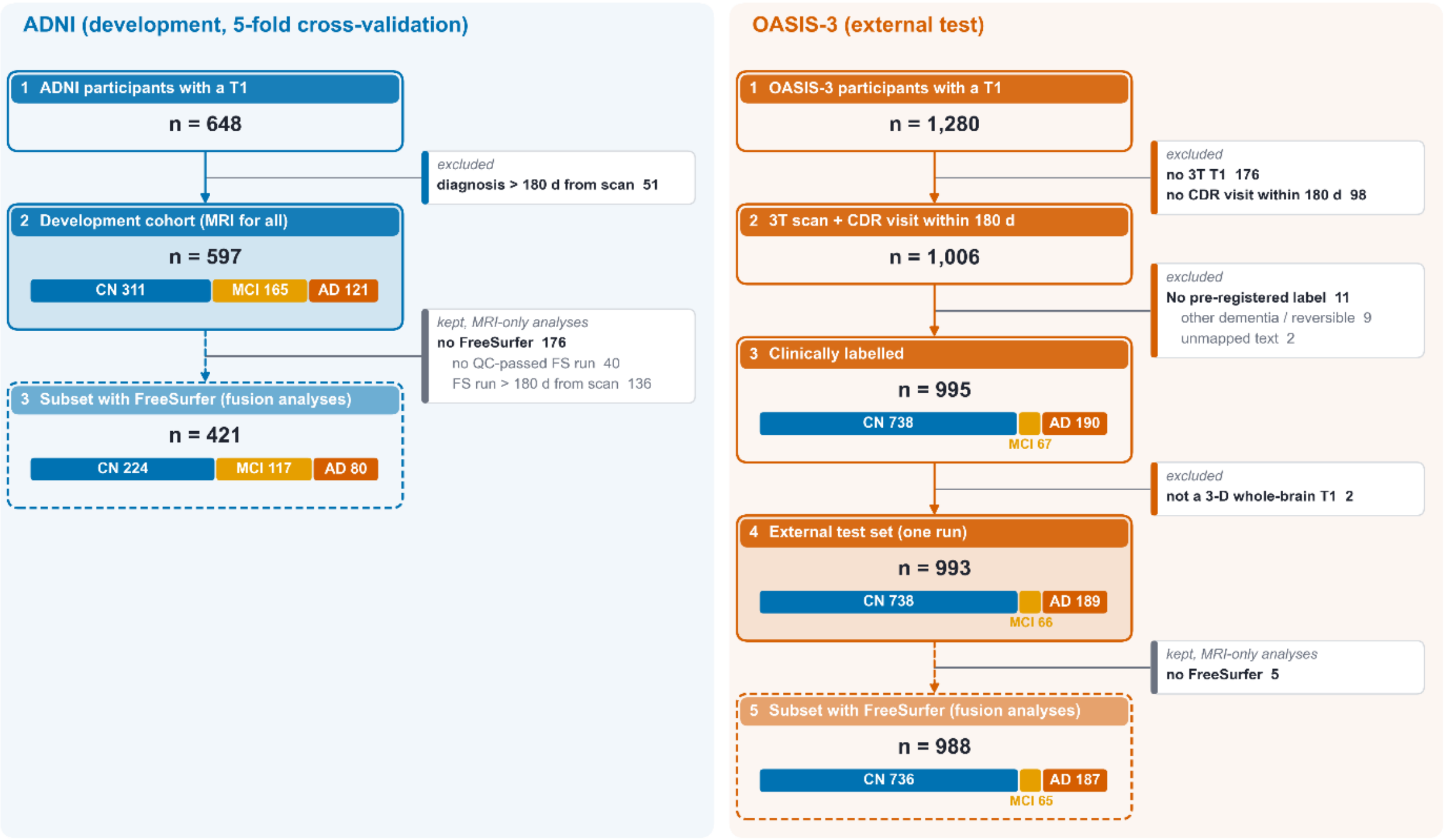

**Fig. 2.** Participant flow for the ADNI development cohort (left) and the OASIS-3 external test cohort (right). Each step shows the number of participants and its composition by diagnosis; side boxes give the exclusions and the participants without FreeSurfer morphometry, who were kept for the MRI-only analyses. Dashed boxes mark the subsets with FreeSurfer used in the fusion analyses.

### 4.1 Datasets

The development cohort was drawn from the Alzheimer's Disease Neuroimaging Initiative (ADNI) [40], a multi-site longitudinal study that acquires structural MRI under a harmonised protocol [41]. Of 648 participants with a usable T1-weighted scan, 597 had a diagnosis recorded in the ADNI diagnostic summary within 180 days of that scan, and these form the cohort: 311 CN, 165 MCI and 121 AD, with one scan per participant (Fig. 2, left; Table 1A). The label is the diagnosis at the date of the scan rather than at study entry, so that the image and its label describe the same point on the continuum. All scans were acquired at 3T between 2017 and 2025, most of them at follow-up rather than baseline visits (460 of the 556 participants with an available visit code, at a median of 24 months after study entry), and the diagnosis was recorded a median of 9 days from the scan (within 30 days for 457 of the 597 participants). Scans were acquired on Siemens (437), Philips (92) and GE (68) systems. FreeSurfer morphometry from the UCSF

**Table 1.** Characteristics of the development (ADNI) and external test (OASIS-3) cohorts.

**(A)** Demographic and clinical characteristics of the ADNI development cohort (n = 597) and the OASIS-3 external test cohort (n = 993) by diagnosis (OASIS-3, primary clinical label). Values are mean ± SD or n (%); p compares CN, MCI and AD within each cohort (Kruskal-Wallis for continuous and chi-square for categorical variables). Where a value is missing for some participants, the number with data is given; nothing was imputed. FreeSurfer available: UCSF FreeSurfer (ADNI) or OASIS FreeSurfer, quality control passed, within 180 days of the scan.

| Characteristic | ADNI CN | ADNI MCI | ADNI AD | ADNI All | p | OASIS-3 CN | OASIS-3 MCI | OASIS-3 AD | OASIS-3 All | p |
|---|---|---|---|---|---|---|---|---|---|---|
| n | 311 | 165 | 121 | 597 | | 738 | 66 | 189 | 993 | |
| Age, years | 74.6 ± 7.5 | 76.5 ± 7.8 | 75.1 ± 8.7 | 75.3 ± 7.9 | 0.035 | 69.5 ± 8.5 | 74.4 ± 5.9 | 75.0 ± 7.0 | 70.9 ± 8.4 | <0.001 |
| Female, n (%) | 185 (59.5) | 68 (41.2) | 59 (48.8) | 312 (52.3) | <0.001 | 421 (57.0) | 33 (50.0) | 87 (46.0) | 541 (54.5) | 0.019 |
| Education, years | 16.8 ± 2.3 | 16.0 ± 2.7 | 15.6 ± 2.6 | 16.3 ± 2.5 | <0.001 | 16.1 ± 2.5 | 15.7 ± 2.3 | 14.8 ± 3.0 | 15.8 ± 2.7 | <0.001 |
| APOE ε4 carrier, n (%) | 105 (34.0) | 55 (33.5) | 58 (64.4) | 218 (38.7) | <0.001 | 261 (35.6) | 28 (42.4) | 125 (67.2) | 414 (42.0) | <0.001 |
| MMSE | 29.1 ± 1.1 (n = 308) | 27.4 ± 2.4 (n = 162) | 21.3 ± 5.6 (n = 118) | 27.1 ± 4.2 (n = 588) | <0.001 | 29.0 ± 1.2 | 27.7 ± 2.1 | 24.9 ± 3.5 | 28.2 ± 2.5 | <0.001 |
| CDR-SB | 0.1 ± 0.2 (n = 296) | 1.5 ± 1.2 (n = 158) | 5.7 ± 3.4 (n = 119) | 1.6 ± 2.7 (n = 573) | <0.001 | 0.0 ± 0.1 | 1.2 ± 0.7 | 3.5 ± 1.8 | 0.8 ± 1.6 | <0.001 |
| FreeSurfer available, n (%) | 224 (72.0) | 117 (70.9) | 80 (66.1) | 421 (70.5) | 0.477 | 736 (99.7) | 65 (98.5) | 187 (98.9) | 988 (99.5) | 0.192 |

*Values are mean ± SD or n (%). p: Kruskal-Wallis test (continuous) or chi-square test (categorical) across CN, MCI and AD within each cohort. Values missing for some participants are shown with their n; nothing was imputed. FreeSurfer available: UCSF FreeSurfer (ADNI) or OASIS FreeSurfer, quality-control passed, within 180 days of the scan.*

**(B)** Partition of the ADNI cohort into training, validation and test sets in each of the five cross-validation folds. Folds are formed at the participant level and stratified by diagnosis and FreeSurfer availability, and the validation set is 15% of the training part. The five test sets are disjoint, cover all 597 participants exactly once and are identical for every method. OASIS-3 is used only for testing, with no training, tuning or recalibration.

| Fold | Set | n | CN | MCI | AD | With FreeSurfer |
|---|---|---|---|---|---|---|
| 0 | Training | 405 | 212 | 111 | 82 | 285 |
| 0 | Validation | 72 | 37 | 20 | 15 | 51 |
| 0 | Test | 120 | 62 | 34 | 24 | 85 |
| 1 | Training | 405 | 211 | 112 | 82 | 286 |
| 1 | Validation | 72 | 37 | 20 | 15 | 51 |
| 1 | Test | 120 | 63 | 33 | 24 | 84 |
| 2 | Training | 406 | 211 | 113 | 82 | 286 |
| 2 | Validation | 72 | 37 | 20 | 15 | 51 |
| 2 | Test | 119 | 63 | 32 | 24 | 84 |
| 3 | Training | 406 | 212 | 113 | 81 | 286 |
| 3 | Validation | 72 | 37 | 20 | 15 | 51 |
| 3 | Test | 119 | 62 | 32 | 25 | 84 |
| 4 | Training | 406 | 213 | 111 | 82 | 286 |
| 4 | Validation | 72 | 37 | 20 | 15 | 51 |
| 4 | Test | 119 | 61 | 34 | 24 | 84 |

*Participant-level folds stratified by diagnosis × FreeSurfer availability; validation = 15% of the training part, stratified in the same way. The five test sets are disjoint and cover all 597 participants exactly once; the partition is identical for every method and ablation. OASIS-3 is used only for testing (no training, tuning or recalibration).*

pipeline (FreeSurfer 6, overall quality control rated pass or partial) acquired within 180 days of the scan was available for 421 participants (70.5%). The other 176 were not excluded: 40 had no FreeSurfer run that passed quality control, and 136 had none close enough in time to the scan. This naturally occurring missingness is the setting for which AdaptAD was designed, and all ADNI comparisons in Section 5 are made under it. The external cohort was drawn from OASIS-3 [42], a longitudinal study of normal ageing and AD conducted at a different institution, with different scanners, recruitment and diagnostic practice. The cohort and its labels were defined before any OASIS-3 prediction was made (Fig. 2, right). Of 1,280 participants with a T1-weighted scan, we kept, for each participant, the earliest 3T session with a Clinical Dementia Rating (CDR) [43] assessment within 180 days, which yielded 1,006 participants. The primary label combines the clinician's diagnosis with the CDR. CN requires a diagnosis of cognitively normal and CDR 0. MCI requires CDR 0.5 with a diagnosis of impairment without dementia. AD requires CDR of at least 0.5 with AD dementia named as the primary diagnosis. Eleven participants whose diagnosis named another dementia or a reversible cause, or could not be mapped, received no label, and two scans that were not 3-D whole-brain acquisitions were excluded. This left 993 participants (738 CN, 66 MCI and 189 AD) scanned on Siemens TrioTim, Biograph mMR and MAGNETOM Vida systems. Two secondary labels were pre-specified because they do not depend on the clinician's free-text diagnosis: a three-class label based on the CDR alone (CDR 0, 0.5 and at least 1) and a binary label of CN versus impaired (CDR at least 0.5). Under the CDR-based label, 126 of the 189 participants with a clinical diagnosis of AD fall into the MCI class, because their global CDR was 0.5. The complete label mapping is given in Table S1. The Mini-Mental State Examination (MMSE) [44] and the CDR Sum of Boxes (CDR-SB) at the matched visit were extracted for the clinical analyses, and FreeSurfer output was available for 988 of the 993 participants. The two cohorts differ in ways that matter for staging (Table 1A). AD participants in OASIS-3 are milder than in ADNI (CDR-SB 3.5 ± 1.8 versus 5.7 ± 3.4; MMSE 24.9 versus 21.3), the MCI group is small (66 participants, 6.6%), and the prevalence of impairment is 26% rather than 48%. The external test therefore probes generalisation across a shift in site, scanner, label definition, disease severity and class prevalence at the same time.

### 4.2 Data preprocessing

Both cohorts were processed by the same code, in the same fixed sequence of steps. Each T1-weighted image was converted to NIfTI and reoriented to a standard orientation. The brain was then extracted with HD-BET with test-time augmentation [45], which is robust to differences in contrast and pathology across scanners. Intensity inhomogeneity was corrected with N4 bias-field correction [46]. Each bias-corrected brain was registered to the 1 mm MNI152 template by symmetric diffeomorphic registration in ANTs (antsRegistrationSyNQuick, with a fixed random seed) [47]. Finally, the registered volume was cropped to 153 × 189 × 159 voxels and z-scored inside the brain mask (Eq. 3). The volumes were cached once. The 16 slice tokens of Section 3.2 were extracted from this cache on the fly during training and inference, so that plane jitter could be applied as augmentation. The 3-D comparison methods (Section 4.5) read the same cached volumes, so every method saw identically preprocessed data. Two checks protected the integrity of the pipeline. First, before OASIS-3 was processed, the pipeline was re-run on 10 ADNI scans and reproduced their cached volumes bit for bit, confirming that the external cohort would be processed exactly as the development cohort had been. Second, all 1,004 OASIS-3 scans that entered preprocessing were checked automatically and visually. The Dice overlap between each registered brain mask and the template brain mask had a median of 0.993 and a minimum of 0.986, and visual review of the flagged scans and a random sample excluded none (Fig. S1). Six further participants passed registration but did not enter the final cohort because their nearest recorded diagnosis was more than 180 days from the scan (217 to 669 days), and their overlap scores (0.989 to 0.993) were within the range of the cohort.

Morphometry was processed as described in Section 3.4. Regional volumes were divided by intracranial volume and multiplied by 1,000, and cortical thickness was averaged per region, giving 132 features. Imputation and standardisation (Eq. 10) were always fitted on the training participants of the current fold and only applied to its validation and test participants. In OASIS-3, morphometry had been computed with FreeSurfer 5.3 rather than FreeSurfer 6, and the FreeSurfer version is known to change volume and thickness estimates [48]. The OASIS-3 output was therefore mapped to the 132 ADNI features by region name: 128 features were matched, and the 4 without an OASIS-3 equivalent were set to missing (Table S10). The mapping was checked by comparing the medians of cognitively normal participants in the two cohorts. The median ratio of OASIS-3 to ADNI was 0.95 for volumes and 0.97 for thickness, close to unity but not identical. No further harmonisation was applied, so that the external test reflects the morphometry a new site would actually supply.

### 4.3 Training, validation and test partitioning

The ADNI cohort was split at the participant level into five folds, stratified jointly by diagnosis and FreeSurfer availability, so that every fold contains the same mix of stages and of missing morphometry. In each round of cross-validation, one-fold served as the test set, and the remaining four were divided into a training set and a validation set holding 15% of them, again stratified by diagnosis and FreeSurfer availability. As Table 1B shows, every round therefore used 405 or 406 training participants (211 to 213 CN, 111 to 113 MCI and 81 or 82 AD, of whom 285 or 286 had FreeSurfer), 72 validation participants (37 CN, 20 MCI and 15 AD, of whom 51 had FreeSurfer) and 119 or 120 test participants (61 to 63 CN, 32 to 34 MCI and 24 or 25 AD, of whom 84 or 85 had FreeSurfer). The five test sets are disjoint and together cover all 597 participants exactly once, and no participant appears in more than one part of the same round. Each part had one role, fixed in advance. Training participants were used to fit the network weights, the FreeSurfer imputation and standardisation, and the morphometry expert inside the stacker (Eq. 25). Validation participants were used for checkpoint selection (Eq. 23), for the choice of MRI expert, and for fitting the stacker and calibration (Eqs. 26 and 27). The stand-alone FreeSurfer logistic regression, which has no checkpoints to select, was fitted on the training and validation participants together. Test participants were used only to report results. Every reported prediction therefore comes from the one model that never saw that participant, and pooling these out-of-fold predictions gives one prediction for each of the 597 participants. The same five folds, with the same training, validation and test membership, were used for AdaptAD, its two variants, the seven comparison methods and all ablations, so that every difference between methods is measured on the same participants. OASIS-3 was used only for testing, and no OASIS-3 participant contributed to training, model selection, calibration or any decision about the method. Each of the five locked fold models, with the calibration fitted on its own ADNI validation set, predicted every OASIS-3 participant, and the five predictions were averaged. The primary external population is the 988 labelled participants with FreeSurfer, so that every method, including those that require morphometry, is scored on the same participants. AdaptAD, which needs only the T1 scan, was additionally evaluated on all 993 participants as a pre-specified sensitivity analysis.

### 4.4 Pre-registered decision gates and locking

Three gates fixed the method, each using validation data only. Gate 1 fixed the final model as the ensemble of three training seeds (42, 142 and 242). Gate 2 chose the backbone. ViT-B/16 was compared with a DINOv2 ViT-B/14 [10] trained in the same way, under a rule written before the comparison: DINOv2 would be adopted only if its pooled validation MRI-only macro-AUC exceeded that of ViT by at least 0.01. ViT reached 0.781 and DINOv2 0.765, so ViT was retained (Table S1). One protocol amendment, dated 2026-10-10, switched off a training-collapse guard that had wrongly stopped DINOv2 runs whose validation AUC was in fact good. The guard never fired for ViT, so the ViT runs were unaffected. Gate 3 locked the method on 2026-10-12 by recording the SHA-256 hashes of all 71 checkpoints, model files and code files involved in inference (Table S9). The ablations reported in Section 5.2 were run before the lock and, by design, could not change the method. After the lock, the external test was run exactly once, and the test script refuses to run a second time.

### 4.5 Comparison methods

AdaptAD was compared with two of its own variants and seven comparison methods, all trained on the same folds. Table 2 lists each method with its input and number of trainable parameters, and Table S5 gives the full parameter counts. The two variants of AdaptAD isolate parts of its design. A single seed (42) isolates the effect of ensembling, and AdaptAD late fusion (Section 3.7) isolates the effect of adding morphometry. ViT full fine-tuning trains all 88.15

**Table 2.** AdaptAD, two of its variants and seven comparison methods: input, trainable parameters and three-class performance in ADNI cross-validation (complete cases, n = 421) and in the OASIS-3 external test (primary endpoint, clinical label, n = 988). Macro-AUC with 95% CI from 2,000 stratified bootstrap resamples; BAcc, balanced accuracy; QWK, quadratic weighted kappa; ECE, expected calibration error. ΔAUC is the paired difference in macro-AUC (method minus AdaptAD; negative, worse than AdaptAD) with 95% CI, and p is the paired bootstrap p-value with Holm adjustment across comparisons. Input: T1 MRI ± FS, the method uses FreeSurfer features when available and the [MISSING] token or its own missing-data handling otherwise; FS only, FreeSurfer features alone.

| Method | Input | Trainable params | ADNI macro-AUC [95% CI] | ADNI BAc | ADNI QWI | ADNI ECI | ADNI ΔAUC [95% CI] | ADNI p (Holm) | OASIS-3 macro-AUC [95% CI] | OASIS-3 ΔAUC [95% CI] | OASIS-3 p (Holm) |
|---|---|---|---|---|---|---|---|---|---|---|---|
| AdaptAD (ours) | T1 MRI | 3.53 M | 0.745 [0.709, 0.781] | 0.564 | 0.545 | 0.045 | reference | | 0.780 [0.748, 0.814] | reference | |
| AdaptAD, single seed | T1 MRI | 3.53 M | 0.742 [0.707, 0.778] | 0.564 | 0.557 | 0.064 | −0.003 −0.016, +0.010 | 0.644 | 0.784 [0.752, 0.817] | +0.004 [−0.002, +0.010 | 1.000 |
| AdaptAD late fusion (ours, MRI + FS) | T1 MRI ± FS | 3.53 M | 0.732 [0.696, 0.767] | 0.540 | 0.527 | 0.076 | −0.014 [−0.029, +0.002] | 0.243 | 0.778 [0.745, 0.813] | −0.002 [−0.011, +0.007] | 1.000 |
| FreeSurfer logistic regression | FS only | 399 | 0.719 [0.683, 0.753] | 0.548 | 0.510 | 0.054 | −0.026 −0.066, +0.013 | 0.344 | 0.744 [0.712, 0.776] | −0.036 [−0.067, −0.003] | 0.200 |
| MedicalNet (Med3D) | T1 MRI | 32.99 M | 0.710 [0.672, 0.746] | 0.536 | 0.488 | 0.021 | −0.035 −0.069, −0.005 | 0.110 | 0.749 [0.715, 0.781] | −0.031 [−0.060, −0.004 | 0.180 |
| ViT full fine-tuning | T1 MRI | 88.15 M | 0.708 [0.670, 0.745] | 0.506 | 0.531 | 0.068 | −0.037 −0.063, −0.012 | 0.014 | 0.780 [0.748, 0.814] | +0.000 [−0.011, +0.011] | 1.000 |
| FiLM | T1 MRI ± FS | 3.63 M | 0.701 [0.660, 0.738] | 0.510 | 0.446 | 0.075 | −0.045 −0.084, −0.005 | 0.112 | 0.748 [0.715, 0.781] | −0.033 [−0.061, −0.002 | 0.252 |
| DAFT | T1 MRI ± FS | 3.62 M | 0.695 [0.657, 0.733] | 0.538 | 0.453 | 0.062 | −0.051 −0.091, −0.012 | 0.090 | 0.770 [0.738, 0.803] | −0.010 [−0.040, +0.021 | 1.000 |
| ShaSpec | T1 MRI ± FS | 3.95 M | 0.669 [0.628, 0.711] | 0.529 | 0.444 | 0.078 | −0.076 −0.120, −0.035 | 0.008 | 0.742 [0.710, 0.775] | −0.038 [−0.072, −0.000 | 0.300 |
| 3D CNN (Wen et al.) | T1 MRI | 6.35 M | 0.616 [0.574, 0.657] | 0.415 | 0.242 | 0.106 | −0.130 −0.175, −0.084 | <0.001 | 0.753 [0.718, 0.788] | −0.027 [−0.055, +0.000] | 0.300 |

*ADNI: complete cases common to all methods (n = 421). OASIS-3: primary endpoint, clinical label, participants every method could score (n = 988). ΔAUC = method minus AdaptAD, paired stratified bootstrap (2,000 resamples), Holm correction within each cohort; bold p, Holm p < 0.05. AdaptAD is the three-seed ensemble of the locked recipe; all other rows are single models. Deep models are calibrated per fold on validation data only. ± FS: FreeSurfer used when available, MRI-only path otherwise. Rows ordered by ADNI macro-AUC.*

M weights of the same architecture with the same recipe, except that the backbone is trained with a learning rate of $10^{-5}$, and isolates the effect of low-rank adaptation. FreeSurfer logistic regression is a class-balanced multinomial logistic regression on the 132 features alone, with its regularisation strength chosen from nine values between $10^{-4}$ and 1 by inner three-fold cross-validation on the macro-AUC. As a stand-alone comparator it is fitted on the training and validation participants of each fold who have morphometry, and it shows what morphometry achieves without the image. Five 3-D networks read the whole registered volume, resampled to 112 × 136 × 112 voxels (about 1.4 mm). They are the convolutional network of Wen et al. [4], trained from scratch, MedicalNet [5], a ResNet-18 pretrained on 23 medical segmentation datasets, and three methods that fuse image and tabular data on a 3-D ResNet trunk, namely DAFT [17], FiLM conditioning [16] and ShaSpec [18] adapted to image and tabular input. The three fusion methods receive the FreeSurfer vector together with a missing indicator, and the vector is zero-filled when morphometry is absent. The 3-D methods shared one training protocol, so that differences reflect the methods rather than their tuning. Augmentation consisted of a left-right flip, a shift of up to ±6 voxels and intensity jitter, and the loss was class-balanced cross-entropy with label smoothing of 0.1. Optimisation used AdamW with a learning rate of $3 \times 10^{-4}$ when trained from scratch, $10^{-4}$ for pretrained trunks and $10^{-3}$ for heads, a weight decay of $10^{-4}$, a linear warm-up over three epochs and cosine decay, with an effective batch of 8. Training ran for up to 80 epochs, with selection on the trailing three-epoch mean of the validation macro-AUC and a patience of 15 epochs, and flip test-time augmentation was applied at inference. Each comparison method is a single model trained with seed 42. All configurations are listed in Table S6.

### 4.6 Evaluation metrics and statistical analysis

The primary metric is the one-versus-rest macro-averaged area under the receiver operating characteristic curve over the three stages,

$$\mathrm{mAUC} = \frac{1}{3} \sum_{c \in \{\mathrm{CN}, \mathrm{MCI}, \mathrm{AD}\}} \mathrm{AUC}\left(\{p_{i,c}\}_i, \{\mathbb{1}[y_i = c]\}_i\right) \quad (28)$$

which measures how well the predicted probability of each stage separates participants in that stage from all others, independently of any decision threshold and of class prevalence. Four secondary metrics use the predicted stage $\hat{y}_i = \arg\max_c p_{i,c}$. Balanced accuracy and macro-F1 average the recall and the F1 score over the three classes:

$$\mathrm{BAcc} = \frac{1}{3} \sum_c \frac{\mathrm{TP}_c}{\mathrm{TP}_c + \mathrm{FN}_c}, \qquad \mathrm{F1}_{\mathrm{macro}} = \frac{1}{3} \sum_c \frac{2\,\mathrm{TP}_c}{2\,\mathrm{TP}_c + \mathrm{FP}_c + \mathrm{FN}_c} \quad (29)$$

Because the stages are ordered, we also report the quadratic weighted kappa. It penalises an error by the square of its distance along the continuum, so that calling an AD participant CN costs four times as much as calling them MCI:

$$\kappa_w = 1 - \frac{\sum_{a,b} \omega_{ab}\, O_{ab}}{\sum_{a,b} \omega_{ab}\, E_{ab}}, \qquad \omega_{ab} = \frac{(a-b)^2}{(C-1)^2} \quad (30)$$

where O is the observed confusion matrix and E is the matrix expected by chance from its margins. Pairwise AUCs (CN versus AD, CN versus MCI and MCI versus AD) show where on the continuum errors occur. Calibration is measured by the expected calibration error over M = 10 equal-width bins of confidence,

$$\mathrm{ECE} = \sum_{b=1}^{M} \frac{|B_b|}{N} \left| \frac{1}{|B_b|} \sum_{i \in B_b} \mathbb{1}\,[\hat{y}_i = y_i] \; - \; \frac{1}{|B_b|} \sum_{i \in B_b} \max_c p_{i,c} \right| \tag{31}$$

where $B_b$ is the set of participants whose confidence falls in bin b. So that methods are compared on an equal footing, every deep model, comparison methods included, was calibrated in the same way [35]: a per-fold temperature and class bias fitted on that fold's validation predictions, as in the second branch of Eq. (26). ADNI results pool the out-of-fold test predictions of all participants. Methods are compared on the 421 complete cases that every method can score, and AdaptAD is additionally reported on all 597 participants and on the 176 without morphometry. OASIS-3 results are reported on the populations defined in Section 4.3. Uncertainty is quantified with a bootstrap stratified by class [49]. For each of R = 2,000 resamples r of the participants, drawn with replacement within each class, the metric is recomputed, and the 95% confidence interval is given by the 2.5th and 97.5th percentiles. Differences between a method m and AdaptAD are assessed with a paired bootstrap that uses the same resamples for both methods:

$$\Delta^{(r)} = \mathrm{mAUC}_m^{(r)} - \mathrm{mAUC}_{\mathrm{AdaptAD}}^{(r)}, \qquad p = \min\left(1,\; 2\min\left\{\frac{1}{R}\sum_r \mathbb{1}\,[\Delta^{(r)} \le 0],\; \frac{1}{R}\sum_r \mathbb{1}\,[\Delta^{(r)} \ge 0]\right\}\right) \tag{32}$$

The two-sided p-values within each table are adjusted for multiplicity with Holm's procedure [38]. The same resamples give, for each method, the distribution of its rank among all methods. Across the five folds, methods are compared with the Friedman test, and the Nemenyi critical difference is reported. Robustness to missing morphometry is assessed by withholding FreeSurfer from a random fraction ρ ∈ {0, 0.25, 0.5, 0.75, 1} of the participants who have it. For each fraction, 20 random draws are made, and the macro-AUC of each method that uses morphometry is recomputed.

Clinical validity is assessed with the disease score of Eq. (2). Spearman correlations with MMSE and CDR-SB at the scan visit are computed in each cohort, overall and within each diagnostic group. A correlation within a group cannot be produced by a model that has only learned the diagnostic labels, so it is the stronger test of whether the score reflects severity. For ADNI participants with MCI, two prognostic analyses use the follow-up diagnoses. In the first, participants who progressed to dementia within 36 months are compared with those who remained free of dementia for at least 36 months, using the AUC of the score. In the second, time to a dementia diagnosis is compared across tertiles of the score with Kaplan-Meier curves and the log-rank test, using all MCI participants with any follow-up. Clinical usefulness is assessed by decision-curve analysis [50] for two decisions, "impaired (MCI or AD)" and "AD". For a rule that acts on every participant whose predicted probability exceeds a threshold $p_t$, the net benefit is

$$\mathrm{NB}(p_t) = \frac{\mathrm{TP}(p_t)}{N} - \frac{\mathrm{FP}(p_t)}{N} \cdot \frac{p_t}{1 - p_t} \tag{33}$$

which weighs true positives against false positives at the exchange rate implied by the threshold. It is compared with the strategies of acting on everyone and on no one.

To examine where the model looks, saliency maps were computed with SmoothGrad [51], as the magnitude of gradient × input for the AD logit of the MRI-only prediction, averaged over 8 copies of the input perturbed with Gaussian noise of standard deviation 0.1 of the intensity range. The maps were computed for correctly classified test participants of all three seeds and projected from the slice tokens back to MNI voxels. They were then summarised over 14 regions of the Harvard-Oxford atlas [52] distributed with FSL [53] by an enrichment ratio,

$$\mathrm{Enr}(R) = \frac{\sum_{\mathbf{x} \in R} S\,(\mathbf{x}) \,/\, \sum_{\mathbf{x} \in \Omega} S\,(\mathbf{x})}{|R| \,/\, |\Omega|} \tag{34}$$

where S is the saliency map and Ω is the brain volume covered by the slice tokens. A ratio of 1 means that a region receives exactly the share of saliency its size predicts, and values above or below 1 indicate more or less attention than its size predicts. Each ratio is accompanied by a bootstrap confidence interval over participants. Finally, a model-randomisation test [54] recomputed the maps after re-initialising the trainable weights. The Spearman correlation between trained and randomised maps was used to check that the saliency reflects what the model learned rather than image contrast alone.

### 4.7 Implementation details

All code is written in PyTorch [55], with the timm library for the vision transformer backbones. AdaptAD and the comparison methods were trained on an NVIDIA GeForce RTX 4060 Laptop GPU (8 GB) and two NVIDIA GeForce RTX 2080 Ti GPUs (11 GB each), with one run per GPU at a time. Every run saves its complete state at the end of each epoch, including the model, optimiser, scheduler, gradient scaler and random number generators, and resumes from it after an interruption, so an interrupted run continues exactly where it stopped. Every run was trained on a single GPU, with its five folds on the same type of GPU. One AdaptAD epoch took a median of 64 to 85 s. Early stopping ended training after a mean of about 49 epochs per fold, so one seed over five folds took 4.4 to 6.1 h and the three-seed model 16.3 h in total. Full fine-tuning of the same ViT, trained with identical settings on an RTX 2080 Ti, took 61 to 65 s per epoch, comparable to the LoRA variants trained on the same GPUs (64 to 77 s). Low-rank adaptation still requires a forward and backward pass through the entire backbone, so its benefits lie in memory, optimiser state and per-task storage rather than in training time; only the frozen linear probe, which needs no backward pass through the backbone, was markedly faster (42 s per epoch). The 3-D convolutional comparison methods took 15 to 26 s per epoch and MedicalNet 38 s, and the DINOv2 backbone evaluated at Gate 2 took 112 s. Training times for every method and ablation are listed in Table S7. Statistical analyses used NumPy, SciPy and scikit-learn. The Kaplan-Meier estimator and the multi-group log-rank test were implemented directly, with the $\chi^2$ reference distribution taken from SciPy. The code for preprocessing, training, comparison methods, statistics and figures will be released, together with the SHA-256 lock file of the evaluated checkpoints, upon acceptance.

## 5. Results

The results are reported in the order in which the study's questions were posed. Section 5.1 gives the accuracy of three-class staging on the development cohort, compared with nine alternatives trained on identical folds. Section 5.2 examines which design components and which anatomical planes this accuracy depends on, and at what parameter cost. Section 5.3 tests whether the model's output behaves as a measure of severity and progression along the disease continuum. Section 5.4 reports the single locked test on an independent cohort, Section 5.5 the robustness of the model to missing morphometry, and Section 5.6 the anatomical distribution of the evidence used by AdaptAD and by the comparison methods. Unless stated otherwise, intervals are 95% bootstrap confidence intervals, and p-values are two-sided and Holm-adjusted within each table.

### 5.1 Three-class staging on the ADNI cohort

Table 2 and Fig. 3 summarise the comparison of AdaptAD, its two variants and seven comparison methods on the 421 ADNI participants with complete data, the set of participants that every method can score. AdaptAD reached a three-class macro-AUC of 0.745 [0.709, 0.781], with a balanced accuracy of 0.564, a quadratic weighted kappa of 0.545 and an expected calibration error of 0.045. Discrimination was strongest for CN versus AD (AUC 0.916), weakest for CN versus MCI (AUC 0.685) and intermediate for MCI versus AD (AUC 0.791). One-versus-rest ROC curves and reliability diagrams for all methods are shown in Fig. S3. It ranked first by macro-AUC and also had the best mean rank over bootstrap resamples, 1.48 among the ten models, placing first or second in 94% of the 2,000 resamples (first in 60%), so its lead does not depend on the particular sample observed. The ordering in Fig. 3a shows a clear gradient. AdaptAD and its two variants lead the comparison, followed by FreeSurfer logistic regression, MedicalNet, ViT full fine-tuning and the methods that fuse image and tabular data, while the 3-D CNN trained from scratch trails at 0.616 [0.574, 0.657]. The pairwise AUCs show the parts of the continuum at which the model succeeds and those at which it struggles. The two ends of the continuum were separated well, with an AUC of 0.916 for CN versus AD. Distinguishing MCI from AD was harder (0.791), and distinguishing CN from MCI was hardest (0.685). This ordering is expected from the biology of the disease. The MCI group combines participants in the prodromal phase of AD with participants whose impairment will remain stable or has other causes, and the structural changes of early MCI overlap with those of normal ageing. The quadratic weighted kappa of 0.545, which penalises an error by the square of its distance along the continuum, indicates that most misclassifications fall between neighbouring stages rather than across the whole continuum.

The paired comparisons in Fig. 3b quantify the size and certainty of the lead of AdaptAD. After Holm correction, AdaptAD was significantly better than three methods, namely ViT full fine-tuning by 0.037 [0.012, 0.063] ($p = 0.014$), ShaSpec by 0.076 [0.035, 0.120] ($p = 0.008$) and the 3-D CNN by 0.130 [0.084, 0.175] ($p < 0.001$). MedicalNet (by 0.035 [0.005, 0.069]), FiLM (by 0.045 [0.005, 0.084]) and DAFT (by 0.051 [0.012, 0.091]) also scored below AdaptAD with intervals that excluded zero, but these differences did not remain significant after correction for multiple comparisons (Holm p between 0.09 and 0.11). FreeSurfer logistic regression was 0.026 lower ($p = 0.34$). This comparison is favourable to the morphometry model, which is evaluated here only on the 70.5% of participants who have FreeSurfer output and cannot produce a prediction for the remainder.

The lead of AdaptAD does not result from ensembling. A single AdaptAD seed scored 0.742 [0.707, 0.778], within 0.003 of the three-seed model, and still ranked above every comparison method. The comparison with ViT full fine-tuning is the most informative in the table, because the two models share the backbone, the input representation and the training pipeline and differ essentially in which weights are trained, with a lower learning rate used for the fully trained backbone. Training all 88.15 M weights instead of 3.53 M lowered the macro-AUC by 0.037 and increased the calibration error from 0.045 to 0.068. Across the five folds, the methods differed significantly (Friedman test, $p = 0.044$). With only five folds, however, the Nemenyi critical difference is 6.1 ranks, too coarse to separate individual methods. The mean rank of AdaptAD across the five folds was 3.0, and its mean rank over paired bootstrap replicates of the pooled complete cases was 1.48. The fold-level ranking is reported for completeness rather than for inference, because five folds give limited power to separate methods (Fig. S2, Table S8).

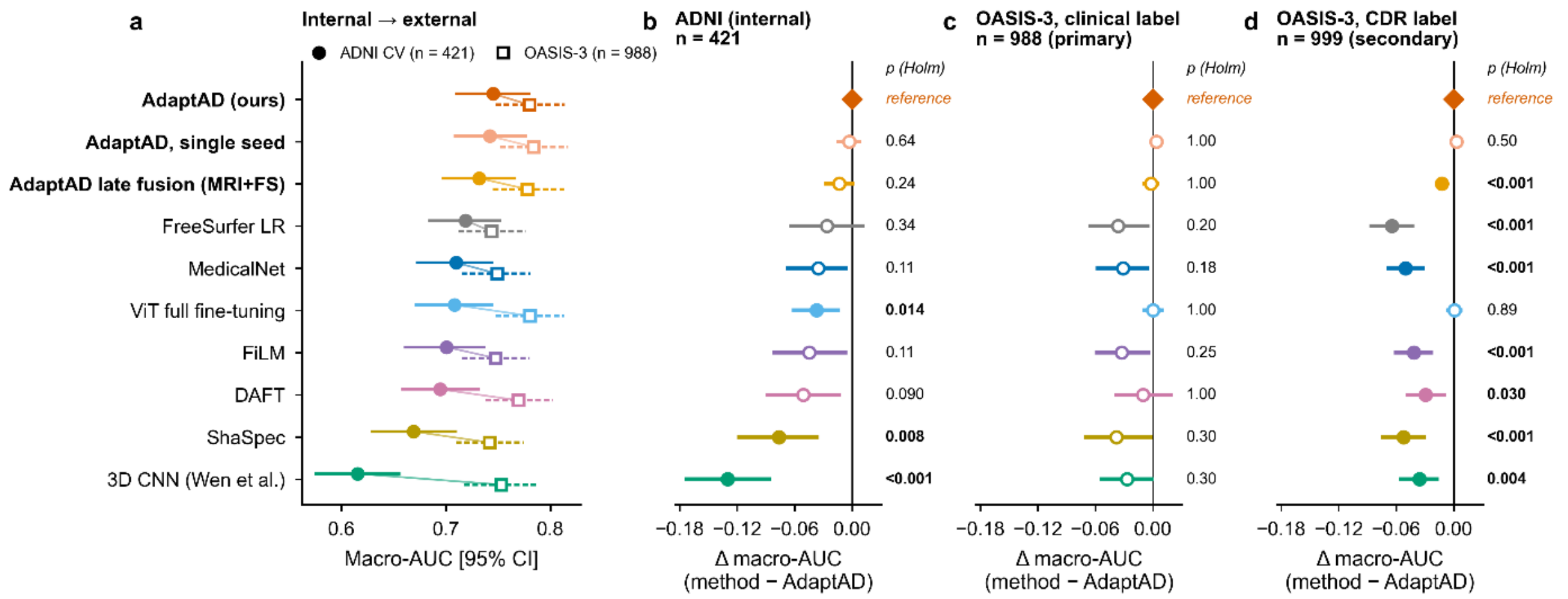

**Fig. 3.** Three-class staging performance of all methods on ADNI and OASIS-3. One row per method, ordered by ADNI macro-AUC. (a) Macro-AUC with 95% CI on ADNI complete cases (n = 421; circles) and on the OASIS-3 primary endpoint (clinical label, n = 988; squares), joined per method. (b to d) Paired difference in macro-AUC (method minus AdaptAD) with 95% CI from 2,000 stratified bootstrap resamples, on (b) ADNI, (c) the OASIS-3 clinical label, the primary endpoint, and (d) the OASIS-3 CDR-based label, a secondary endpoint (n = 999). Holm-adjusted p-values are printed beside each estimate; filled markers indicate Holm $p < 0.05$.

## 5.2 Contribution of design components, parameter efficiency and anatomical planes

Table 3 and Fig. 4 report the ablations of the locked recipe. Each variant changes one element and is trained with the same seed. The reference reached a macro-AUC of 0.741 [0.710, 0.770] for the MRI expert on all 597 participants, and each variant is compared with this value. Only changes to the way the backbone is adapted moved performance noticeably. Freezing the backbone and training only the token projection, fusion layers and heads, which amounts to a linear probe, lowered the macro-AUC by 0.037 [0.011, 0.064]. This was the largest effect of any ablation, with an unadjusted p of 0.006, although after correction for eleven comparisons it fell just short of significance (Holm p =

**Table 3.** Ablations of the locked recipe (ADNI five-fold cross-validation, seed 42). Each variant changes one component of the locked recipe; ΔAUC is the paired difference in macro-AUC (variant minus locked recipe; negative, the change reduces performance) with 95% CI from a paired stratified bootstrap. MRI expert: all participants (n = 597), with Holm-adjusted p over the 11 pre-specified comparisons. Late fusion: complete cases (n = 421), unadjusted p (secondary). ΔAUC, FS removed: late-fusion AUC with FreeSurfer removed for every participant minus AUC with none removed. The ablations were pre-specified and did not change the locked method.

| Variant | MRI expert AUC [95% CI] | ΔAUC [95% CI] | p (Holm | Late fusion AUC | Late fusion ΔAUC [95% CI] | p | ΔAUC, FS removed |
|---|---|---|---|---|---|---|---|
| Locked recipe (reference) | 0.741 [0.710, 0.770] | | | 0.731 | | | +0.011 |
| [MISSING] token → zero vector | 0.738 [0.709, 0.767] | −0.003 [−0.018, +0.011] | 1.000 | 0.721 | −0.010 [−0.025, +0.005] | 0.214 | +0.007 |
| [MISSING] token → attention mask | 0.727 [0.697, 0.758] | −0.014 [−0.033, +0.003] | 1.000 | 0.724 | −0.007 [−0.026, +0.013] | 0.509 | −0.000 |
| No full → MRI distillation | 0.743 [0.712, 0.773] | +0.002 [−0.014, +0.017] | 1.000 | 0.733 | +0.002 [−0.014, +0.018] | 0.783 | +0.004 |
| No FreeSurfer dropout in training | 0.739 [ 0.709, 0.769] | −0.001 [−0.019, +0.016] | 1.000 | 0.727 | −0.004 [−0.022, +0.012] | 0.623 | +0.003 |
| No MRI-only warm-up stage | 0.727 [0.695, 0.761] | −0.013 [−0.034, +0.008] | 1.000 | 0.728 | −0.003 [−0.024, +0.017] | 0.762 | −0.001 |
| No auxiliary MRI-token head | 0.735 [0.705, 0.766] | −0.005 [−0.023, +0.011] | 1.000 | 0.733 | +0.002 [−0.016, +0.020] | 0.843 | +0.004 |
| Stage 2 starts from last stage-1 epoch | 0.730 [0.697, 0.761] | −0.010 [−0.030, +0.006] | 1.000 | 0.729 | −0.001 [−0.019, +0.016] | 0.906 | −0.010 |
| LoRA rank 4 (α = 8) | 0.739 [0.709, 0.772] | −0.001 [−0.018, +0.017] | 1.000 | 0.727 | −0.004 [−0.021, +0.013] | 0.650 | +0.003 |
| LoRA rank 16 (α = 32) | 0.732 [0.702, 0.763] | −0.008 [−0.026, +0.008] | 1.000 | 0.718 | −0.013 [−0.031, +0.005] | 0.160 | +0.004 |
| Frozen backbone (linear probe, no LoRA) | 0.704 [0.672, 0.737] | −0.037 [−0.064, −0.011] | 0.066 | 0.715 | −0.016 [−0.042, +0.010] | 0.237 | −0.019 |
| Full fine-tuning instead of LoRA | 0.720 [0.688, 0.752] | −0.021 [−0.042, +0.001] | 0.610 | 0.711 | −0.019 [−0.040, +0.001] | 0.054 | −0.003 |

*MRI expert: all participants (n = 597), Holm correction over the 11 pre-specified comparisons. Late fusion: complete cases (n = 421), unadjusted p (secondary). ΔAUC, FS removed: late-fusion AUC with FreeSurfer removed for every participant minus AUC with none removed. The ablations explain the method; none changed it (pre-registered).*

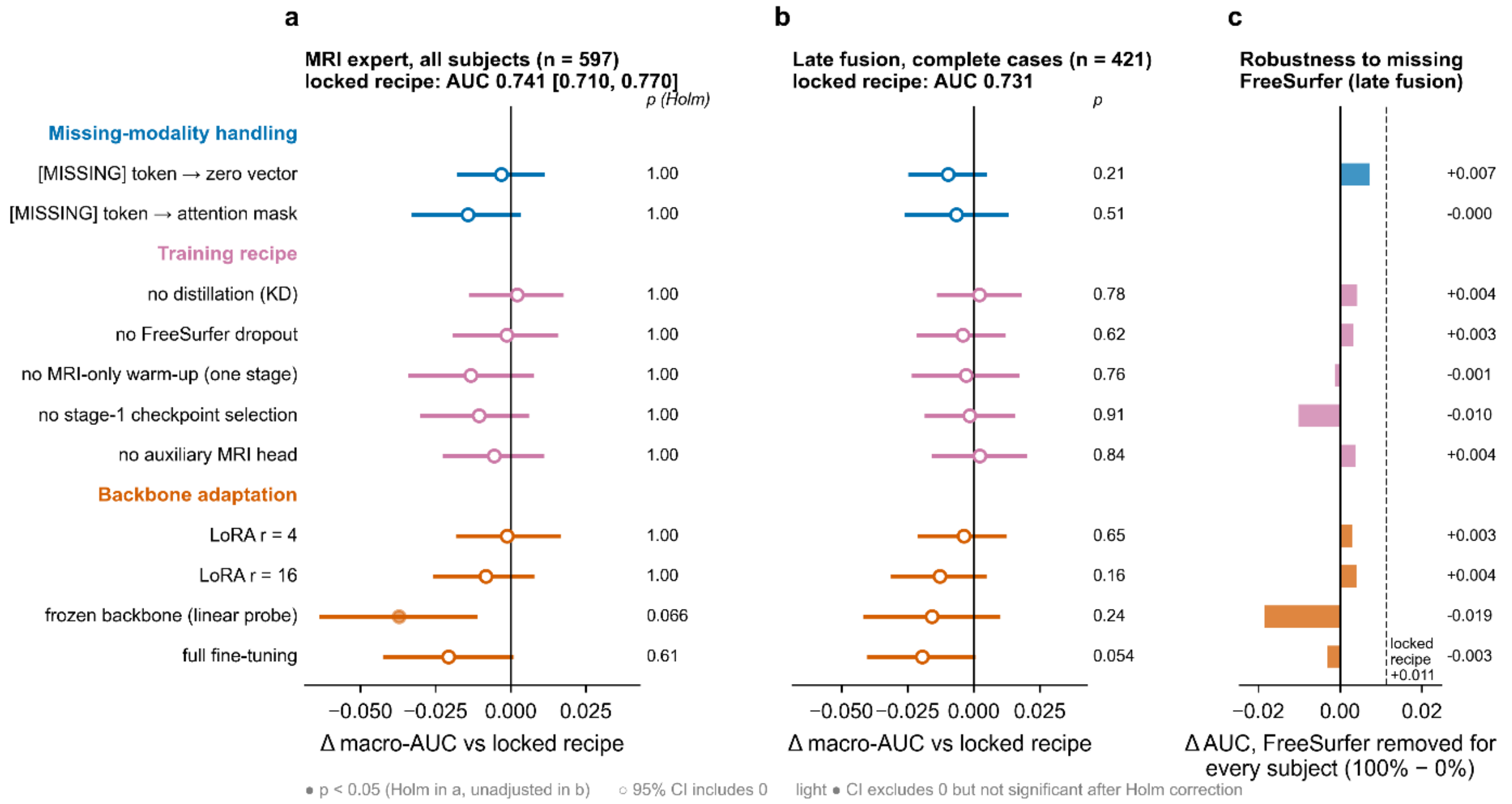

**Fig. 4.** Ablations grouped by the part of the method they change (ADNI five-fold cross-validation, paired stratified bootstrap versus the locked recipe). (a) MRI expert, all participants (n = 597): ΔAUC versus the locked recipe with 95% CI and Holm-adjusted p over the pre-specified family of 11 comparisons. (b) Late fusion, complete cases (n = 421): ΔAUC with 95% CI and unadjusted p. (c) Late-fusion AUC with FreeSurfer removed for every complete case minus AUC with none removed; the dashed line marks the locked recipe. Filled markers, $p < 0.05$ (Holm-adjusted in a, unadjusted in b); light filled markers, CI excludes zero but not significant after Holm correction; open markers, CI includes zero.

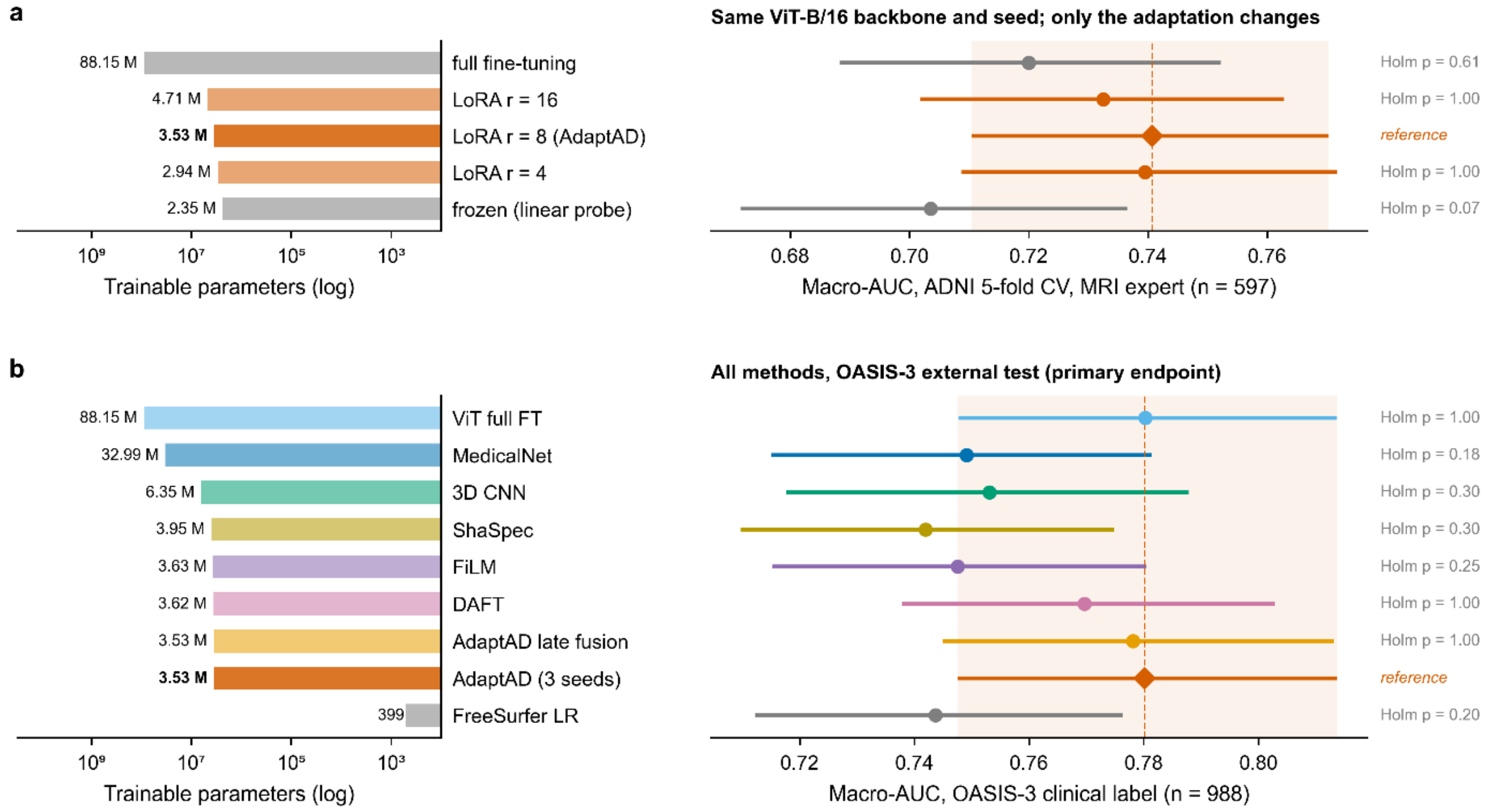

**Fig. 5.** Accuracy against trainable parameters. Each row shows the number of trainable parameters (left, logarithmic scale) and macro-AUC with 95% CI (right); the dashed line and shaded band mark the reference estimate and its CI. (a) Controlled comparison within one ViT-B/16 backbone and seed, changing only the adaptation: frozen linear probe, LoRA with rank 4, 8 and 16, and full fine-tuning (ADNI, MRI expert, n = 597; Holm p versus LoRA rank 8 within the pre-specified family of 11 ablation comparisons). (b) AdaptAD, its late-fusion variant and the seven comparison methods on the OASIS-3 primary endpoint (n = 988; Holm p versus AdaptAD); the late-fusion row counts the parameters of AdaptAD and the FreeSurfer logistic regression together.

0.066). Training all backbone weights instead of low-rank updates lowered the AUC by 0.021 [−0.001, 0.042] (Holm p = 0.61). The rank of the low-rank update had little influence, since rank 4 changed the AUC by −0.001 and rank 16 by −0.008, both with intervals centred close to zero. The training and missing-modality components did not change the M RI expert detectably, and every one of these variants had an interval that included zero. Replacing the [MISSING] token with a zero vector changed the AUC by −0.003 and replacing it with attention masking by −0.014. Removing distillation changed it by +0.002, removing FreeSurfer dropout by −0.001, removing the auxiliary head by

−0.005, removing the MRI-only warm-up by −0.013 and removing stage-1 checkpoint selection by −0.010. The late-fusion results in Fig. 4b agree with this picture, and Fig. 4c shows that removing FreeSurfer at test time changed the late-fusion AUC by between −0.019 and +0.011 across all variants. Taken together, the ablations indicate that the gain of AdaptAD comes primarily from adapting the pretrained backbone with a small number of parameters. The individual training and missing-modality components are not necessary for the accuracy of the MRI expert, although they do no harm. Fig. 5a places these results on a parameter axis. With the backbone, seed and pipeline held fixed, performance rose from 0.704 with the frozen backbone (2.35 M trainable parameters) to 0.739 and 0.741 with low-rank adaptation at ranks 4 and 8 (2.94 and 3.53 M), remained at 0.732 at rank 16, and fell to 0.720 when all 88.15 M weights were trained. The relationship between performance and the number of trained parameters is therefore not monotonic. A small, targeted adaptation is beneficial, but adapting every weight with about 400 training participants per fold gives part of that benefit back, which is the behaviour expected when the capacity of a model exceeds what the data can constrain. The low-rank update itself contributes only 1.18 M parameters, about 1.4% of the backbone. Training time per epoch was not reduced by low-rank adaptation, since on the same RTX 2080 Ti the low-rank variants took 69 to 75 s per epoch and full fine-tuning 63 s (Table S5), because both require a backward pass through the entire backbone.

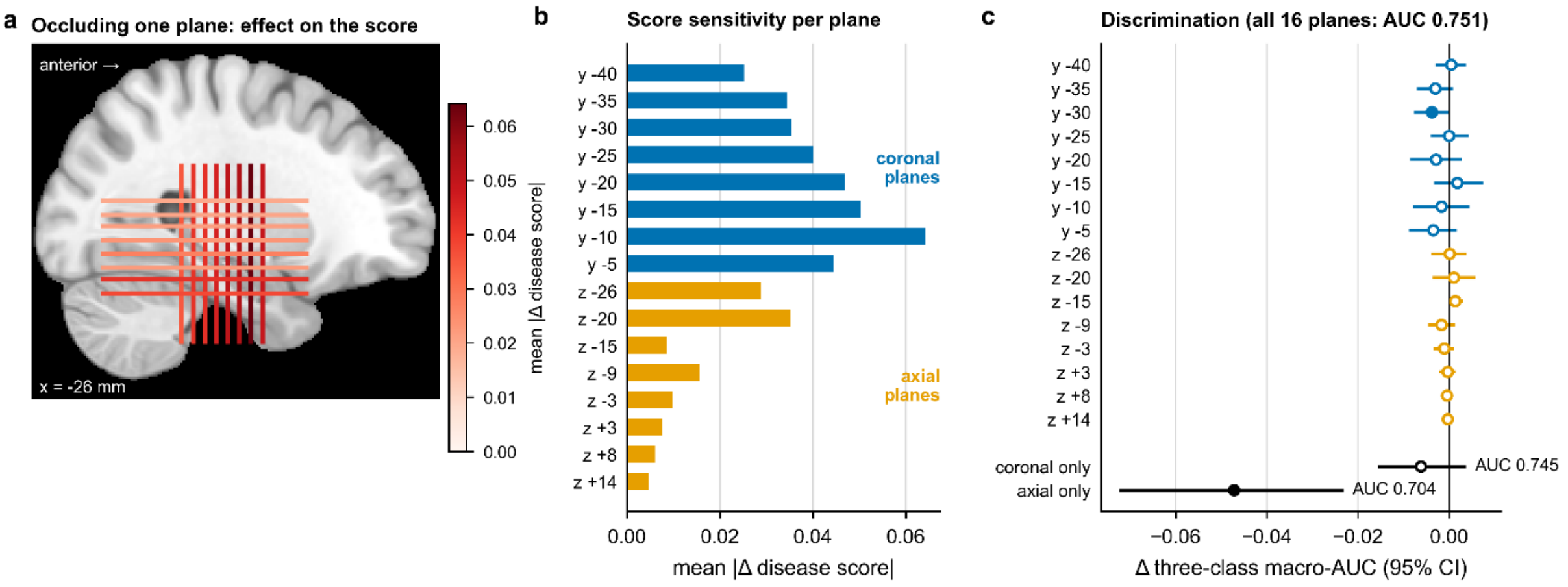

**Fig. 6.** Contribution of the 16 slice planes to AdaptAD's decision (token occlusion, ADNI, n = 597, out-of-fold, MRI-only pathway with FreeSurfer replaced by the [MISSING] token; predictions averaged over three seeds and the flipped volume). A plane is occluded by replacing its token with the mean token at that position over the fold's validation participants; calibration is fitted on unoccluded validation predictions and applied unchanged. (a) The 16 planes on a sagittal MNI152 slice through the hippocampus (x = −26 mm), coloured by the mean absolute change in disease score when the plane alone is occluded. (b) The same sensitivity for each plane, grouped into coronal and axial planes. (c) Change in three-class macro-AUC with 95% CI (paired stratified bootstrap, 2,000 resamples) when one plane, all axial planes (coronal only) or all coronal planes (axial only) are occluded, relative to the unoccluded model (macro-AUC 0.751 on all 597 participants); the absolute AUCs of the two subsets are printed. Occluding any single plane changed macro-AUC by 0.004 or less.

The practical saving of low-rank adaptation therefore lies in memory, optimiser state and the storage needed for each adapted model rather than in training time. The corresponding view across all methods on the external cohort is given in Fig. 5b and discussed in Section 5.4. Fig. 6 addresses a different question, namely which of the 16 anatomical planes the decision depends on. Each plane was occluded in turn by replacing its token with the average token at that position over the validation participants of the fold, and the change in output was measured on the held-out test participants. With all planes present, the analysis reproduced the reported macro-AUC of AdaptAD on all 597 participants exactly (0.751), confirming that it measures the headline model rather than an approximation of it. No single plane proved decisive. Occluding any one plane changed the macro-AUC by at most 0.004, and fifteen of the sixteen intervals included zero (Fig. 6c). The one exception, the coronal plane at y = −30 mm, excluded zero only marginally and is not corrected for sixteen comparisons. The information used by the model is therefore distributed redundantly across neighbouring planes, a property that should make the model robust to a single poorly registered or artefacted slice. The two orientations, by contrast, were far from equivalent. Keeping only the eight coronal planes preserved almost all of the accuracy (macro-AUC 0.745; ΔAUC −0.006 [−0.016, 0.004]), whereas keeping only the eight axial planes lowered it substantially (0.704; ΔAUC −0.047 [−0.072, −0.023]; Fig. 6c). The sensitivity map in Fig. 6a and b shows that the disease score responded most strongly to the coronal planes at y = −10, −15 and −20 mm, which pass through the head and anterior body of the hippocampus and the adjacent temporal horns of the lateral ventricles. Occluding one of these planes changed the score by 0.047 to 0.064 on average, whereas occluding one of the superior axial planes at z = −3 to 14 mm changed it by only 0.005 to 0.010. The coronal planes through the medial temporal lobe thus carry most of the decision, and the superior axial planes, together with the axial plane at z = −15 mm, contribute least.

### 5.3 Clinical validity along the disease continuum

Fig. 7 illustrates how AdaptAD reads five participants placed along the continuum. To avoid selecting cases by eye, they were chosen by a fixed rule, which took, among ADNI participants with complete data whose stage AdaptAD predicted correctly, the participant whose disease score was closest to the median of their group. One OASIS-3 participant with AD was chosen in the same way. Each ADNI participant was read by the model of the fold that held them out, and the OASIS-3 participant by the locked fold models. Read from left to right, Fig. 7 shows anatomy, model output and clinical severity changing together. The cognitively normal participant, with a disease score of 0.48, an MMSE of 29 and a CDR-SB of 1.0, has narrow lateral ventricles and full medial temporal structures. At the other end, the ADNI participant with AD, with a score of 1.88, an MMSE of 12 and a CDR-SB of 14.0, shows widened lateral ventricles and temporal horns and a loss of medial temporal volume that is visible on the raw scan. The two participants with MCI lie between these ends and are ordered as their later clinical course would predict. The participant who remained free of dementia for 36 months, and received a dementia diagnosis at 38 months, scored 0.92, whereas the participant diagnosed with dementia at 24 months scored 1.24 and had the higher probability of AD of the two (0.34). The OASIS-3 participant with AD scored 1.68 with milder impairment (MMSE 18, CDR-SB 6.0), consistent with the milder AD group of that cohort. The coronal saliency concentrates on the lateral ventricles, the periventricular white matter and the inferior temporal region in every column. The axial saliency at z = −15 mm lies mostly along the midline, in agreement with the small contribution of this plane found in Fig. 6 (mean absolute change in score 0.008). The strip at the bottom of the figure places the five participants within the full distribution of scores, in which cognitively normal participants concentrate at low scores, participants with AD at high scores, and participants with MCI spread across the middle of the range.

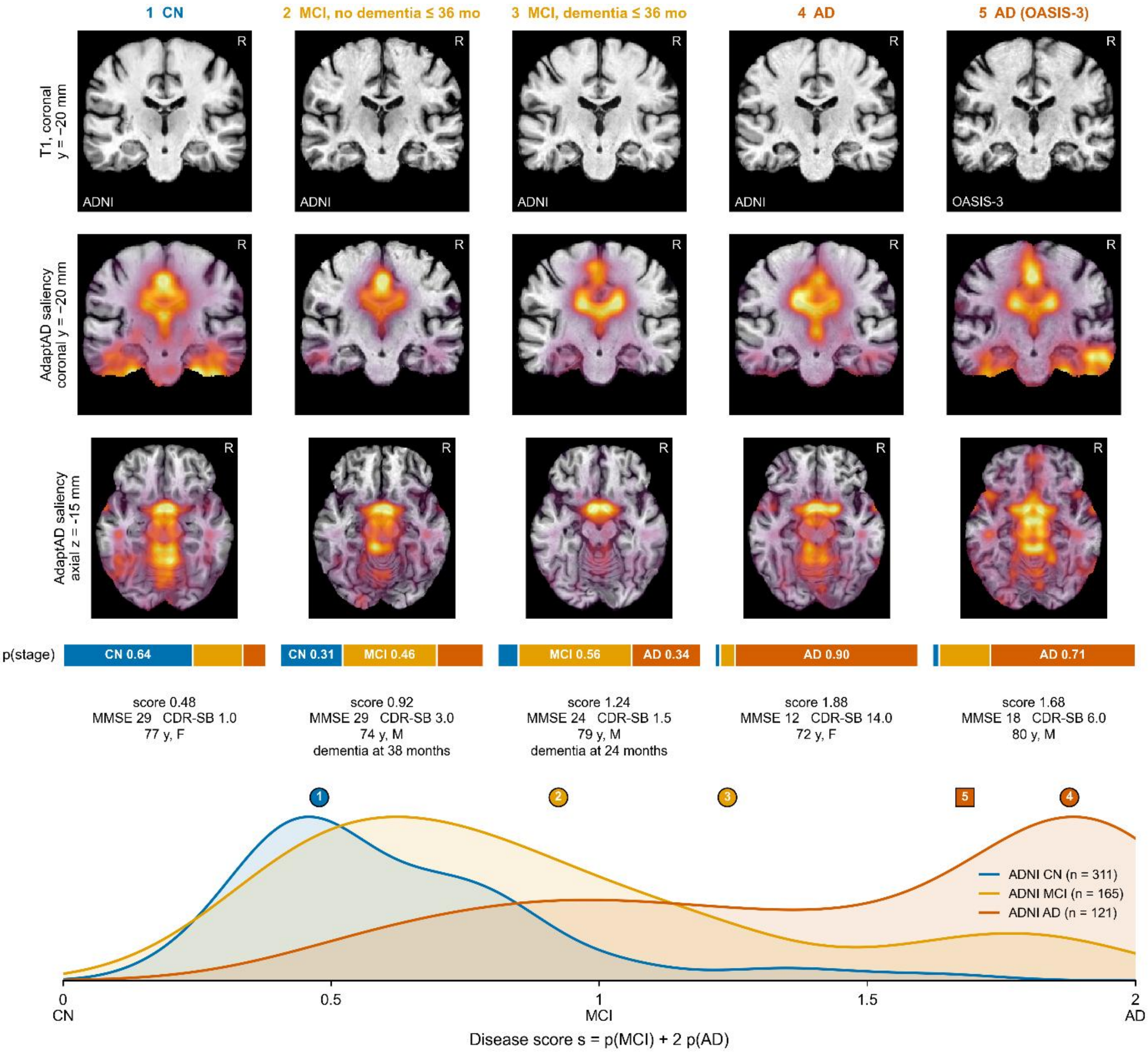

**Fig. 7.** The disease continuum as read by AdaptAD. Columns: five participants chosen by a fixed rule (complete cases with FreeSurfer, correctly staged by AdaptAD, and closest to their group's median disease score; ADNI participants read by the fold model that held them out, the OASIS-3 participant by all locked fold models), numbered and ordered along the continuum; MCI groups follow the pre-specified 36-month definitions, so a participant labelled MCI without dementia within 36 months may be diagnosed later. Rows: the participant's own registered T1 at y = −20 mm; AdaptAD's SmoothGrad saliency for the AD output (averaged over three seeds) on the same coronal token plane and on the axial token plane at z = −15 mm, each showing only that plane's own token, with 10 mm display smoothing; calibrated stage probabilities; disease score, MMSE, CDR-SB, age, sex and, for MCI, the time of dementia diagnosis where it occurred. Bottom: distribution of the out-of-fold disease scores of all 597 ADNI participants by diagnosis (kernel density, each scaled to its peak), with the five participants marked by number (circles, ADNI; square, OASIS-3).

Fig. 8 quantifies these observations across the cohort. The disease score correlated with the MMSE (Spearman ρ = −0.545, n = 588) and with the CDR-SB (ρ = 0.563, n = 573), and the binned medians in Fig. 8a and b rose steadily with CDR-SB and fell steadily with MMSE. Correlations across the whole cohort could, however, be produced by a model that had learned only the three diagnostic labels. The stronger test is therefore the correlation within each diagnostic group, which a relabelled diagnosis cannot produce. Within the AD group the score correlated with CDR-SB at ρ = 0.591 and with MMSE at ρ = −0.461, and within the MCI group at 0.214 and −0.364. Within the CN group the correlations were weak (0.099 and −0.148), as expected in a group with little impairment to grade. The score therefore grades severity within a stage as well as between stages, which is the property expected of a continuous marker of disease. The score also anticipated progression from MCI to dementia. Of the 165 ADNI participants with MCI, 143 had follow-up after the scan. Among the 65 who either progressed to dementia within 36 months (12) or remained free of dementia for at least 36 months (53), the score separated the two groups with an AUC of 0.786 [0.612, 0.926] (Fig. 8c; Table S10). With only 12 converters this estimate is exploratory, and its interval is wide. The time-to-event analysis uses all 143 participants and provides the stronger evidence. Of the 21 dementia diagnoses recorded during follow-up, 17 occurred in the highest score tertile, 3 in the middle tertile and 1 in the lowest. Time to diagnosis differed markedly between tertiles (log-rank $\chi^2$ = 27.2, p = 1.3 × $10^{-6}$; Fig. 8d). The three curves were indistinguishable at 12 months, when between 97% and 100% of each tertile remained free of dementia. The highest tertile separated from the others by 24 months, when 86% of it remained free of dementia against 96% and 97% of the lowest and middle tertiles, and it continued to diverge thereafter. By 36 months 65% of the highest tertile remained free of dementia, and by 57 months 35%, against 96% in the lowest and 88% in the middle tertile. Few participants remained at risk after 48 months (5 in the highest tertile at 57 months), so the late estimates are imprecise. In the decision curves on the complete cases (Fig. 8e, f), AdaptAD provided a net benefit above treating all and treating none across thresholds from 20% to 80% for impairment and from 20% to 60% for AD. Its net benefit was similar to that of FreeSurfer logistic regression for impairment, with differences within 0.02, and higher for AD between thresholds of 20% and 60% (for example 0.088 against 0.065 at 30%), while requiring no morphometry. Results in subgroups defined by age, sex, APOE e4 status, education, scanner manufacturer and FreeSurfer availability are given in Table S9 and Fig. S4. Macro-AUC was highest in APOE ε4 carriers (0.810) and in participants younger than 70 years (0.794), and lowest in participants aged 80 years or older (0.709), in APOE ε4 non-carriers (0.708) and in those with fewer than 16 years of education (0.709), whereas it was similar across scanner manufacturers (0.749 to 0.753) and with or without FreeSurfer (0.745 and 0.766).

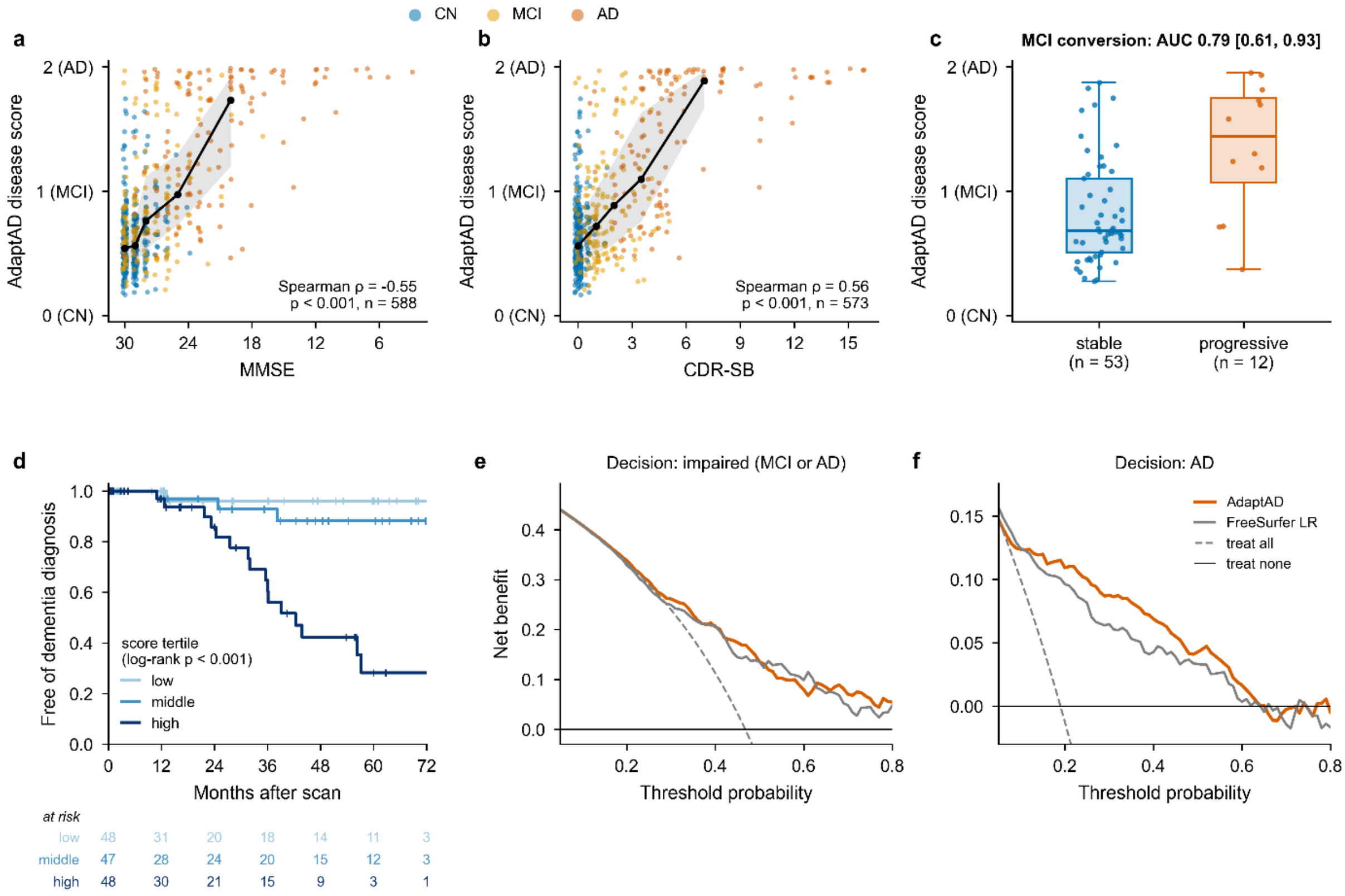

**Fig. 8.** Clinical validity of the AdaptAD disease score s = p(MCI) + 2 p(AD), from out-of-fold ADNI predictions. (a, b) Score against MMSE (n = 588) and CDR-SB (n = 573), coloured by diagnosis, with Spearman's ρ; black line and band, median and interquartile range of the score in eight equal-count bins of the clinical score. (c) MCI participants who progressed to dementia within 36 months (n = 12) against those who remained free of dementia for at least 36 months (n = 53); AUC with 95% CI from 2,000 stratified bootstrap resamples. (d) Kaplan-Meier curves of time to dementia diagnosis by tertile of the score (n = 143 MCI participants with follow-up; ticks, censored participants), with numbers at risk below and the log-rank p. (e, f) Decision curves for the decisions "impaired (MCI or AD)" and "AD" over threshold probabilities from 0.05 to 0.80, using AdaptAD's calibrated probabilities (1 − p(CN) and p(AD)) against FreeSurfer logistic regression and the strategies of treating all or none (complete cases, n = 421).

### 5.4 One-shot external validation on OASIS-3

Table 4 and Fig. 9 report the single test of the locked models on OASIS-3, an independent cohort that differs from ADNI in its scanners, recruitment and diagnostic practice. On the pre-registered primary endpoint, the three-class macro-AUC with the clinical label, AdaptAD reached 0.780 [0.748, 0.814] (n = 988; Table 4, Fig. 9). Of the 993 OASIS-3 participants with a primary clinical label, comparisons used the 988 with FreeSurfer available, so that every method, including the morphometry-only model, could be scored on the same participants. In the pre-specified sensitivity analysis on all 993 labelled participants, which adds the five participants without FreeSurfer whom methods based on morphometry cannot score, it reached 0.782 [0.750, 0.813]. Every method scored higher on OASIS-3 than on ADNI (Fig. 3a), and the 3-D CNN rose the most, from 0.616 to 0.753. A rise shared by all methods points to differences between the cohorts rather than to any single model performing better. OASIS-3 contains a large proportion of clearly normal participants, its MCI group is small, and its labels combine a clinical diagnosis with the CDR, all of which make the three classes easier to separate. On the primary endpoint AdaptAD was ahead of FreeSurfer logistic regression by 0.036 [0.003, 0.067], of MedicalNet by 0.031 [0.004, 0.060], of FiLM by 0.033 [0.002, 0.061], of ShaSpec by 0.038 [0.000, 0.072] and of the 3-D CNN by 0.027 [0.000, 0.055], although none of these differences remained significant after Holm correction (adjusted p between 0.18 and 0.30). DAFT trailed AdaptAD by 0.010, and ViT full fine-tuning tied with it at 0.780. The secondary endpoints were defined in advance precisely because they do not depend on the clinician's diagnosis, and they sharpened the comparison. On the CDR-based three-class label (n = 999; Fig. 3d), AdaptAD reached 0.804 [0.781, 0.827] and was significantly better, after Holm correction, than every method that does not use a vision transformer (all adjusted $p \leq 0.03$). Its lead was 0.065 over FreeSurfer logistic regression, 0.052 over ShaSpec, 0.050 over MedicalNet, 0.042 over FiLM, 0.036 over the 3-D CNN and 0.029 over DAFT. It was also better than its own late fusion, by 0.012 [0.006, 0.019]. For the binary distinction between cognitively normal and impaired participants (Fig. 9a), AdaptAD reached an AUC of 0.846 [0.817, 0.874], which tied with full fine-tuning (0.849) and exceeded every 3-D method (0.813 to 0.834). Fig. 5b relates these external results

**Table 4.** One-shot external test on OASIS-3 across all pre-registered endpoints. Clinical label (primary, n = 988) and CDR-based label (secondary, n = 999): three-class macro-AUC with 95% CI, balanced accuracy (BAcc), expected calibration error (ECE) and Holm-adjusted p of the paired difference from AdaptAD. CN versus impaired (CDR ≥ 0.5): AUC of 1 − p(CN) with 95% CI. CIs and p-values from 2,000 stratified bootstrap resamples (paired for p). Models, calibration and the decision rule (highest calibrated probability) were locked before any OASIS-3 prediction, no OASIS-3 data were used for training, tuning or recalibration, and the test was run once.

| Method | Clinical label: macro-AUC [95% CI] | BAcc | ECE | p (Holm) | CDR label: macro-AUC [95% CI] | BAcc | ECE | p (Holm) | CN vs impaired: AUC [95% CI] |
|---|---|---|---|---|---|---|---|---|---|
| AdaptAD (ours) | 0.780 [0.748, 0.814] | 0.545 | 0.167 | reference | 0.804 [0.781, 0.827] | 0.613 | 0.132 | reference | 0.846 [0.817, 0.874] |
| AdaptAD, single seed | 0.784 [0.752, 0.817] | 0.552 | 0.175 | 1.000 | 0.807 [0.784, 0.830] | 0.602 | 0.125 | 0.504 | 0.847 [0.818, 0.875] |
| AdaptAD late fusion (ours, MRI + FS) | 0.778 [0.745, 0.813] | 0.558 | 0.138 | 1.000 | 0.792 [0.769, 0.815] | 0.600 | 0.096 | <0.001 | 0.846 [0.817, 0.874] |
| FreeSurfer logistic regression | 0.744 [0.712, 0.776] | 0.522 | 0.109 | 0.200 | 0.740 [0.715, 0.763] | 0.575 | 0.068 | <0.001 | 0.818 [0.786, 0.849] |
| 3D CNN (Wen et al.) | 0.753 [0.718, 0.788] | 0.551 | 0.342 | 0.300 | 0.769 [0.741, 0.795] | 0.590 | 0.296 | 0.004 | 0.819 [0.788, 0.850] |
| MedicalNet (Med3D) | 0.749 [0.715, 0.781] | 0.538 | 0.201 | 0.180 | 0.754 [0.729, 0.779] | 0.602 | 0.182 | <0.001 | 0.834 [0.806, 0.864] |
| ViT full fine-tuning | 0.780 [0.748, 0.814] | 0.548 | 0.144 | 1.000 | 0.805 [0.781, 0.828] | 0.601 | 0.131 | 0.889 | 0.849 [0.820, 0.876] |
| DAFT | 0.770 [0.738, 0.803] | 0.514 | 0.142 | 1.000 | 0.775 [0.751, 0.798] | 0.564 | 0.057 | 0.030 | 0.827 [0.797, 0.857] |
| FiLM | 0.748 [0.715, 0.781] | 0.513 | 0.211 | 0.252 | 0.763 [0.740, 0.784] | 0.574 | 0.141 | <0.001 | 0.825 [0.795, 0.855] |
| ShaSpec | 0.742 [0.710, 0.775] | 0.540 | 0.160 | 0.300 | 0.752 [0.727, 0.776] | 0.555 | 0.111 | <0.001 | 0.813 [0.780, 0.844] |

*Clinical label (primary endpoint): n = 988 (CN 736, MCI 65, AD 187). CDR label (secondary): CDR 0 / 0.5 / ≥ 1, n = 999. CN vs impaired: CDR ≥ 0.5. p (Holm): paired stratified bootstrap of the macro-AUC difference from AdaptAD, Holm-corrected within each endpoint; bold, Holm p < 0.05. Models, calibration and thresholds were locked (Gate 3) before any OASIS-3 prediction, and the test was run once.*

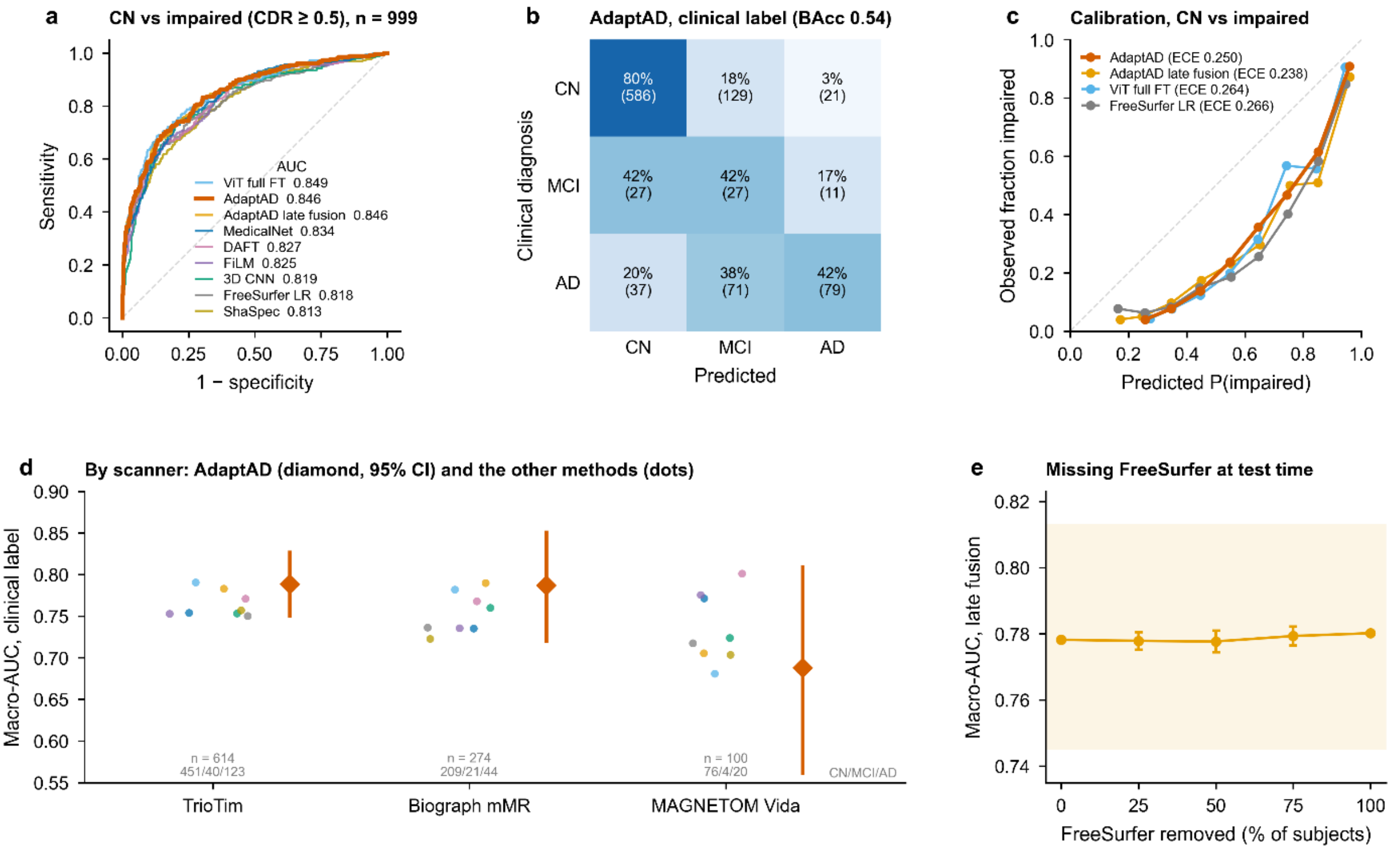

**Fig. 9.** External test on OASIS-3 in detail. (a) ROC curves for the pre-registered binary endpoint CN versus impaired (CDR ≥ 0.5) for AdaptAD and the comparison methods, AdaptAD in bold, with AUCs in the legend (participants scored by every method). (b) AdaptAD confusion matrix for the primary clinical label (row-normalised percentages with counts; balanced accuracy in the title). (c) Reliability diagram for the probability of impairment, 1 − p(CN), against CDR-based impairment, for AdaptAD, AdaptAD late fusion, FreeSurfer logistic regression and ViT full fine-tuning (ten equal-width bins, bins with at least 10 participants shown; ECE in the legend). (d) Macro-AUC for the primary clinical label by scanner model, for scanners with at least 20 participants and all three classes; AdaptAD with 95% CI from 1,000 stratified bootstrap resamples (diamond) and the other methods (dots), with n and CN/MCI/AD counts per scanner. (e) Late fusion with FreeSurfer removed for a random fraction of participants (0%, 25%, 50%, 75% and 100%; mean ± SD over 20 draws for the intermediate fractions), among participants with FreeSurfer and a clinical label; the shaded band is the late-fusion 95% CI with FreeSurfer present.

to the number of trainable parameters. AdaptAD reached 0.780 with 3.53 M trainable parameters, a level matched only by full fine-tuning with 88.15 M, whereas MedicalNet with 32.99 M and the 3-D CNN with 6.35 M reached 0.749 and 0.753.

The confusion matrix in Fig. 9b shows a clear structure in the errors of AdaptAD. It recognised 80% of cognitively normal participants and 42% of participants with AD, and assigned 38% of participants with AD to the MCI stage. This pattern follows from the composition of the cohort rather than from a failure of the model. Participants with AD in OASIS-3 were markedly milder than those in ADNI, with a CDR-SB of 3.5 ± 1.8 against 5.7 ± 3.4 and a mean MMSE of 24.9 against 21.3 (Table 1A), and 126 of the 189 had a global CDR of only 0.5, so many of them resemble the ADNI participants with MCI from whom the boundary between MCI and AD was learned. The errors are therefore concentrated at the adjacent boundary of the continuum rather than spread across it. Calibration shifted in the same direction for every method (Fig. 9c). The three-class expected calibration error of AdaptAD was 0.167 (Table 4), and for the binary probability of impairment, predicted risk exceeded observed risk across the whole range for AdaptAD, its late fusion, full fine-tuning and FreeSurfer logistic regression, with expected calibration errors between 0.24 and 0.27. This is the expected consequence of applying models calibrated in a cohort in which 48% of participants are impaired to a cohort in which only 26% are. Performance was consistent across scanner models (Fig. 9d). AdaptAD reached 0.788 [0.748, 0.829] on the Siemens TrioTim (n = 614) and 0.787 [0.718, 0.853] on the Biograph mMR (n = 274). On the MAGNETOM Vida (n = 100, of whom only four had MCI) it reached 0.688 [0.559, 0.811], an interval that is wide and overlaps both the other scanners and the point estimates of most methods. The external summary and manifest are provided in Table S11. Finally, the relation between the disease score and clinical severity carried over to the external cohort, with a Spearman ρ of −0.477 with MMSE and 0.535 with CDR-SB (n = 993, both p < 0.001).

## 5.5 Robustness to missing morphometry

Participants without FreeSurfer output were not harder for AdaptAD to stage. On all 597 ADNI participants the model reached a macro-AUC of 0.751, and on the 176 participants without morphometry it reached 0.766. Because its MRI-only path requires no morphometry, every participant receives a prediction from the same model, and the absence of FreeSurfer neither excludes a participant nor degrades the prediction. The controlled removal experiments confirm this robustness. Withholding FreeSurfer from a random fraction of the OASIS-3 participants who have it changed AdaptAD's late fusion by less than 0.003 at every fraction from 0% to 100%, with the macro-AUC moving only between 0.778 and 0.780 (Fig. 9e). On ADNI complete cases, removing FreeSurfer from every participant changed the AUC of the late-fusion model by +0.011 in the ablation reference run (Table 3) and by +0.014 in the full three-seed model (Fig. S5). In contrast, the AUC of the image and tabular fusion methods DAFT, FiLM and ShaSpec fell by 0.045, 0.062 and 0.041. Removal curves for all methods on ADNI are shown in Fig. S5. Across all ablation variants, complete removal of morphometry changed the late-fusion AUC by between −0.019 and +0.011 (Table 3, Fig. 4c). Adding morphometry did not improve accuracy either. On ADNI, the late fusion scored 0.732 against 0.745 for the MRI expert alone (ΔAUC −0.014 [−0.029, 0.002]), and on OASIS-3 it scored 0.778 against 0.780. The model based on morphometry alone gained least between the two cohorts. It reached 0.719 on ADNI and, on OASIS-3, 0.744 with the clinical label and 0.740 with the CDR label, compared with 0.780 and 0.804 for AdaptAD. This is plausibly related to the change of FreeSurfer version, from UCSF FreeSurfer 6 in ADNI to FreeSurfer 5.3 in OASIS-3, which alters regional volume and thickness estimates. In this study, therefore, morphometry was optional rather than helpful, and it was the image pathway that transferred between cohorts.

### 5.6 Anatomical distribution of model evidence

Fig. 10 summarises the saliency of AdaptAD, computed on the MRI-only pathway, over the test participants of all folds. In correctly classified participants with AD (n = 76), the share of saliency relative to the share of covered brain volume was highest in the lateral ventricles (2.82 [2.70, 2.94]), followed by the parahippocampal gyrus (2.06), the temporal fusiform cortex (1.95), the hippocampus (1.74) and the posterior cingulate cortex (1.59). It was lowest in the frontal pole (0.32) and the lateral occipital cortex (0.28), regions that are comparatively spared in early AD and therefore, serve as anatomical controls (Fig. 10c; Table S12). Correctly classified cognitively normal participants (n = 222) showed the same ordering of regions, with enrichment above 1 in the lateral ventricles (2.50) and the hippocampus (1.47), which indicates that the model inspects these structures in every brain rather than only in diseased ones. Enrichment was nevertheless higher in AD than in CN in the lateral ventricles, the hippocampus, the posterior cingulate cortex (1.59 against 1.22) and the thalamus (1.50 against 1.27), whereas participants with CN drew relatively more saliency in the frontal pole, the lateral occipital cortex and the insula. The difference maps in Fig. 10b correspondingly place the signal specific to AD around the lateral ventricles and in the inferior medial temporal lobe on the coronal plane at y = −20 mm, at the level of the hippocampal body. When the LoRA and classification-head weights were re-initialised while the pretrained backbone was kept, the resulting maps correlated with those of the trained model at only ρ = 0.22 (30 test participants of fold 0; Fig. 10d), which indicates that the maps reflect what the model learned during adaptation rather than image contrast alone. Saliency maps of individual participants are shown in Fig. S6.

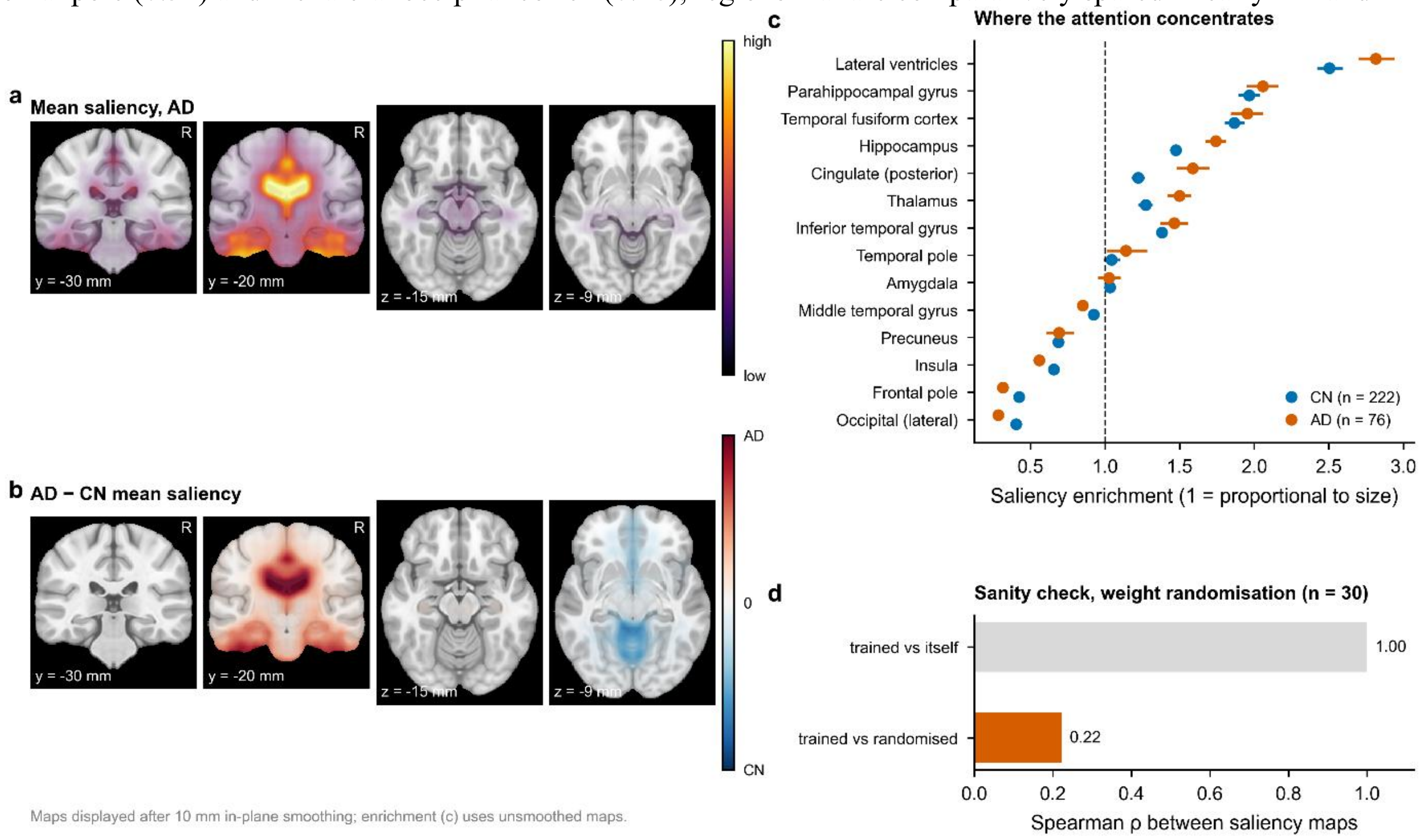

**Fig. 10.** Fig. 10. Saliency of AdaptAD (SmoothGrad of |gradient × input| for the AD logit of the MRI-only pathway, with FreeSurfer replaced by the [MISSING] token; ADNI test folds, correctly classified participants; maps averaged over three seeds and normalised to sum 1 per participant). Attribution exists only on the 16 token planes (three slices each, ±2 mm), and all statistics are computed within these covered voxels. (a) Mean saliency of participants with AD on the MNI152 template at two coronal and two axial token planes through the medial temporal lobe. (b) Difference between the mean saliency of participants with AD and CN on the same planes (red, more in AD; blue, more in CN). (c) Region enrichment, the share of saliency divided by the share of covered brain volume, for AD (n = 76) and CN (n = 222) participants with 95% CI from bootstrap over participants, over Harvard-Oxford regions; a value of 1 means no more attention than the region's size predicts. (d) Model-randomisation check on 30 test participants of fold 0: rank correlation between the group map of the trained model and that of a model whose LoRA and classification-head weights were re-initialised (the trained model against itself gives 1). Maps are displayed after 10 mm in-plane smoothing; enrichment is computed on the unsmoothed maps.

Fig. 11 places the evidence of AdaptAD beside that of the six other methods that read the image, for the same participants as Fig. 7 and the same coronal plane. Every method was attributed with the same SmoothGrad procedure, and each participant was read by the fold model that had held them out, so that differences between columns reflect differences between methods rather than between procedures. The methods fall into recognisable patterns. AdaptAD and full fine-tuning produce nearly interchangeable maps in every row, with their evidence on the lateral ventricles, the periventricular white matter and the inferior temporal region. MedicalNet is the only method whose evidence sits directly on the hippocampal region, bilaterally and consistently across participants. The 3-D CNN spreads its saliency over the whole brain and brainstem, and its calibrated output is close to uniform, about 0.33 for each stage, consistent with its low discrimination. DAFT, FiLM and ShaSpec show scattered focal spots in the cortex, brainstem and cerebellum. The predicted stages beneath the maps agree with the accuracy results: the comparison methods misclassify the participants with MCI more often, mostly towards AD. The population analysis over all 421 ADNI participants with complete data (Table S13) confirms and quantifies this picture. On the 16 token planes, every method concentrated its saliency on the lateral ventricles, at between 2.2 and 4.0 times the share expected from their area: 2.5 [2.5, 2.6] for AdaptAD, 3.2 [3.1, 3.3] for full fine-tuning and 3.0 [2.9, 3.0] for MedicalNet. In the medial temporal lobe, by contrast, the methods diverged. MedicalNet placed 2.7 [2.7, 2.8] times the expected share of its saliency there, AdaptAD 1.6 [1.6, 1.7] and full fine-tuning 1.7 [1.7, 1.8], whereas the 3-D CNN and the three fusion methods remained close to the level expected by chance (0.9 to 1.2). The single coronal plane at y = −20 mm gave the same ordering of methods, so the conclusion does not depend on the plane shown in the figure.

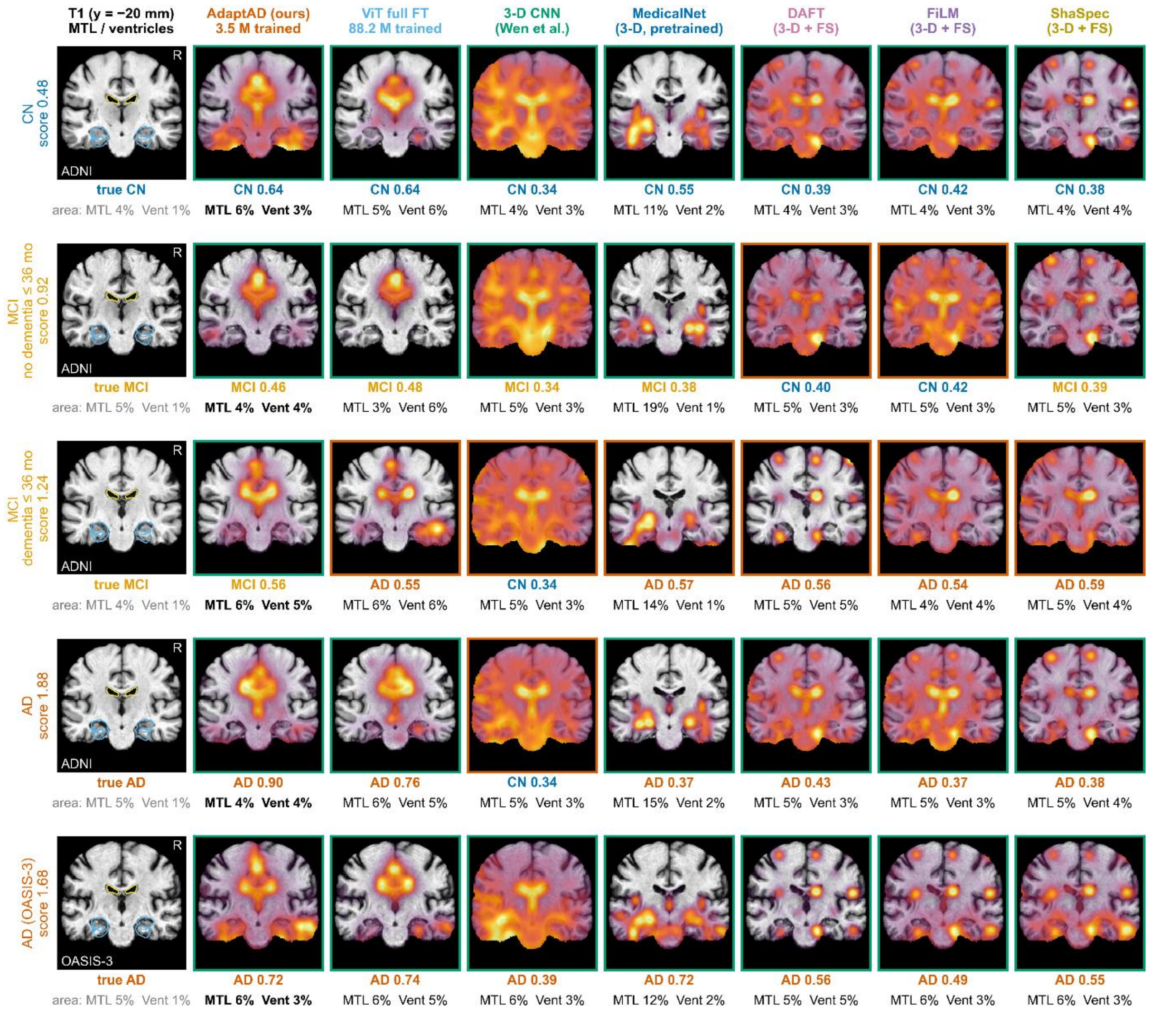

**Fig. 11.** The same participants read by seven image-based methods. Rows: the participants of Fig. 7, each read by the fold model that held them out (ADNI) or by all five-fold models (OASIS-3), with AdaptAD averaged over three seeds. Columns: the participant's registered T1 at y = −20 mm with the true stage, the medial temporal lobe (MTL; hippocampus, amygdala and parahippocampal gyrus, Harvard-Oxford; light blue) and the lateral ventricles (yellow) outlined, followed by the SmoothGrad saliency (|gradient × input|, eight noisy copies, noise SD 0.6 z-units) of the AD logit of each method. AdaptAD and ViT full fine-tuning are read through the MRI-only pathway with FreeSurfer replaced by the [MISSING] token; DAFT, FiLM and ShaSpec receive FreeSurfer, with gradients taken with respect to the image only. Each map is restricted to the brain and normalised to sum 1, and is shown relative to its own 99.5th percentile with 10 mm display smoothing. Below each map: the predicted stage with its calibrated probability (frame green when correct, vermillion when wrong), and the share of the plane's unsmoothed saliency in the MTL and lateral ventricles (Vent); the T1 column gives each region's share of brain area, the value expected without anatomical preference. For slice-token models, voxels where token planes intersect carry the average of the overlapping tokens' attributions. The atlas ventricle mask is smaller than the enlarged ventricles of older participants, so Vent is conservative. Rows are conditioned on AdaptAD being correct and are illustrative; population results over all 421 complete cases are given in Table S13 and the full probabilities in Table S14.

Three conclusions follow from these analyses. First, ventricular enlargement, an indirect but robust marker of atrophy, is a source of evidence shared by all the models. Second, low-rank adaptation retains the anatomical evidence of full adaptation: the maps and regional enrichment of AdaptAD closely track those of full fine-tuning, although AdaptAD trains 25 times fewer parameters. Third, the location of the evidence did not determine accuracy. The most hippocampus-focused model, MedicalNet, ranked below AdaptAD in three-class staging (Table 2), and the least focused, DAFT, ranked in the middle of the comparison.

## 6. Discussion

This study asked whether a vision foundation model pretrained on natural images, adapted with a small number of parameters, can stage the full continuum of Alzheimer's disease from a single T1-weighted scan, and whether it can do so without depending on morphometry that is often unavailable. Under a pre-registered protocol and a single locked external test, the answer is yes. In three-class staging on ADNI, AdaptAD ranked first of the ten models evaluated, ahead of seven comparison methods and two of its own variants, with a macro-AUC of 0.745 [0.709, 0.781]. It was significantly better than full fine-tuning of the same backbone, than ShaSpec and then a 3-D CNN, and it matched or exceeded full adaptation while training 25 times fewer parameters. It did not need morphometry, since removing FreeSurfer features changed its external accuracy by less than 0.003 and adding them did not improve accuracy. On the OASIS-3 external test, AdaptAD reached a macro-AUC of 0.780 on the primary clinical endpoint and 0.804 on the pre-specified CDR-based label, and on the latter it was significantly better than every comparison method that does not use a vision transformer. Its continuous disease score also carried clinical meaning, tracking cognitive and functional severity and stratifying the time to dementia diagnosis in participants with MCI. The following subsections discuss the reasons for these results, their relation to previous work, their implications for clinical practice and the limits within which they hold.

### 6.1 Transfer of a two-dimensional foundation model to volumetric MRI

Three results point to a consistent explanation of why a two-dimensional model pretrained on natural images transfers to three-dimensional brain MRI. First, adaptation was necessary, but its size was not critical. Freezing the backbone cost 0.037 in macro-AUC, the largest effect of any ablation, whereas low-rank updates of rank 4, 8 and 16 performed within 0.01 of one another (Table 3, Fig. 5a). The pretrained features are therefore useful but not sufficient on their own: a modest change to how the backbone processes MRI slices is needed, and a low-rank change is enough to provide it. Second, adapting all 88 M weights did not help. Full fine-tuning was 0.021 below low-rank adaptation with the same seed and 0.037 below the three-seed ensemble on ADNI, with a higher calibration error, and it only tied with AdaptAD on OASIS-3. This is the pattern expected when model capacity exceeds what about 400 training participants per fold can constrain, and it is the situation that motivated parameter-efficient fine-tuning in the first place [13]. Our results agree with recent evidence that parameter-efficient methods are underused in medical image analysis and most advantageous when labelled data are scarce [14], and with reports that low-rank adaptation can match or exceed full fine-tuning in medical diagnosis [27]. They extend that evidence to the staging of a neurodegenerative continuum from volumetric MRI.

Third, the 3-D CNN trained from scratch was the weakest model on ADNI (0.616), and the 3-D ResNet with medical pretraining, MedicalNet [5], did not close the gap to the pretrained two-dimensional model. This mirrors the observation that carefully controlled 3-D networks offer limited gains on cohorts of this size [4]. A pretrained two-dimensional backbone, a small trainable update and a fixed anatomical sampling of the volume together give the model a strong prior and few ways to overfit. The occlusion analysis (Fig. 6) clarifies the role of the anatomical sampling. No single plane was decisive, so the model does not depend on any one slice being perfectly registered. Yet the eight coronal planes through the medial temporal lobe preserved nearly all of the macro-AUC (0.745 against 0.751 with all sixteen planes), whereas the eight axial planes alone lost almost 0.05 (0.704). The fixed placement of planes in MNI space therefore does work that a three-dimensional network would have to learn from data: it guarantees that the anatomy most affected early in the disease is presented to the model in every participant, at the same token position and in the same orientation. The result also suggests that the representation could be simplified. A model trained from the start on the coronal planes alone may retain the accuracy of the present design at lower cost, although this remains to be tested rather than assumed.

These findings position AdaptAD relative to the emerging family of foundation models built specifically for brain MRI. Self-supervised models trained on large collections of three-dimensional brain scans [22], [23] aim to learn domain-specific representations that transfer across neurological tasks, and the rapid growth of vision transformer studies for AD [19], [20], [21] reflects the same ambition. Such models require large-scale three-dimensional pretraining that few groups can reproduce. AdaptAD shows that a widely available two-dimensional backbone, adapted with about 1% of its weights, already provides a strong and reproducible baseline for staging. The two approaches are complementary rather than competing: low-rank adaptation could equally be applied to a brain-specific backbone, and whether such a backbone would add to the anatomical prior provided by fixed slice placement is an open question for future work.

### 6.2 Morphometry as an optional input

AdaptAD was designed for settings in which FreeSurfer morphometry may be missing, and the evaluation turned that design into a finding: once the MRI expert was calibrated, morphometry added nothing. Late fusion scored 0.732 against 0.745 for the MRI expert alone on ADNI and 0.778 against 0.780 on OASIS-3. Removing morphometry from every participant changed the late fusion by at most 0.011, and participants without FreeSurfer were, if anything, staged slightly better than those with it (0.766 against 0.751 overall). The ablations add an important qualification, which we report as it stands. The [MISSING] token, the dropout of morphometry during training and the distillation from the full to the MRI-only prediction did not individually change the accuracy of the MRI expert. The robustness of AdaptAD to missing morphometry therefore comes mainly from the strength of its MRI-only pathway, and from a late-fusion design in which no prediction depends on morphometry being present, rather than from any one missing-modality component. This is consistent with the observation that the robustness of multimodal transformers to missing inputs depends strongly on how the modalities are combined [29]. In our case, the most robust combination was one in which the image carries the decision and the tabular input is used only when it adds information.

This finding contrasts with the assumptions behind much of the fusion literature. Methods such as FiLM [16], DAFT [17] and ShaSpec [18] were designed to exploit tabular data alongside images, and recent multimodal systems for dementia combine imaging with a broad range of clinical variables [6], [7] or learn to accept any combination of available modalities [30]. These approaches are valuable when the inputs are available and harmonised. In our evaluation, however, the image and tabular methods did not outperform the image-only AdaptAD on either cohort, and the model based on morphometry alone transferred least well between cohorts. A plausible contributor is the change of FreeSurfer version between cohorts, from UCSF FreeSurfer 6 in ADNI to FreeSurfer 5.3 in OASIS-3, which is known to alter volume and thickness estimates [48]. Image-derived features learned end to end from a standardised preprocessing pipeline appear to be less sensitive to this kind of pipeline shift than pre-computed morphometry. For clinical practice, the implication is direct. A site can run AdaptAD on a T1-weighted scan after standard registration, without a FreeSurfer run that takes hours, requires visual quality control and, as the version shift suggests, does not transfer cleanly between centres.

### 6.3 Generalisation under label, severity and prevalence shift

Every method, including the weakest, scored higher on OASIS-3 than on ADNI, and this uniform rise must be interpreted with care. It reflects differences between the cohorts rather than an improvement of the models. OASIS-3 contains a large proportion of clearly normal participants, its MCI group is small (66 participants, 6.6%), and its labels combine a clinician's diagnosis with the CDR. We anticipated this by pre-specifying two secondary endpoints that do not depend on the free-text diagnosis: a three-class label based on the CDR alone, and the binary distinction between normal and impaired participants. On the CDR-based label, AdaptAD's lead over every method that does not use a vision transformer became significant after correction, and on the binary endpoint it was among the two best methods. The ranking of methods was therefore stable across the external endpoints, whereas the absolute values moved with the definition of the label. The structure of the external errors is informative. AdaptAD recognised most cognitively normal participants but assigned 38% of participants with AD to the MCI stage. OASIS-3 participants with AD were considerably milder than those in ADNI (mean CDR-SB 3.5 against 5.7), so many of them fall near the boundary between MCI and AD that the model learned from the development cohort. The errors are therefore concentrated at the adjacent boundary of the continuum, which is the error a staging model should make when disease severity shifts between populations, rather than distributed across it.

The cost of the shift appeared in calibration more than in discrimination. All models over-predicted impairment in OASIS-3. This is the expected consequence of applying models calibrated in a cohort where 48% of participants are impaired to one where only 26% are, a form of dataset shift known to degrade calibration [36] and to alter several common performance measures at deployment [37]. The remedy is simple and well established: the class priors can be re-estimated and the outputs adjusted for the new prevalence [56], or the class bias of the calibration can be refitted on a small local sample. Because such adjustments change probabilities but not the ranking of participants, a new site can use the ranking by risk as it is and should recalibrate before interpreting the probabilities themselves. The lower

AUC on the MAGNETOM Vida scanner (0.688) should be read in the light of its precision. It rests on 100 participants, only four of whom had MCI, and its interval overlaps those of the other scanners. It does not, on its own, demonstrate a failure on that scanner, but it marks a setting in which further data are needed.

### 6.4 Clinical and anatomical interpretation of the model output

A staging model is more useful if its output behaves as a measure of disease rather than merely a label, and several results indicate that the AdaptAD disease score does. The score correlated with MMSE and CDR-SB in both cohorts. More importantly, it did so within diagnostic groups: within participants with AD, its correlation with CDR-SB was 0.591. This property cannot arise from learning the three labels alone, and it shows that the score grades severity within a stage as well as between stages. The score also anticipated the clinical course of participants with MCI. Seventeen of the twenty-one dementia diagnoses during follow-up occurred in the highest score tertile, the tertiles separated by 24 months, and by 57 months only 35% of the highest tertile remained free of dementia, against 96% of the lowest. In decision-curve analysis, the score derived from MRI alone provided greater net benefit than the morphometry model for the decision to classify a participant as having AD. These properties, rather than the AUC alone, are what a clinician would need before trusting such a score, and they justify presenting the model as a stager of the continuum (Figs. 7 and 8). The anatomy of the model's evidence is consistent with the known topography of the disease. AdaptAD's saliency was enriched most in the lateral ventricles, the parahippocampal gyrus, the temporal fusiform cortex and the hippocampus, and least in the frontal pole and lateral occipital cortex (Fig. 10). This mirrors the early involvement of the medial temporal lobe described by neuropathological staging [3] and the established role of medial temporal atrophy and ventricular enlargement as imaging markers of AD [1]. Ventricular enlargement is an indirect marker of the surrounding tissue loss, so the model partly reads the space that atrophy leaves behind, a robust signal that is less sensitive to registration than the boundaries of small structures.

The comparison across methods (Fig. 11, Table S13) adds a nuance that we consider important for the interpretation of saliency in this field. Every model concentrated its evidence on the ventricles, but the methods differed in the medial temporal lobe. MedicalNet, whose backbone was pretrained on segmentation tasks, was the most focused on the hippocampal region. AdaptAD and full fine-tuning were moderately focused, and the 3-D CNN trained from scratch and the fusion methods were not focused at all. This location did not determine accuracy: the most hippocampus-focused model ranked below AdaptAD, and the least focused ranked in the middle. Saliency maps therefore show what evidence a model uses, not why it performs better, and they should not be taken as proof of superiority. The close resemblance between the maps of AdaptAD and of full fine-tuning is the more meaningful observation, because it indicates that low-rank adaptation retains the anatomical evidence of full adaptation while training a small fraction of the parameters. These interpretations are bounded by what attribution methods can show. A saliency map indicates which voxels most influenced a given output; it does not localise pathology. The model-randomisation test [54], in which re-initialising the trainable weights reduced the correlation with the trained maps to 0.22, supports the view that the maps reflect learned behaviour rather than image contrast. It does not prove that the attribution is a faithful description of the model's reasoning.

### 6.5 Limitations

Several limitations qualify these results. The study used one development cohort and one external cohort, and both are research cohorts whose participants are, on average, well-educated and carefully characterised. Neither represents the heterogeneity of a general memory clinic. OASIS-3 also contributed only 66 participants with MCI and 100 scans from the MAGNETOM Vida scanner, too few for precise estimates in these subgroups. A second external cohort, such as AIBL or NACC, would test generalisation further. The prognostic analyses rest on short follow-up and few events. Only 12 participants with MCI converted to dementia within 36 months, so the corresponding AUC should be regarded as exploratory. The survival analysis over 143 participants provides stronger evidence, but few participants remained at risk after 48 months, and the late survival estimates are therefore imprecise. The comparison methods and the ablations are single models trained with one seed under one shared training protocol rather than with per-method tuning. A single AdaptAD seed still ranked above every comparison method on both cohorts, which limits the concern that ensembling alone explains its lead. The seed-to-seed variability of the comparison methods was nevertheless not measured, and their performance under method-specific tuning may differ. The 2.5-D representation samples 16 planes centred on the medial temporal lobe. Atrophy outside these planes, for example in the parietal cortex, is seen only where the planes happen to intersect it, so the model may underuse information that a whole-brain representation would capture. The model was also trained and evaluated only for the three stages CN, MCI and AD. Participants with other dementias or reversible causes of cognitive impairment were excluded from the external labels, and the behaviour of the model in these groups, which are common in clinical practice, remains unknown. Morphometry came from different FreeSurfer versions in the two cohorts and was deliberately not harmonised, so that the external test would reflect what a new site would actually supply. This choice, however, limits the conclusions that can be drawn about fusion with harmonised morphometry. Three analyses were carried out after the method was locked and are descriptive, namely the token occlusion, the comparison of saliency across methods and its population summary. None of them influenced the method or any reported performance. The example participants in Figs. 7 and 11 were chosen by a fixed rule but are conditioned on AdaptAD staging them correctly, so they illustrate rather than test the model;

two confident errors are shown in Fig. S6. In addition, the calibration of all models shifted with the prevalence of impairment in the external cohort, and their probabilities therefore require local recalibration before they can be interpreted at a new site. The ADNI cohort comprised one recent 3T scan per participant from a subset of the full database, and training on a larger part of ADNI, including additional visits and earlier phases, may further improve performance.

### 6.6 Implications and future directions

AdaptAD requires only a T1-weighted scan, a standard registration and about 3.5 M trained parameters. It returns a calibrated stage and a continuous severity score, and it does not depend on a morphometry pipeline. These properties make it a practical candidate for settings in which a structural scan is acquired routinely but quantitative analysis is not. Its adoption would require two steps that this study motivates directly: local recalibration of the class priors at each site, and prospective evaluation in a memory clinic, where the case mix, scanner protocols and diagnostic workflow differ from those of research cohorts. Several directions follow from the present findings. The occlusion results suggest testing a model trained from the start on the coronal planes through the medial temporal lobe, which may retain the accuracy of the current design at lower cost. Evaluation in further cohorts would establish how well the model generalises across populations and scanners, and extension of the label set would test whether the same approach supports the differential diagnosis of other dementias, a task for which multimodal approaches have recently shown promise [7]. Finally, because the disease score behaves as a continuous measure of severity, its change across repeated scans of the same individual may offer a way to monitor progression. That possibility can be examined in the longitudinal data that ADNI and OASIS-3 already provide.

## 7. Conclusion

This study shows that a vision foundation model pretrained on natural images can stage the full Alzheimer's disease continuum from a single T1-weighted scan when it is adapted efficiently and presented with the right anatomy. AdaptAD reads a registered scan as sixteen anatomically fixed slice tokens through the medial temporal lobe and adapts a frozen ViT-B/16 with low-rank updates, training 3.5 M of its 89 M parameters. Under a pre-registered protocol with identical folds for ten methods, it ranked first in three-class staging on ADNI, with a macro-AUC of 0.745, and was significantly better than full fine-tuning of the same backbone, an image and tabular fusion method and a 3-D CNN. In a single locked test on OASIS-3 it reached 0.780 on the primary endpoint and 0.804 on the CDR-based label, where it significantly outperformed every method that does not use a vision transformer. Low-rank adaptation matched or exceeded full adaptation with 25 times fewer trainable parameters, and the coronal planes carried most of the decision, since occluding all of them lowered macro-AUC by 0.047 whereas occluding all axial planes changed it by only 0.006, with the highest sensitivity in the coronal planes at $y = -10$ to $-20$ mm. Morphometry proved optional, since the model needed no FreeSurfer output, removing it changed external macro-AUC by less than 0.003, and adding it did not improve staging. Beyond discrete labels, the disease score graded clinical severity within each diagnostic group and stratified the time to dementia in mild cognitive impairment, with seventeen of twenty-one diagnoses occurring in its highest tertile. Its evidence concentrated on the lateral ventricles, the parahippocampal and fusiform cortex and the hippocampus, regions whose atrophy is characteristic of the disease, whereas the amygdala received no more attention than its size predicts. Together, these results indicate that accurate and robust staging of the Alzheimer's continuum does not require large three-dimensional models or quantitative morphometry. A parameter-efficient adaptation of a widely available backbone, combined with local recalibration of class priors, offers a practical and reproducible route to staging from the routine structural scan acquired in nearly every memory clinic. Validation in further cohorts and scanners, together with prospective clinical evaluation, are the next steps towards that use.

## Ethics statement

This study used data from the Alzheimer's Disease Neuroimaging Initiative (ADNI) and the third release of the Open Access Series of Imaging Studies (OASIS-3), both de-identified research databases approved by institutional review boards. Data collection procedures for ADNI and OASIS-3 were approved by the institutional review board at each participating site, and written informed consent was obtained from all participants or their authorised representatives at the time of original data collection. No additional ethical approval was required for the present study, which is a secondary analysis of fully de-identified imaging and clinical data obtained under the ADNI and OASIS-3 data use agreements.

## Funding

This study was supported by the National Science and Technology Council (NSTC), Taiwan, under grant number 113-2923-E-038-002-MY3. The funder had no role in study design, data analysis, interpretation or the decision to submit.

## Acknowledgements

Data used in the preparation of this article were obtained from the Alzheimer's Disease Neuroimaging Initiative (ADNI) database (adni.loni.usc.edu). As such, the investigators within the ADNI contributed to the design and implementation of ADNI and/or provided data but did not participate in the analysis or writing of this report. A complete listing of ADNI investigators can be found at http://adni.loni.usc.edu/wp-content/uploads/how_to_apply/ADNI_Acknowledgement_List.pdf.

Data collection and sharing for ADNI were funded by the Alzheimer's Disease Neuroimaging Initiative (National Institutes of Health Grant U01 AG024904) and DOD ADNI (Department of Defense award number W81XWH-12-2-0012). ADNI is funded by the National Institute on Aging, the National Institute of Biomedical Imaging and Bioengineering, and through generous contributions from the following: AbbVie, Alzheimer's Association, Alzheimer's Drug Discovery Foundation, Araclon Biotech, BioClinica Inc., Biogen, Bristol-Myers Squibb Company, CereSpir Inc., Cogstate, Eisai Inc., Elan Pharmaceuticals Inc., Eli Lilly and Company, EuroImmun, F. Hoffmann-La Roche Ltd and its affiliated company Genentech Inc., Fujirebio, GE Healthcare, IXICO Ltd., Janssen Alzheimer Immunotherapy Research & Development LLC, Johnson & Johnson Pharmaceutical Research & Development LLC, Lumosity, Lundbeck, Merck & Co. Inc., Meso Scale Diagnostics LLC, NeuroRx Research, Neurotrack Technologies, Novartis Pharmaceuticals Corporation, Pfizer Inc., Piramal Imaging, Servier, Takeda Pharmaceutical Company, and Transition Therapeutics. The Canadian Institutes of Health Research provides funds to support ADNI clinical sites in Canada. Private sector contributions are facilitated by the Foundation for the National Institutes of Health (www.fnih.org). The grantee organisation is the Northern California Institute for Research and Education, and the study is coordinated by the Alzheimer's Therapeutic Research Institute at the University of Southern California. ADNI data are disseminated by the Laboratory for Neuro Imaging at the University of Southern California. Data were provided in part by OASIS-3: Longitudinal Multimodal Neuroimaging. Principal investigators: T. Benzinger, D. Marcus, J. Morris; NIH P30 AG066444, P50 AG00561, P30 NS09857781, P01 AG026276, P01 AG003991, R01 AG043434, UL1 TR000448, R01 EB009352. AV-45 doses were provided by Avid Radiopharmaceuticals, a wholly owned subsidiary of Eli Lilly.

# Appendix

**Table S1.** Mapping of OASIS-3 clinical diagnoses to the pre-registered labels, and derivation of the external cohort.

**(A) Clinician's diagnosis text (dx1) mapped to label class.** n is the number of CDR visits with that text, before cohort selection.

| dx1 text | n visits | Class | → CN | → MCI | → AD |
|---|---|---|---|---|---|
| Cognitively normal | 6401 | CN | 6400 | 0 | 0 |
| AD Dementia | 1145 | AD | 0 | 0 | 1145 |
| uncertain dementia | 475 | MCI | 0 | 471 | 0 |
| Unc: ques. Impairment | 80 | MCI | 0 | 80 | 0 |
| DLBD, primary | 48 | OTHER | 0 | 0 | 0 |
| AD dem w/depresss, not contribut | 46 | AD | 0 | 0 | 46 |
| "." | 43 | UNMAPPED | 0 | 0 | 0 |
| No dementia | 37 | NODEM | 37 | 0 | 0 |
| AD dem w/depresss, contribut | 25 | AD | 0 | 0 | 25 |
| Vascular Demt, primary | 24 | OTHER | 0 | 0 | 0 |
| 0.5 in memory only | 23 | MCI | 0 | 23 | 0 |
| Incipient demt PTP | 23 | MCI | 0 | 23 | 0 |
| DAT | 19 | AD | 0 | 0 | 19 |
| Frontotemporal demt. prim | 19 | OTHER | 0 | 0 | 0 |
| uncertain, possible NON AD dem | 18 | OTHER | 0 | 0 | 0 |
| Dementia/PD, primary | 18 | OTHER | 0 | 0 | 0 |
| Non AD dem, Other primary | 17 | OTHER | 0 | 0 | 0 |
| AD dem Language dysf with | 16 | AD | 0 | 0 | 16 |
| AD dem Language dysf after | 14 | AD | 0 | 0 | 14 |
| AD dem w/CVD not contrib | 13 | AD | 0 | 0 | 13 |
| AD dem w/PDI after AD dem not contrib | 12 | AD | 0 | 0 | 12 |
| AD dem Language dysf prior | 10 | AD | 0 | 0 | 10 |
| AD dem w/oth (list B) contribut | 10 | AD | 0 | 0 | 10 |
| AD dem w/CVD contribut | 10 | AD | 0 | 0 | 10 |
| Unc: impair reversible | 7 | OTHER | 0 | 0 | 0 |
| AD dem w/PDI after AD dem contribut | 7 | AD | 0 | 0 | 7 |
| AD dem cannot be primary | 6 | UNMAPPED | 0 | 2 | 2 |
| AD dem distrubed social, after | 6 | AD | 0 | 0 | 6 |
| AD dem distrubed social, with | 6 | AD | 0 | 0 | 6 |
| AD dem w/oth (list B) not contrib | 5 | AD | 0 | 0 | 5 |
| Non DAT, Other primary | 4 | UNMAPPED | 0 | 0 | 0 |
| Incipient Non-AD dem | 4 | OTHER | 0 | 0 | 0 |
| Vascular Demt, secondary | 3 | OTHER | 0 | 0 | 0 |
| AD dem w/oth unusual features/demt on | 3 | AD | 0 | 0 | 3 |
| AD dem visuospatial, with | 3 | AD | 0 | 0 | 3 |
| DAT w/depresss, not contribut | 3 | UNMAPPED | 0 | 0 | 0 |
| AD dem/FLD prior to AD dem | 2 | UNMAPPED | 0 | 0 | 0 |
| DAT Language dysf after | 2 | UNMAPPED | 0 | 0 | 0 |
| AD dem w/oth unusual feat/subs demt | 2 | AD | 0 | 0 | 2 |
| DAT w/CVD contribut | 2 | UNMAPPED | 0 | 0 | 0 |
| AD dem visuospatial, prior | 1 | AD | 0 | 0 | 1 |
| AD dem visuospatial, after | 1 | AD | 0 | 0 | 1 |
| AD dem w/Frontal lobe/demt at onset | 1 | AD | 0 | 0 | 1 |
| AD dem distrubed social, prior | 1 | AD | 0 | 0 | 1 |
| AD dem w/oth unusual features | 1 | AD | 0 | 0 | 1 |
| DAT distrubed social, after | 1 | UNMAPPED | 0 | 0 | 0 |
| DLBD, secondary | 1 | OTHER | 0 | 0 | 0 |
| DAT w/depresss, contribut | 1 | UNMAPPED | 0 | 0 | 0 |
| DAT w/CVD not contrib | 1 | UNMAPPED | 0 | 0 | 0 |
| DAT w/PDI after DAT not contrib | 1 | OTHER | 0 | 0 | 0 |
| DAT distrubed social, prior | 1 | UNMAPPED | 0 | 0 | 0 |
| Frontotemporal demt. secn | 1 | OTHER | 0 | 0 | 0 |
| ProAph w/o dement | 1 | UNMAPPED | 0 | 0 | 0 |
| Q | 1 | UNMAPPED | 0 | 0 | 0 |

**(B) Derivation of the external cohort**

| Step | Result |
|---|---|
| CDR visits with a diagnosis text | 8,625: primary labels CN 6,437, MCI 599, AD 1,359; excluded 230 (other dementia or reversible cause 161, unmapped text 64, MCI text with CDR ≠ 0.5: 4, normal text with CDR > 0: 1) |
| Participants with a 3T T1 scan | 1,104 |
| With a CDR visit within 180 days of the earliest such scan | 1,006 (median gap 75 days) |
| Primary label at that visit | CN 738, MCI 67, AD 190, excluded 11 |
| CDR-only label at that visit | CDR 0: 738, CDR 0.5: 202, CDR ≥ 1: 66 |
| After excluding 2 scans that were not 3-D whole-brain | 993 (CN 738, MCI 66, AD 189) |
| FreeSurfer (passed or passed with edits, within 180 days) | 988 of 993 |

**(C) Primary label against CDR-only label (n = 1,006)**

| Primary \ CDR-only | CDR 0 (CN) | CDR 0.5 (MCI) | CDR ≥ 1 (AD) |
|---|---|---|---|
| CN | 738 | 0 | 0 |
| MCI | 0 | 67 | 0 |
| AD | 0 | 126 | 64 |
| Excluded | 0 | 9 | 2 |

*Classes: CN requires a normal text and CDR 0; MCI requires CDR 0.5 with an impairment-without-dementia text; AD requires CDR ≥ 0.5 with AD named as the primary diagnosis; OTHER (non-AD or reversible) and UNMAPPED texts received no label. The mapping and labels were fixed before any OASIS-3 prediction. Panel C shows that two-thirds of the participants with a clinical AD diagnosis had CDR 0.5, which is why the clinical and CDR-based labels differ.*

**Table S2.** Mapping of OASIS-3 FreeSurfer 5.3 output to the 132 ADNI (UCSF FreeSurfer 6) features.

**(A) Summary**

| | **n** |
|---|---|
| ADNI features | 132 |
| Matched automatically by region name | 119 |
| Matched manually (global measures) | 9 |
| No OASIS-3 equivalent, set to missing | 4 (ST21SV, ST66SV, ST80SV, ST125SV) |
| Median OASIS-3 / ADNI ratio of cognitively normal medians, volumes | 0.95 |
| Median OASIS-3 / ADNI ratio of cognitively normal medians, thickness | 0.97 |

**(B) Manually mapped features**

| ADNI code | OASIS-3 column |
|---|---|
| ST147SV | lhCortexVol |
| ST148SV | rhCortexVol |
| ST149SV | CortexVol |
| ST150SV | lhCorticalWhiteMatterVol |
| ST151SV | rhCorticalWhiteMatterVol |
| ST152SV | CorticalWhiteMatterVol |
| ST153SV | SubCortGrayVol |
| ST154SV | TotalGrayVol |
| ST155SV | SupraTentorialVol |

*Volumes are divided by intracranial volume × 1,000 in both cohorts. The full 132-row table, with ADNI code, ADNI description, OASIS-3 column, mapping source, cognitively normal medians in each cohort and their ratio, is provided as a machine-readable file (fs_map_report.csv). Individual ratios range widely (for example 0.78 for the right pallidum and 1.12 for the right putamen), and no harmonisation was applied.*

**Table S3.** Gate 2: choice of backbone on validation data only.

| Backbone | Expert source | Pooled validation AUC (3-seed ensemble) | Auxiliary head | MRI-only | Per seed | Per fold |
|---|---|---|---|---|---|---|
| ViT-B/16 (LoRA) | MRI-only | **0.781** | 0.765 | 0.781 | 0.766, 0.747, 0.754 | 0.758, 0.803, 0.830, 0.749, 0.826 |
| DINOv2 ViT-B/14 (LoRA) | MRI-only | 0.765 | 0.750 | 0.765 | 0.749, 0.747, 0.696 | 0.766, 0.789, 0.846, 0.729, 0.819 |

*Pre-registered rule: adopt DINOv2 only if its pooled validation MRI-expert AUC is at least 0.01 above that of ViT (threshold 0.791). DINOv2 minus ViT = −0.016, so ViT-B/16 was retained. Test data were not read. DINOv2 was re-run with the training-collapse guard switched off (protocol amendment, 2026-10-10); the guard never fired for ViT, so both backbones share the same effective recipe.*

**Table S4.** Gate 3 lock of the evaluated method.

| Field | Value |
|---|---|
| Locked | 2026-10-12, 12:35 |
| Backbone | vit_base_patch16_224.augreg2_in21k_ft_in1k |
| Reference runs | 3 (seeds 42, 142, 242) |
| Files hashed (SHA-256) | 71 (checkpoints, model files and inference code) |
| Rule | Ablations explain, never change, the method; the external test runs once on this lock |
| External test run | 2026-10-12, 13:51 (once) |

*The complete list of the 71 files with their SHA-256 hashes is provided as a machine-readable file (LOCK.json).*

**Table S5.** Model size and training cost.

**(A) Parameters**

| Method | Family | Total | Trainable | of which LoRA | Trainable (%) |
|---|---|---|---|---|---|
| AdaptAD (LoRA r = 8, locked) | 2.5-D ViT | 89,325,574 | 3,526,918 | 1,179,648 | 3.95 |
| AdaptAD, LoRA r = 4 | 2.5-D ViT | 88,735,750 | 2,937,094 | 589,824 | 3.31 |
| AdaptAD, LoRA r = 16 | 2.5-D ViT | 90,505,222 | 4,706,566 | 2,359,296 | 5.20 |
| AdaptAD, linear probe | 2.5-D ViT | 88,145,926 | 2,347,270 | 0 | 2.66 |
| ViT full fine-tuning | 2.5-D ViT | 88,145,926 | 88,145,926 | 0 | 100 |
| 3D CNN (Wen et al.) | 3-D CNN | 6,351,623 | 6,351,623 | 0 | 100 |
| MedicalNet (Med3D) | 3-D CNN | 32,987,715 | 32,987,715 | 0 | 100 |
| DAFT | 3-D CNN + tabular | 3,616,668 | 3,616,668 | 0 | 100 |
| FiLM | 3-D CNN + tabular | 3,630,637 | 3,630,637 | 0 | 100 |
| ShaSpec | 3-D CNN + tabular | 3,953,125 | 3,953,125 | 0 | 100 |
| FreeSurfer logistic regression | tabular | 399 | 399 | 0 | 100 |

**(B) Training time, five folds**

| Run | Machine | Epochs (all folds) | Median s/epoch | Total (h) |
|---|---|---|---|---|
| AdaptAD, seed 42 | L | 246 | 85.0 | 5.79 |
| AdaptAD, seed 142 | L | 236 | 84.0 | 6.14 |
| AdaptAD, seed 242 | not recorded | 244 | 64.0 | 4.38 |
| ViT full fine-tuning | S | 250 | 63.0 | 4.35 |
| 3D CNN (Wen et al.) | L | 95 | 17.0 | 0.50 |
| MedicalNet (Med3D) | S | 156 | 38.0 | 1.66 |
| DAFT | L | 100 | 15.0 | 0.41 |
| FiLM | L | 120 | 15.0 | 0.49 |
| ShaSpec | L | 121 | 15.0 | 0.51 |
| DINOv2 (Gate 2, seed 42) | not recorded | 248 | 112.0 | 9.99 |
| Ablation: linear probe | S | 249 | 42.0 | 2.92 |
| Ablation: LoRA r = 16 | S | 255 | 69.0 | 4.91 |
| Ablation: LoRA r = 4 | S | 233 | 75.0 | 4.89 |
| Ablation: [MISSING] → mask | S | 234 | 66.0 | 4.31 |
| Ablation: [MISSING] → zero | S | 239 | 67.0 | 4.54 |
| Ablation: no auxiliary head | S | 250 | 72.0 | 5.06 |
| Ablation: no FreeSurfer dropout | S | 243 | 71.0 | 4.77 |
| Ablation: no distillation | S | 241 | 64.0 | 4.32 |
| Ablation: no stage-1 selection | S | 240 | 65.0 | 4.32 |
| Ablation: one stage | S | 201 | 66.0 | 3.68 |

*Machine: L, NVIDIA GeForce RTX 4060 Laptop GPU (8 GB); S, NVIDIA GeForce RTX 2080 Ti (11 GB). Seconds per epoch are medians over all epochs and folds and are comparable only within a machine. ViT full fine-tuning and the LoRA runs share every training setting except the set of trainable weights (and the backbone learning rate, Table S6).*

**Table S6.** Training configurations.

**(A) AdaptAD, its ablations and ViT full fine-tuning**

| Setting | Locked recipe | Differs in |
|---|---|---|
| Adaptation | LoRA, r = 8, α = 16 | linear probe: frozen; r = 4 (α = 8); r = 16 (α = 32); full fine-tuning: all weights |
| Learning rate, LoRA other trainable | $2 \times 10^{-4}$ / $3 \times 10^{-4}$ | full fine-tuning: backbone $1 \times 10^{-5}$ |
| Weight decay, LoRA / other | $10^{-4}$ / 0.05 | |
| Batch × gradient accumulation | 4 × 2 (effective 8) | |
| Epochs / warm-up / patience | 80 / 5 / 15 | |
| Stage-1 (MRI-only) epochs | 30 | one stage: 0 |
| FreeSurfer dropout p | 0.3 | no FreeSurfer dropout: 0 |
| Distillation weight / temperature | 1.0 / 2.0 | no distillation: 0 |
| MRI-only loss weigh / auxiliary weight | 0.5 / 1.0 | no auxiliary head: 0 |
| Label smoothing / dropout / stochastic depth | 0.1 / 0.1 / 0.1 | |
| Missing FreeSurfer | learned [MISSING] token | zero vector; attention mask |
| Stage-1 checkpoint selection | on | stage 2 starts from the last stage-1 epoch |
| Collapse guard | at stage-1 epoch 8, a fold whose mean training loss over the last 3 epochs was ≥ 1.08 would be restarted with a new seed (at most 2 restarts); not triggered in the locked runs | |
| Backbone | ViT-B/16, 224 × 224 input, pretrained on ImageNet-21k and fine-tuned on ImageNet-1k (timm identifier vit_base_patch16_224.augreg2_in21k_ft_in1k) | |

**(B) 3-D comparison methods (3D CNN, MedicalNet, DAFT, FiLM, ShaSpec)**

| Setting | Value |
|---|---|
| Learning rate, trunk / heads | $3 \times 10^{-4}$ (MedicalNet, pretrained: $1 \times 10^{-4}$) / $1 \times 10^{-3}$ |
| Weight decay | $10^{-4}$ |
| Batch × gradient accumulation | 4 × 2 (effective 8) |
| Epochs / warm-up / patience | 80 / 3 / 15 |
| Label smoothing | 0.1 |
| ShaSpec FreeSurfer dropout during training | 0.3 |

*All runs: class-balanced cross-entropy, AdamW, linear warm-up followed by cosine decay (to 1% of the peak learning rate for AdaptAD), mixed precision, flip test-time augmentation*

**Table S7.** Results on all ADNI participants each method can score, under natural FreeSurfer availability (N = 597).

| Method | Coverage | n | Macro-AUC | BAcc | Macro-F1 | QWK |
|---|---|---|---|---|---|---|
| AdaptAD (ours) | 100.0% | 597 | 0.751 | 0.562 | 0.563 | 0.557 |
| AdaptAD, single seed | 100.0% | 597 | 0.741 | 0.553 | 0.555 | 0.563 |
| AdaptAD late fusion (ours, MRI + FS) | 100.0% | 597 | 0.740 | 0.545 | 0.541 | 0.544 |
| ViT full fine-tuning | 100.0% | 597 | 0.720 | 0.516 | 0.523 | 0.532 |
| FreeSurfer logistic regression | 70.5% | 421 | 0.719 | 0.548 | 0.523 | 0.510 |
| MedicalNet (Med3D) | 100.0% | 597 | 0.711 | 0.537 | 0.520 | 0.491 |
| FiLM | 100.0% | 597 | 0.678 | 0.473 | 0.464 | 0.397 |
| DAFT | 100.0% | 597 | 0.667 | 0.481 | 0.466 | 0.393 |
| ShaSpec | 100.0% | 597 | 0.656 | 0.494 | 0.460 | 0.345 |
| 3D CNN (Wen et al.) | 100.0% | 597 | 0.625 | 0.423 | 0.430 | 0.236 |

*Image and tabular methods use FreeSurfer where available and their MRI-only path otherwise; FreeSurfer logistic regression can score only the 421 participants with FreeSurfer.*

**Table S8.** Fold-level results and bootstrap rank probabilities (ADNI complete cases, n = 421).

**(A) Macro-AUC per test fold**

| Method | Fold 0 | Fold 1 | Fold 2 | Fold 3 | Fold 4 |
|---|---|---|---|---|---|
| AdaptAD (ours) | 0.750 | 0.811 | 0.758 | 0.655 | 0.776 |
| AdaptAD, single seed | 0.763 | 0.800 | 0.750 | 0.650 | 0.802 |
| AdaptAD late fusion | 0.750 | 0.777 | 0.753 | 0.653 | 0.779 |
| FreeSurfer logistic regression | 0.770 | 0.719 | 0.705 | 0.700 | 0.739 |
| 3D CNN (Wen et al.) | 0.569 | 0.706 | 0.603 | 0.678 | 0.545 |
| MedicalNet (Med3D) | 0.733 | 0.756 | 0.783 | 0.647 | 0.660 |
| ViT full fine-tuning | 0.749 | 0.689 | 0.727 | 0.620 | 0.745 |
| DAFT | 0.741 | 0.678 | 0.718 | 0.669 | 0.670 |
| FiLM | 0.740 | 0.702 | 0.719 | 0.667 | 0.692 |
| ShaSpec | 0.741 | 0.705 | 0.640 | 0.640 | 0.683 |

**(B) Probability of each rank over 2,000 bootstrap resamples (rank 1 = highest macro-AUC)**

| Method | 1 | 2 | 3 | 4 | 5 | 6 | 7 | 8 | 9 | 10 |
|---|---|---|---|---|---|---|---|---|---|---|
| AdaptAD (ours) | **0.601** | 0.338 | 0.044 | 0.013 | 0.004 | 0.001 | 0 | 0 | 0 | 0 |
| AdaptAD, single seed | 0.294 | **0.504** | 0.124 | 0.059 | 0.016 | 0.005 | 0 | 0.001 | 0 | 0 |
| AdaptAD late fusion | 0.021 | 0.090 | **0.636** | 0.187 | 0.049 | 0.014 | 0.003 | 0.001 | 0 | 0 |
| FreeSurfer logistic regression | 0.073 | 0.043 | 0.101 | **0.307** | 0.212 | 0.174 | 0.070 | 0.021 | 0.002 | 0 |
| MedicalNet (Med3D) | 0.007 | 0.012 | 0.055 | 0.178 | 0.237 | **0.239** | 0.142 | 0.105 | 0.027 | 0 |
| ViT full fine-tuning | 0 | 0.002 | 0.019 | 0.165 | **0.268** | 0.234 | 0.178 | 0.117 | 0.019 | 0 |
| FiLM | 0.005 | 0.009 | 0.015 | 0.063 | 0.130 | 0.195 | **0.300** | 0.249 | 0.036 | 0.001 |
| DAFT | 0.001 | 0.005 | 0.007 | 0.030 | 0.083 | 0.132 | 0.282 | **0.422** | 0.041 | 0 |
| ShaSpec | 0 | 0 | 0 | 0.001 | 0.003 | 0.008 | 0.026 | 0.087 | **0.854** | 0.023 |
| 3D CNN (Wen et al.) | 0 | 0 | 0 | 0 | 0 | 0 | 0 | 0.001 | 0.023 | **0.977** |

*Friedman test across folds p = 0.044; Nemenyi critical difference 6.1 ranks (five folds). Bold, most probable rank. Fold 3 was the hardest for most methods.*

**Table S9.** AdaptAD macro-AUC in ADNI subgroups (all 597 participants, out-of-fold).

| Subgroup | Level | n | Macro-AUC [95% CI] |
|---|---|---|---|
| All | | 597 | 0.751 [0.720, 0.781] |
| Age | < 70 years | 157 | 0.794 [0.737, 0.850] |
| Age | 70 to 80 years | 278 | 0.751 [0.711, 0.794] |
| Age | ≥ 80 years | 162 | 0.709 [0.647, 0.762] |
| Sex | Female | 312 | 0.770 [0.728, 0.809] |
| Sex | Male | 285 | 0.720 [0.679, 0.762] |
| APOE | ε4 non-carrier | 345 | 0.708 [0.666, 0.748] |
| APOE | ε4 carrier | 218 | 0.810 [0.766, 0.852] |
| Education | ≥ 16 years | 419 | 0.778 [0.744, 0.812] |
| Education | < 16 years | 178 | 0.709 [0.659, 0.759] |
| Scanner | Siemens | 437 | 0.749 [0.716, 0.782] |
| Scanner | Philips | 92 | 0.753 [0.682, 0.819] |
| Scanner | GE | 68 | 0.751 [0.656, 0.851] |
| FreeSurfer | available | 421 | 0.745 [0.710, 0.781] |
| FreeSurfer | missing | 176 | 0.766 [0.713, 0.812] |

*Bootstrap 95% CI stratified by class. Subgroups differ in their stage composition (for example, APOE ε4 carriers include a larger proportion of AD), so differences between levels reflect both case mix and model behaviour and are descriptive. APOE status was unavailable for 34 participants.*

**Table S10.** Progression from MCI to dementia in ADNI (n = 165 with MCI at the scan).

**(A) Groups and events**

| | n |
|---|---|
| MCI with follow-up after the scan | 143 |
| Progressed to dementia within 36 months (pMCI) | 12 |
| Free of dementia for at least 36 months (sMCI) | 53 |
| Neither (follow-up < 36 months without event, or event after 36 months) | 78 |
| Dementia diagnoses during follow-up (all 143) | 21 (low tertile 1, middle 3, high 17) |
| AUC of the disease score, pMCI vs sMCI | 0.786 [0.612, 0.926] |
| Log-rank test across score tertiles | $\chi^2 = 27.2$, $p = 1.3 \times 10^{-6}$ |

**(B) Kaplan-Meier proportion free of dementia (number at risk) by score tertile**

| Tertile | 6 mo | 12 mo | 18 mo | 24 mo | 30 mo | 36 mo | 48 mo | 57 mo |
|---|---|---|---|---|---|---|---|---|
| Low (n = 48) | 1.00 (34) | 1.00 (31) | 0.96 (21) | 0.96 (20) | 0.96 (18) | 0.96 (18) | 0.96 (14) | 0.96 (12) |
| Middle (n = 47) | 1.00 (32) | 0.97 (28) | 0.97 (25) | 0.97 (24) | 0.93 (21) | 0.93 (20) | 0.88 (15) | 0.88 (12) |
| High (n = 48) | 1.00 (33) | 0.97 (30) | 0.94 (25) | 0.86 (21) | 0.78 (18) | 0.65 (15) | 0.42 (9) | 0.35 (5) |

*Disease score s = p(MCI) + 2 p(AD), out-of-fold. Participant-level data are not reproduced here and are available with the code release under the ADNI data use agreement.*

**Table S11.** One-shot external test on OASIS-3, extended metrics.

**(A) Primary endpoint, clinical label (n = 988; CN 736, MCI 65, AD 187)**

| Method | Macro-AUC [95% CI] | BAcc [95% CI] | Macro-F1 | QWK | ECE | ΔAUC, AdaptAD minus method [95% CI] | p (Holm) |
|---|---|---|---|---|---|---|---|
| AdaptAD (ours) | 0.780 [0.748, 0.814] | 0.545 [0.497, 0.594] | 0.520 | 0.567 | 0.167 | reference | |
| AdaptAD, single seed | 0.784 [0.752, 0.817] | 0.552 [0.504, 0.601] | 0.527 | 0.573 | 0.175 | −0.004 [−0.010, +0.002] | 1.000 |
| AdaptAD late fusion | 0.778 [0.745, 0.813] | 0.558 [0.513, 0.605] | 0.539 | 0.565 | 0.138 | +0.002 [−0.007, +0.011] | 1.000 |
| FreeSurfer logistic regression | 0.744 [0.712, 0.776] | 0.522 [0.490, 0.557] | 0.492 | 0.473 | 0.109 | +0.036 [+0.003, +0.067] | 0.200 |
| 3D CNN (Wen et al.) | 0.753 [0.718, 0.788] | 0.551 [0.508, 0.596] | 0.526 | 0.528 | 0.342 | +0.027 [−0.000, +0.055] | 0.300 |
| MedicalNet (Med3D) | 0.749 [0.715, 0.781] | 0.538 [0.493, 0.586] | 0.509 | 0.525 | 0.201 | +0.031 [+0.004, +0.060] | 0.180 |
| ViT full fine-tuning | 0.780 [0.748, 0.814] | 0.548 [0.500, 0.598] | 0.528 | 0.589 | 0.144 | −0.000 [−0.011, +0.011] | 1.000 |
| DAFT | 0.770 [0.738, 0.803] | 0.514 [0.483, 0.546] | 0.456 | 0.386 | 0.142 | +0.010 [−0.021, +0.040] | 1.000 |
| FiLM | 0.748 [0.715, 0.781] | 0.513 [0.480, 0.549] | 0.475 | 0.424 | 0.211 | +0.033 [+0.002, +0.061] | 0.252 |
| ShaSpec | 0.742 [0.710, 0.775] | 0.540 [0.496, 0.586] | 0.491 | 0.441 | 0.160 | +0.038 [+0.000, +0.072] | 0.300 |

**(B) Secondary endpoint, CDR label (n = 999; CDR 0: 736, 0.5: 199, ≥ 1: 64)**

| Method | Macro-AUC [95% CI] | BAcc [95% CI] | Macro-F1 | QWK | ECE | ΔAUC, AdaptAD minus method [95% CI] | p (Holm) |
|---|---|---|---|---|---|---|---|
| AdaptAD (ours) | 0.804 [0.781, 0.827] | 0.613 [0.569, 0.659] | 0.558 | 0.572 | 0.132 | reference | |
| AdaptAD, single seed | 0.807 [0.784, 0.830] | 0.602 [0.556, 0.646] | 0.547 | 0.574 | 0.125 | −0.003 [−0.007, +0.002] | 0.504 |
| AdaptAD late fusion | 0.792 [0.769, 0.815] | 0.600 [0.557, 0.642] | 0.513 | 0.543 | 0.096 | +0.012 [+0.006, +0.019] | <0.001 |
| FreeSurfer logistic regressio | 0.740 [0.715, 0.763] | 0.575 [0.541, 0.604] | 0.421 | 0.413 | 0.068 | +0.065 [+0.041, +0.088] | <0.001 |
| 3D CNN (Wen et al.) | 0.769 [0.741, 0.795] | 0.590 [0.548, 0.629] | 0.481 | 0.480 | 0.296 | +0.036 [+0.015, +0.057] | 0.004 |
| MedicalNet (Med3D) | 0.754 [0.729, 0.779] | 0.602 [0.561, 0.642] | 0.498 | 0.483 | 0.182 | +0.050 [+0.030, +0.070] | <0.001 |
| ViT full fine-tuning | 0.805 [0.781, 0.828] | 0.601 [0.556, 0.648] | 0.548 | 0.579 | 0.131 | −0.001 [−0.009, +0.008] | 0.889 |
| DAFT | 0.775 [0.751, 0.798] | 0.564 [0.539, 0.586] | 0.382 | 0.321 | 0.057 | +0.029 [+0.008, +0.050] | 0.030 |
| FiLM | 0.763 [0.740, 0.784] | 0.574 [0.541, 0.602] | 0.412 | 0.371 | 0.141 | +0.042 [+0.021, +0.062] | <0.001 |

| ShaSpec | 0.752 [0.727, 0.776] | 0.555 [0.511, 0.594] | 0.438 | 0.397 | 0.111 | +0.052 [+0.029, +0.076] | <0.001 |
|---|---|---|---|---|---|---|---|

**(C) Macro-AUC by scanner, primary label**

| Scanner | n | AdaptAD | Single seed | Late fusion | FS LR | 3D CNN | MedicalNet | ViT full FT | DAFT | FiLM | ShaSpec |
|---|---|---|---|---|---|---|---|---|---|---|---|
| Siemens TrioTim | 614 | 0.788 | 0.791 | 0.783 | 0.750 | 0.753 | 0.754 | 0.790 | 0.771 | 0.753 | 0.757 |
| Siemens Biograph mMR | 274 | 0.787 | 0.786 | 0.790 | 0.736 | 0.760 | 0.735 | 0.782 | 0.768 | 0.736 | 0.723 |
| Siemens MAGNETOM Vida | 100 | 0.688 | 0.719 | 0.705 | 0.717 | 0.724 | 0.771 | 0.681 | 0.801 | 0.775 | 0.703 |

**(C) AdaptAD late fusion with FreeSurfer removed for a random fraction of participants**

| Removed | 0% | 25% | 50% | 75% | 100% |
|---|---|---|---|---|---|
| Macro-AUC (mean ± SD, 20 draws) | 0.778 ± 0.000 | 0.778 ± 0.003 | 0.778 ± 0.003 | 0.779 ± 0.003 | 0.780 ± 0.000 |

*1,004 participants entered preprocessing; 999 were predicted by every method. ΔAUC here is AdaptAD minus method, so positive values favour AdaptAD (the opposite sign convention from Table 2). Holm correction within each endpoint. Test run once, 2026-10-12, 13:51. On the Vida scanner, only 4 participants had MCI, so the per-scanner estimates there are imprecise for every method.*

**Table S12.** Regional enrichment of AdaptAD saliency (ADNI, correctly classified test participants, three seeds).

| Region (Harvard-Oxford) | Covered voxels | AD (n = 76) [95% CI] | CN (n = 222) [95% CI] |
|---|---|---|---|
| Lateral ventricles | 9,519 | 2.82 [2.70, 2.94] | 2.50 [2.42, 2.59] |
| Parahippocampal gyrus | 9,862 | 2.06 [1.95, 2.16] | 1.97 [1.89, 2.04] |
| Temporal fusiform cortex | 11,083 | 1.95 [1.84, 2.06] | 1.87 [1.80, 1.93] |
| Hippocampus | 9,171 | 1.74 [1.67, 1.81] | 1.47 [1.44, 1.51] |
| Cingulate gyrus, posterior | 8,770 | 1.59 [1.48, 1.70] | 1.22 [1.18, 1.27] |
| Thalamus | 16,445 | 1.50 [1.42, 1.58] | 1.27 [1.22, 1.32] |
| Inferior temporal gyrus | 20,046 | 1.46 [1.37, 1.56] | 1.38 [1.34, 1.42] |
| Temporal pole | 8,547 | 1.14 [1.01, 1.28] | 1.04 [0.99, 1.10] |
| Amygdala | 3,799 | 1.02 [0.95, 1.10] | 1.03 [1.00, 1.07] |
| Middle temporal gyrus | 30,465 | 0.85 [0.82, 0.89] | 0.92 [0.91, 0.94] |
| Precuneus | 4,023 | 0.69 [0.61, 0.79] | 0.69 [0.66, 0.72] |
| Insula | 13,462 | 0.56 [0.53, 0.59] | 0.66 [0.64, 0.68] |
| Frontal pole | 41,513 | 0.32 [0.29, 0.34] | 0.42 [0.40, 0.45] |
| Lateral occipital cortex | 19,667 | 0.28 [0.26, 0.31] | 0.40 [0.38, 0.43] |

*Enrichment = share of saliency in the region divided by the region's share of the brain volume covered by the 16 token planes (1 = no preference). SmoothGrad of |gradient × input| for the AD output; bootstrap 95% CI over participants. Sorted by AD enrichment. The enrichment is higher in AD than in CN for the hippocampus, posterior cingulate, thalamus and ventricles.*

**Table S13.** Distribution of saliency of seven image-based methods over all ADNI complete cases (n = 421).

| Method | MTL, plane y = −20 mm | MTL, 16 token planes | Ventricles, plane y = −20 mm | Ventricles, 16 token planes |
|---|---|---|---|---|
| AdaptAD (ours) | 1.38 [1.35, 1.41] | 1.64 [1.61, 1.67] | 3.24 [3.16, 3.32] | 2.54 [2.48, 2.60] |
| ViT full fine-tuning | 1.64 [1.59, 1.70] | 1.74 [1.68, 1.80] | 3.96 [3.80, 4.11] | 3.18 [3.06, 3.30] |
| 3D CNN (Wen et al.) | 1.14 [1.11, 1.17] | 1.02 [1.00, 1.05] | 2.62 [2.53, 2.70] | 2.57 [2.48, 2.68] |
| MedicalNet (Med3D) | **2.99 [2.93, 3.05]** | **2.73 [2.68, 2.78]** | 1.63 [1.56, 1.70] | 2.96 [2.90, 3.02] |
| DAFT | 0.98 [0.96, 0.99] | 0.91 [0.90, 0.92] | 3.90 [3.82, 3.99] | 2.44 [2.42, 2.47] |
| FiLM | 1.07 [1.04, 1.10] | 1.03 [1.00, 1.06] | 3.36 [3.24, 3.48] | 2.33 [2.28, 2.37] |
| ShaSpec | 1.27 [1.21, 1.34] | 1.18 [1.13, 1.24] | 2.98 [2.86, 3.11] | 2.21 [2.14, 2.29] |

*Values are the ratio of the mean share of saliency in the region to the region's share of brain area within the same plane set (1 = no anatomical preference), with bootstrap 95% CI over participants. Expected shares (chance): MTL 0.045 (plane) and 0.035 (token planes); ventricles 0.011 (plane) and 0.013 (token planes). MTL = hippocampus + amygdala + parahippocampal gyrus; ventricles = lateral ventricles (Harvard-Oxford, 25% threshold). Each participant was read by the fold model that held it out; same SmoothGrad procedure for every method. Post-lock, descriptive analysis.*

**Table S14.** Calibrated probabilities and saliency shares for the participants of Figs. 7 and 11.

| Participant | Method | Predicted | p(CN) | p(MCI) | p(AD) | MTL share | Vent share |
|---|---|---|---|---|---|---|---|
| CN | AdaptAD (ours) | **CN** ✓ | 0.639 | 0.244 | 0.117 | 0.060 | 0.027 |
| | ViT full fine-tuning | CN ✓ | 0.642 | 0.277 | 0.082 | 0.050 | 0.061 |
| | 3D CNN (Wen et al.) | CN ✓ | 0.337 | 0.329 | 0.333 | 0.037 | 0.027 |
| | MedicalNet (Med3D) | CN ✓ | 0.547 | 0.303 | 0.150 | 0.106 | 0.016 |
| | DAFT | CN ✓ | 0.390 | 0.356 | 0.255 | 0.038 | 0.035 |
| | FiLM | CN ✓ | 0.425 | 0.363 | 0.212 | 0.036 | 0.034 |
| | ShaSpec | CN ✓ | 0.385 | 0.367 | 0.249 | 0.039 | 0.042 |
| MCI, no dementia ≤ 36 mo | AdaptAD (ours) | **MCI** ✓ | 0.306 | 0.464 | 0.230 | 0.042 | 0.041 |
| | ViT full fine-tuning | MCI ✓ | 0.254 | 0.479 | 0.267 | 0.034 | 0.056 |
| | 3D CNN (Wen et al.) | MCI ✓ | 0.331 | 0.335 | 0.334 | 0.048 | 0.025 |
| | MedicalNet (Med3D) | MCI ✓ | 0.337 | 0.376 | 0.287 | 0.190 | 0.010 |
| | DAFT | CN ✗ | 0.397 | 0.366 | 0.237 | 0.048 | 0.029 |
| | FiLM | CN ✗ | 0.422 | 0.364 | 0.214 | 0.049 | 0.030 |
| | ShaSpec | MCI ✓ | 0.358 | 0.394 | 0.248 | 0.051 | 0.032 |
| MCI, dementia ≤ 36 mo | AdaptAD (ours) | **MCI** ✓ | 0.102 | 0.558 | 0.341 | 0.059 | 0.053 |
| | ViT full fine-tuning | AD ✗ | 0.073 | 0.379 | 0.548 | 0.065 | 0.059 |
| | 3D CNN (Wen et al.) | CN ✗ | 0.339 | 0.330 | 0.331 | 0.049 | 0.027 |
| | MedicalNet (Med3D) | AD ✗ | 0.057 | 0.373 | 0.570 | 0.138 | 0.010 |
| | DAFT | AD ✗ | 0.086 | 0.358 | 0.556 | 0.055 | 0.052 |
| | FiLM | AD ✗ | 0.083 | 0.378 | 0.538 | 0.043 | 0.039 |
| | ShaSpec | AD ✗ | 0.185 | 0.227 | 0.588 | 0.046 | 0.036 |
| AD | AdaptAD (ours) | **AD** ✓ | 0.026 | 0.072 | 0.903 | 0.043 | 0.042 |
| | ViT full fine-tuning | AD ✓ | 0.059 | 0.178 | 0.763 | 0.055 | 0.052 |
| | 3D CNN (Wen et al.) | CN ✗ | 0.340 | 0.327 | 0.333 | 0.049 | 0.029 |
| | MedicalNet (Med3D) | AD ✓ | 0.259 | 0.370 | 0.371 | 0.148 | 0.018 |
| | DAFT | AD ✓ | 0.245 | 0.326 | 0.429 | 0.054 | 0.035 |
| | FiLM | AD ✓ | 0.295 | 0.337 | 0.367 | 0.051 | 0.033 |
| | ShaSpec | AD ✓ | 0.293 | 0.327 | 0.379 | 0.052 | 0.038 |
| AD (OASIS-3) | AdaptAD (ours) | **AD** ✓ | 0.033 | 0.252 | 0.715 | 0.062 | 0.035 |
| | ViT full fine-tuning | AD ✓ | 0.037 | 0.224 | 0.739 | 0.064 | 0.049 |
| | 3D CNN (Wen et al.) | AD ✓ | 0.286 | 0.320 | 0.395 | 0.062 | 0.027 |
| | MedicalNet (Med3D) | AD ✓ | 0.074 | 0.211 | 0.715 | 0.123 | 0.016 |
| | DAFT | AD ✓ | 0.137 | 0.304 | 0.559 | 0.046 | 0.046 |
| | FiLM | AD ✓ | 0.227 | 0.285 | 0.488 | 0.059 | 0.033 |
| | ShaSpec | AD ✓ | 0.166 | 0.283 | 0.550 | 0.060 | 0.030 |

*Calibrated probabilities exactly as reported in Tables 2 and 4. ✓/✗, predicted stage correct/incorrect. Shares are on the coronal plane $y = -20$ mm. Brain-area shares (chance): MTL 0.045 (0.045 to 0.046 across participants), ventricles 0.011. Participants were chosen by a fixed rule conditioned on AdaptAD being correct, so AdaptAD's 5/5 is by construction and is not a performance measure.*

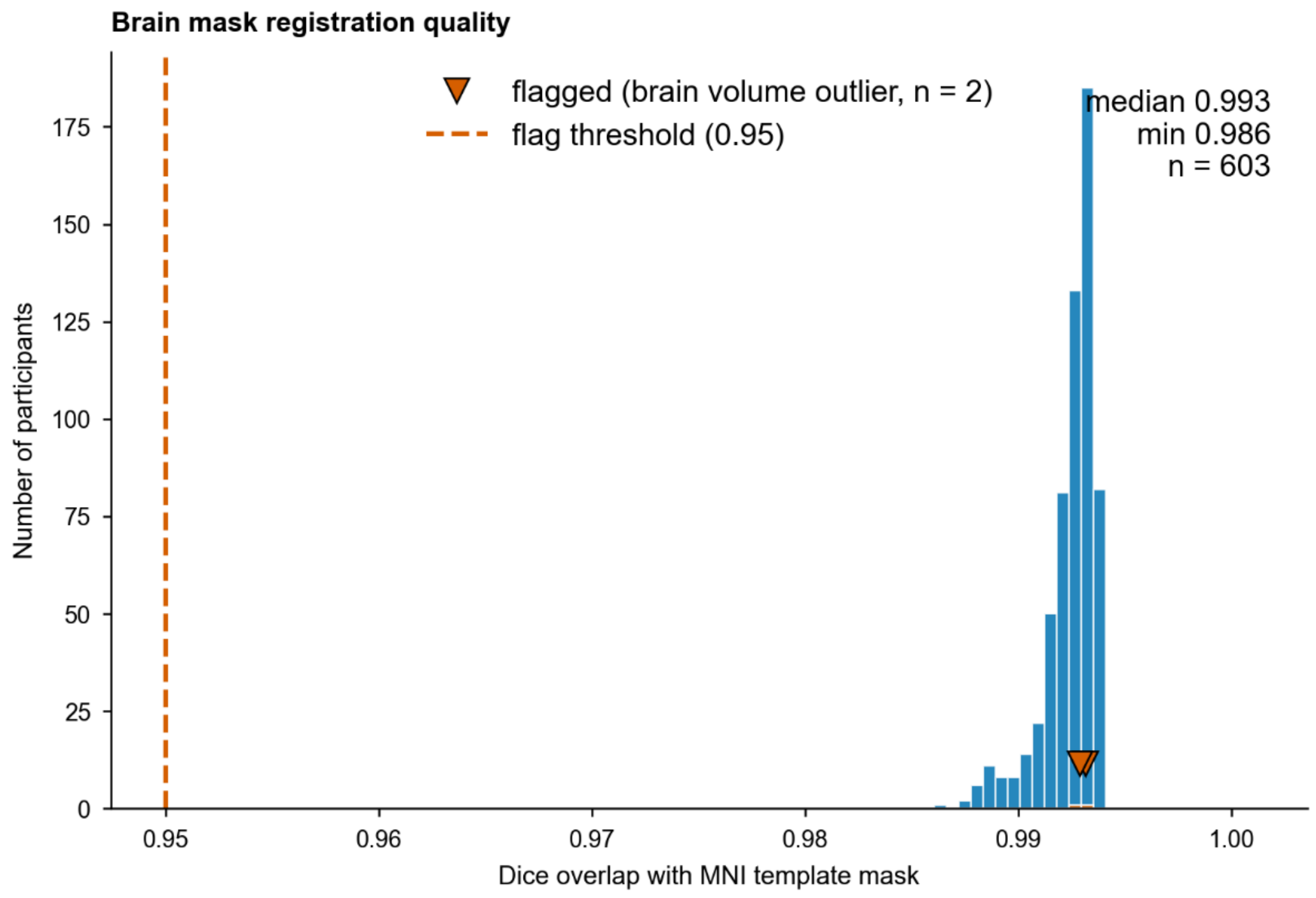

**Fig. S1. Quality control of MRI registration to the MNI template.** The histogram shows the Dice overlap between each participant's registered brain mask and the MNI152 template mask for the 597 participants in the final cohort (median 0.993, range 0.986 to 0.994). The dashed vertical line marks the pre-specified flag threshold of 0.95. No scan fell below this threshold, so the two scans that were flagged were identified by the brain volume outlier criterion rather than by overlap, and they are marked with red triangles.

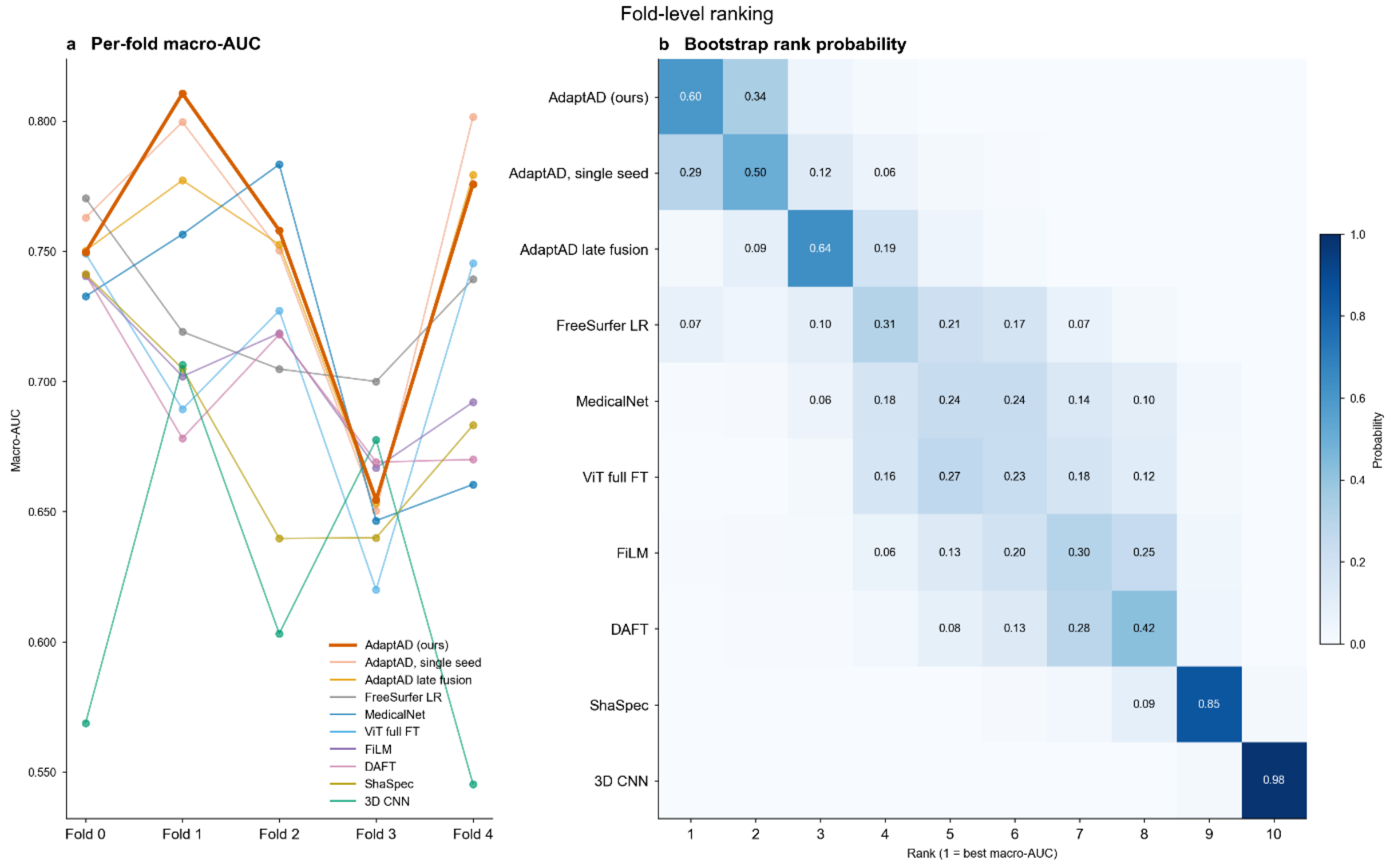

**Fig. S2. Fold-level ranking of all methods.** Panel a shows the macro-AUC of each method in each of the five cross-validation test folds, with AdaptAD drawn as a thicker line. Panel b shows the probability that each method occupies each rank, estimated from 2,000 paired stratified bootstrap replicates on the 421 complete cases, with methods ordered by mean rank. Because five folds give limited power to separate methods, these rankings describe the data and are not used for inference.

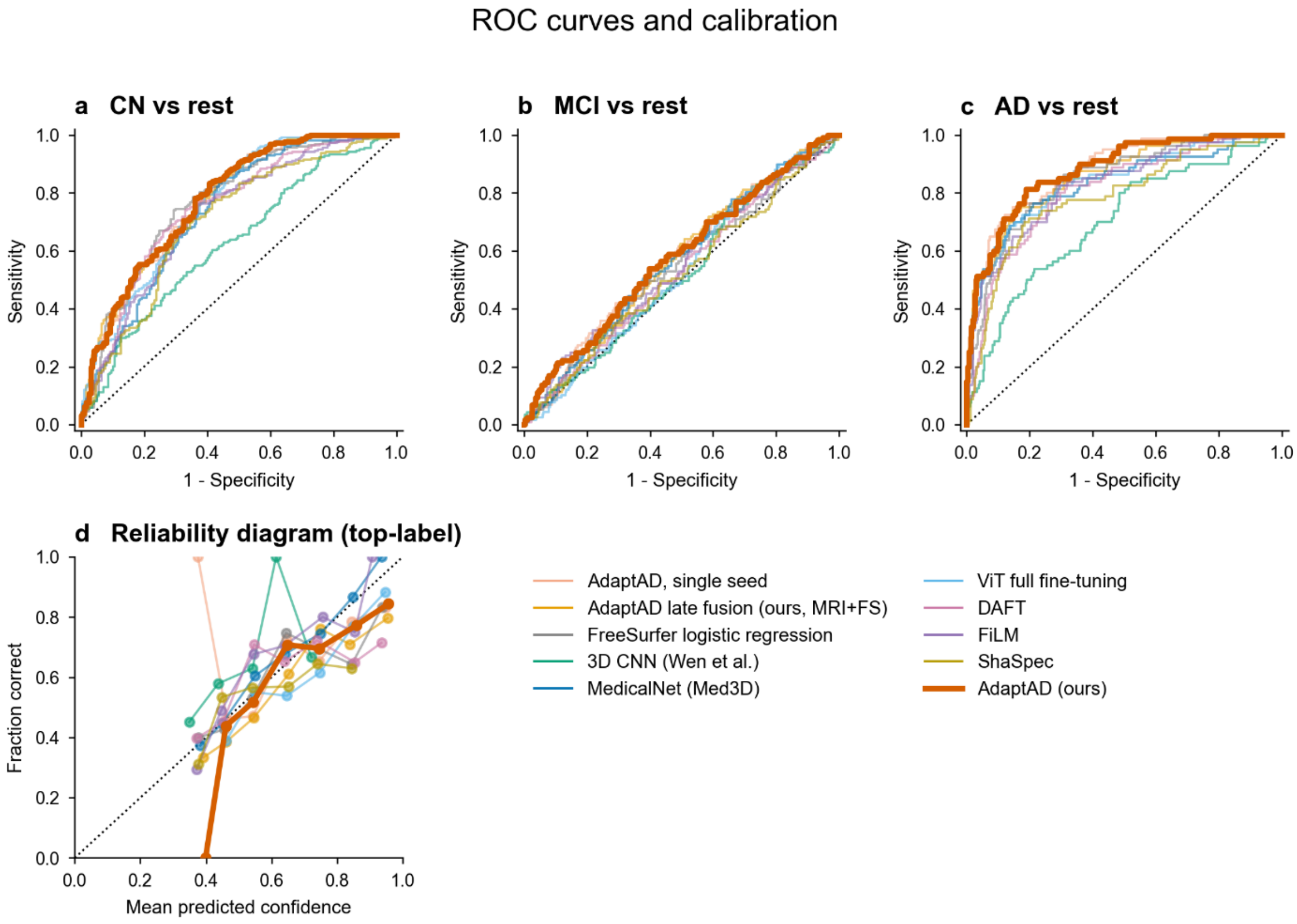

**Fig. S3. Discrimination and calibration on ADNI.** Panels a to c show one-versus-rest receiver operating characteristic curves for CN, MCI and AD, pooled over the five test folds for the 421 complete cases. Panel d shows the reliability diagram for the top-label prediction, which plots the fraction of correct predictions against the mean predicted confidence in ten bins. The dotted diagonal marks perfect discrimination in panels a to c and perfect calibration in panel d, and AdaptAD is drawn as the thicker line in every panel.

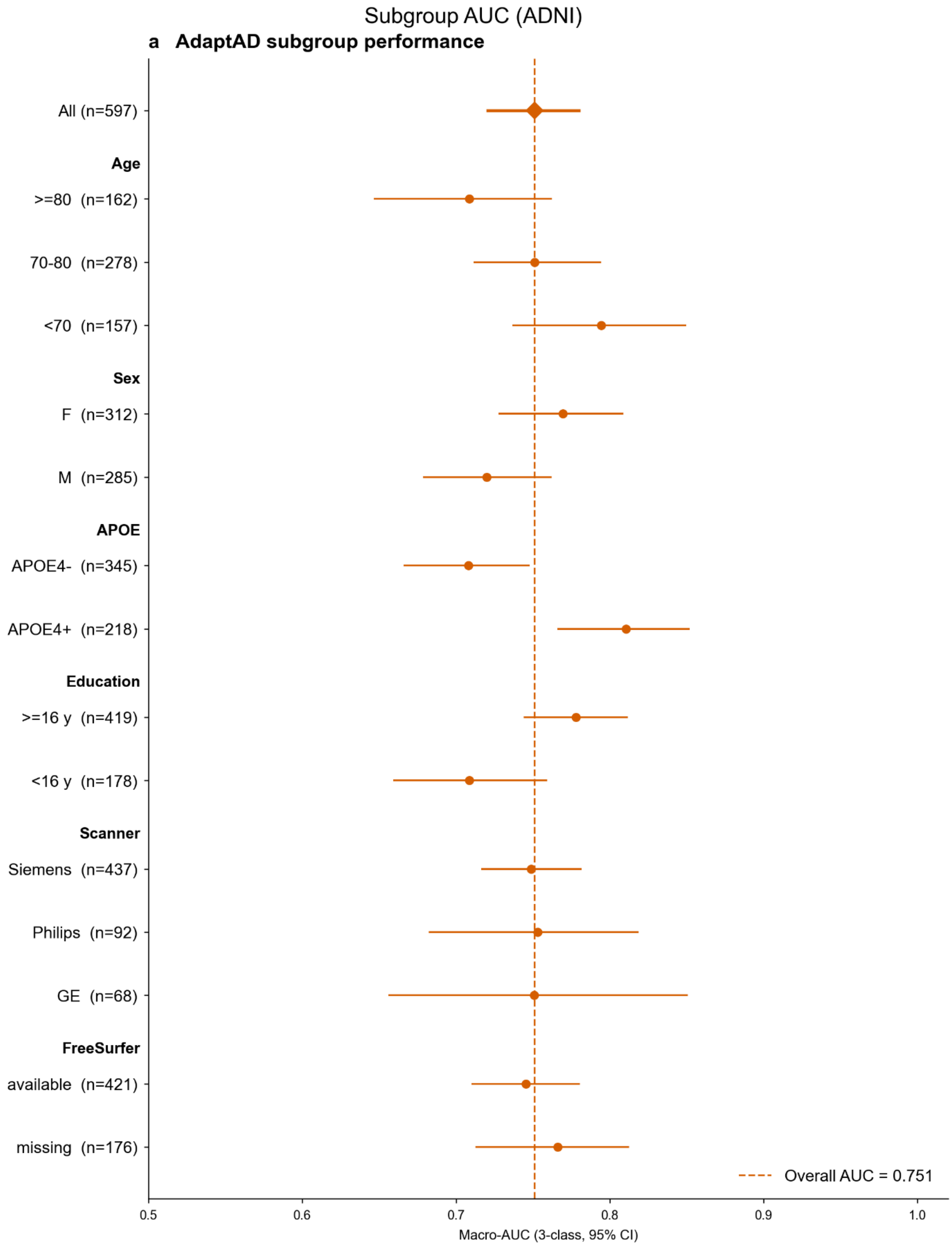

**Fig. S4. Performance of AdaptAD across participant subgroups.** The forest plot shows the three-class macro-AUC of out-of-fold AdaptAD predictions for subgroups defined by age (below 70, 70 to 80, and 80 or older), sex, APOE ε4 carrier status, education (16 years or more versus fewer), scanner manufacturer, and FreeSurfer availability. Horizontal bars show bootstrap 95% confidence intervals, group sizes are given in the row labels, and the dashed line marks the overall macro-AUC on all 597 participants (0.751). The 34 participants without an APOE result are not included in the APOE rows. Intervals are not corrected for multiple comparisons, the differences between subgroups were not formally tested, and subgroups with small samples have wide intervals and should be read with caution.

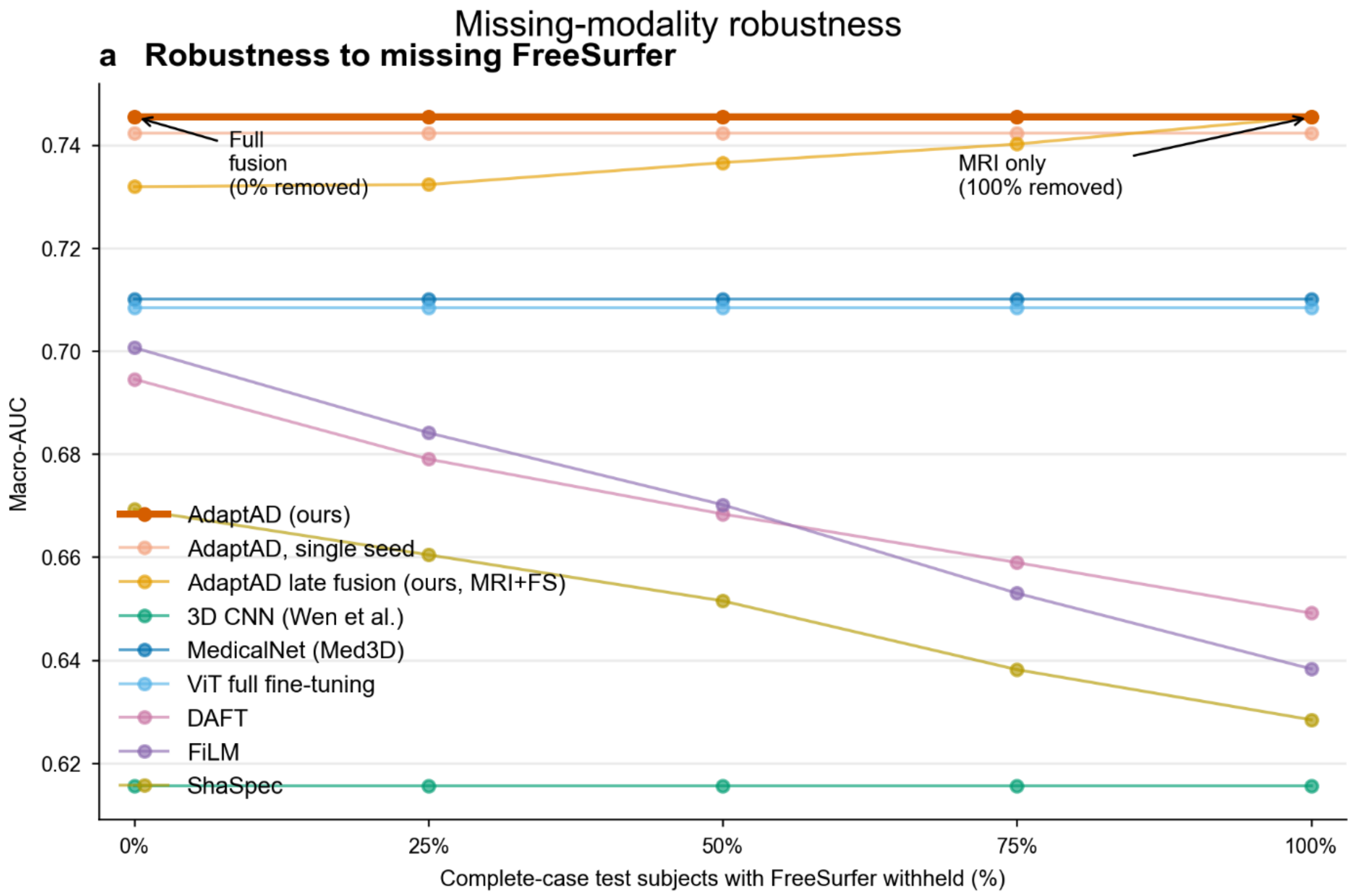

**Fig. S5. Robustness to missing FreeSurfer morphometry.** Macro-AUC is plotted against the percentage of complete-case test participants (n = 421) for whom FreeSurfer measurements were withheld, from 0% (all participants with FreeSurfer) to 100% (none). Each point is the mean of 20 random draws of the withheld participants. Methods that never use FreeSurfer are flat by construction. The late-fusion model improved slightly as FreeSurfer was removed (0.732 to 0.745, the level of the MRI-only model), whereas the image and tabular fusion methods DAFT, FiLM and ShaSpec declined steadily, by 0.045, 0.062 and 0.041.

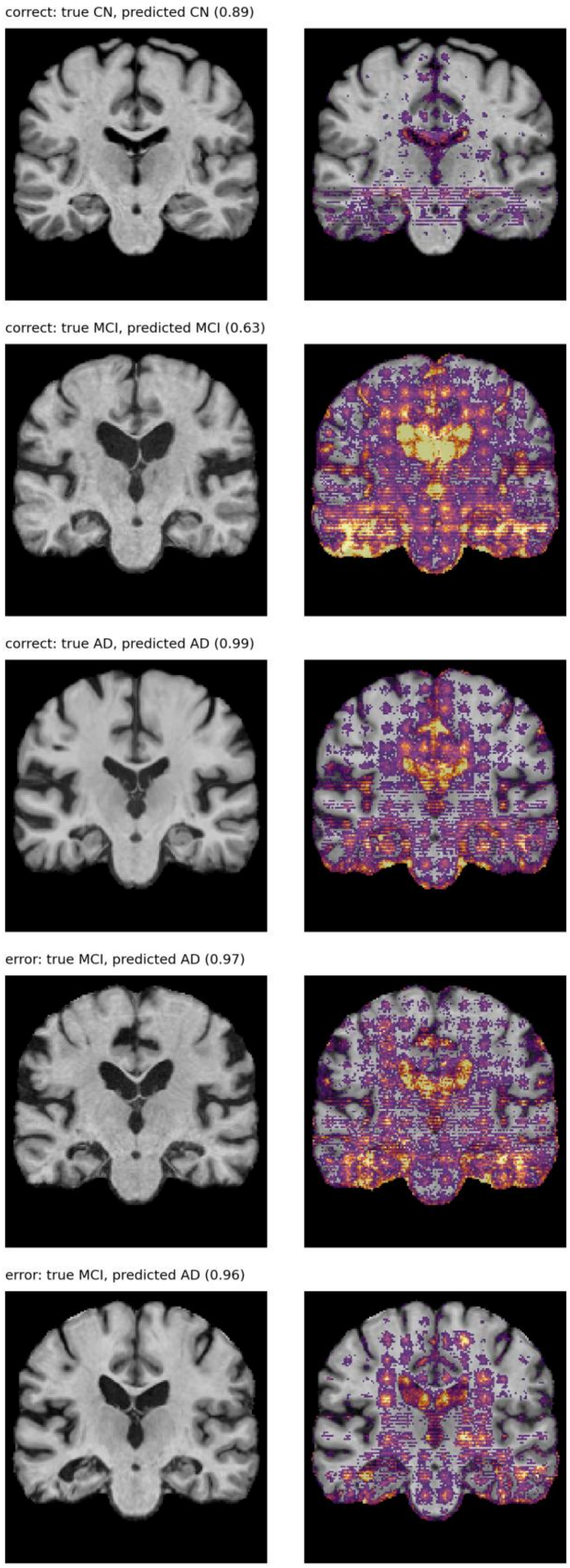

**Fig. S6. Saliency maps of individual participants.** Each row shows one ADNI participant in the same coronal plane, with the registered T1 image on the left and the SmoothGrad gradient-times-input saliency for the AD output of the MRI-only model overlaid on the right. The rows show the highest-confidence correct predictions for CN, MCI and AD, followed by the two most confident errors, both MCI participants predicted as AD. The number in each panel title is the predicted probability of the predicted class. Brighter regions contributed more to the AD output, and values below 15% of the maximum are not shown. Maps are not smoothed, so the patch grid of the vision transformer and the slice-plane banding are visible. Because the target is the AD output, the CN row shows where the model attended when it assigned low AD probability, and the maps are descriptive and not evidence that any region is causal.